%% file: acl_latex.tex
\pdfoutput=1

\documentclass[11pt]{article}

\usepackage{acl}

\usepackage{times}
\usepackage{latexsym}

\usepackage[T1]{fontenc}

\usepackage[utf8]{inputenc}

\usepackage{microtype}

\usepackage{inconsolata}
\usepackage[ruled]{algorithm2e}
\usepackage{ bbold }

\usepackage[ruled]{algorithm2e}
\usepackage{ bbold }

\usepackage{hyperref}
\usepackage{url}
\usepackage{mathtools}
\usepackage{ textcomp }
\usepackage{multirow}
\usepackage{graphicx}
\usepackage{amsthm}
\usepackage{amssymb}

\usepackage{subcaption}
\newtheorem{definition}{Definition}

\usepackage{amsmath, amsthm}
\newtheorem{Assumption}{Assumption}
\newcommand\norm[1]{\left\lVert#1\right\rVert}

\usepackage{color}

\title{Language Agnostic Code Embeddings}

\author{Saiteja Utpala \\
  Cohere For AI \\
  \texttt{saitejautpala@gmail.com} \\\And
  Alex Gu \\
 MIT \\
  \texttt{gua@mit.edu} \\\And
  Pin Yu Chen \\
  IBM Research \\
  \texttt{pin-yu.chen@ibm.com} \\ }

\begin{document}
\maketitle
\begin{abstract}
Recently, code language models have achieved notable advancements in addressing a diverse array of essential code comprehension and generation tasks. Yet, the field lacks a comprehensive deep dive and understanding of the code embeddings of multilingual code models. In this paper, we present a comprehensive study on multilingual code embeddings, focusing on the cross-lingual capabilities of these embeddings across different programming languages. Through probing experiments, we demonstrate that code embeddings comprise two distinct components: one deeply tied to the nuances and syntax of a specific language, and the other remaining agnostic to these details, primarily focusing on semantics. Further, we show that when we isolate and eliminate this language-specific component, we witness significant improvements in downstream code retrieval tasks, leading to an absolute increase of up to +17 in the Mean Reciprocal Rank (MRR).
\end{abstract}

\section{Introduction}

Large language models (LLMs) have made remarkable progress in code-related tasks, exemplified by models such as Codex, which powers GitHub Copilot and offers automated code suggestions within integrated development environments (IDEs) \cite{chen2021evaluating}. These models achieve their proficiency through extensive training on vast code datasets, providing them with versatile contextual understanding for a range of coding tasks \cite{husain2019codesearchnet, athiwaratkun2022multi, zhu2022xlcost}. However, it's worth noting that decoder-only models may not always be the optimal choice for retrieval tasks when compared to encoder models \cite{nijkamp2023codegen2, wang2021codet5, wang2023codet5+}.

While previous studies indicate that language models trained on a variety of natural languages exhibit strong cross-lingual traits \cite{pires2019multilingual}, their multilingual representations can be dissected into a language-specific syntax component and a language-agnostic semantic component \cite{chang2022geometry}. Moreover, eliminating the language-specific elements can enhance retrieval tasks and counteract "language bias", a tendency for representations to cluster by language instead of meaning \cite{roy2020lareqa, yang2021simple, xie2022discovering}. We aim to determine if similar patterns are evident in multilingual code models pretrained on programming languages (e.g., C, C++, Python) as opposed to natural languages (e.g., English, French, Spanish). We specifically address:

\begin{enumerate}
    \item Can representations of these code models be categorized into language-specific and language-agnostic components?
    \item If so, does removing the language-specific components enhance the consistency and comparability of code representations (alignment) across programming languages, thereby improving downstream code retrieval tasks? 
\end{enumerate}



\begin{figure*}[t!]
\centering
\includegraphics[height=0.4\linewidth]{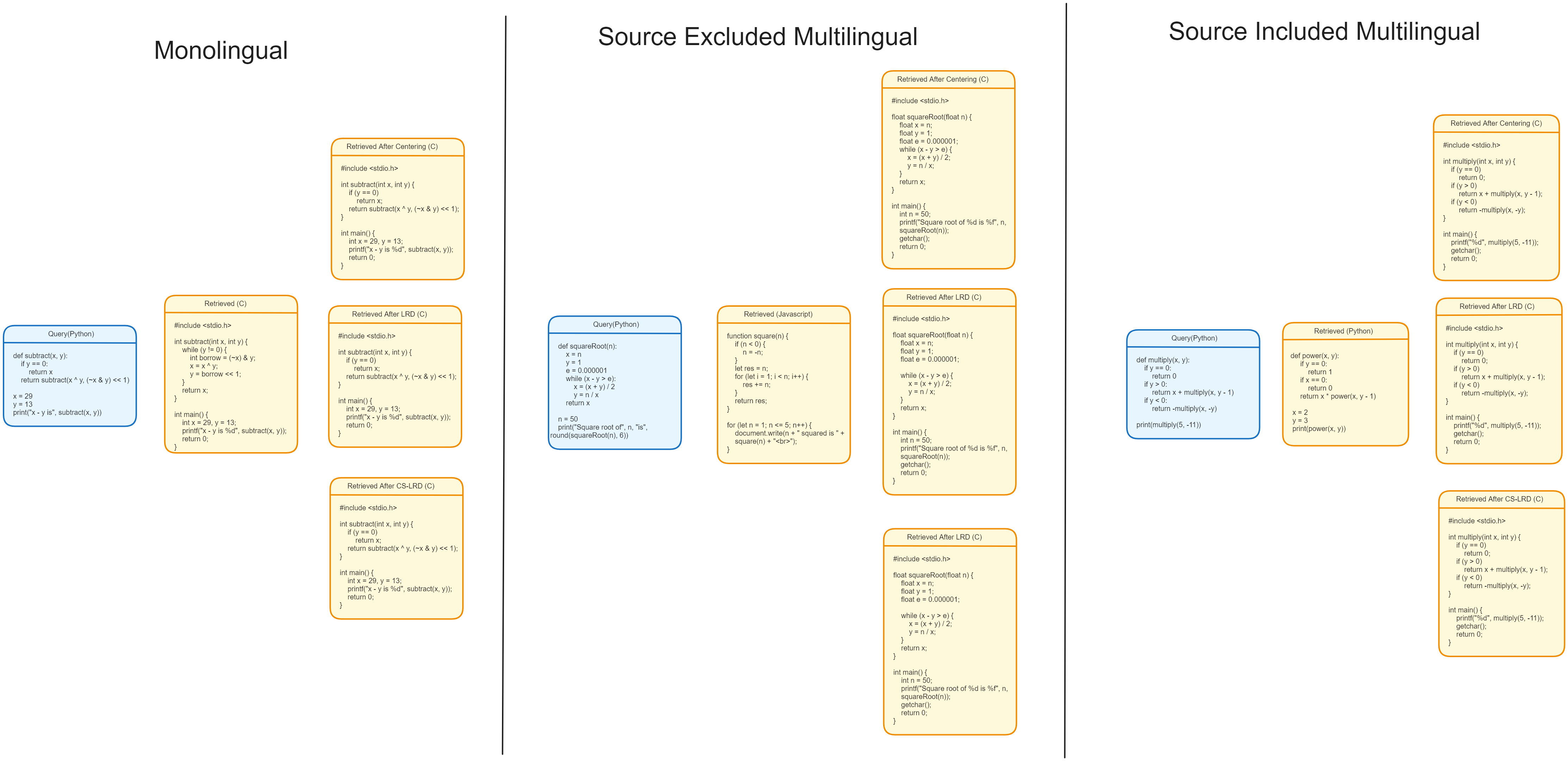}
\caption{Illustration of the top retrieved results for code-to-code search, where the query is in Python, and the target is in C. Language composition varies across retrieval databases: 'Monolingual' (C only), 'Source Excluded Multilingual' (several languages except Python), 'Source Included Multilingual' (several languages including Python). Demonstrates improved semantic matching and reduced language bias after removing language-specific components. }
\end{figure*}


Our comprehensive evaluations confirm these patterns in code language models. Our contributions are summarized as follows: 
\begin{itemize} 

    \item We investigate the cross-lingual properties of pretrained multilingual code language models, examining them through the lens of both language-specific (syntax) and language-agnostic (semantic) attributes. Through various probing experiments on five models, we demonstrate that the embeddings of these code language models include both language-specific (syntax) and language-agnostic (semantic) components.
    \item We demonstrate that removing these language-specific components and using only the language-agnostic component in downstream tasks can significantly enhance code retrieval tasks providing improvement upto +17 increase in MRR. Importantly, such improvements are achieved using inexpensive operations such as centering and projections, without using parallel language data or finetuning.

    \item Additionally, our extensive ablation studies suggest that as few as 100 samples per language suffice for these MRR improvements. We also confirm that the improvements are not restricted to a single type of embedding but can be realized across all common types, including mean, cls, and pooler embeddings.
\end{itemize}


\section{Language Agnostic Code Embeddings}

Let $\mathcal{M}$ represent a multilingual code language model trained on a set of programming languages $\{1, \dots, \ell\}$. Given a code snippet $c$ in a specific programming language $l$, this model produces an embedding $\textbf{e} \in \mathbb{R}^{d}$, denoted as $\mathcal{M}(c) = \textbf{e} \in \mathbb{R}^{d}$.  We hypothesize that the embedding $\textbf{e} \in \mathbb{R}^{d}$ of a code snippet can be decomposed into two components: a syntax component, $ \textbf{e}^{\text{s}} \in \mathbb{R}^{d}$, which depends on the programming language $l$, and semantic component, \(\textbf{e}^{\text{a}}\) which is language-agnostic. This relationship can be expressed as:


\begin{align} 
    \textbf{e} &= \textbf{e}^{\text{s}} + \textbf{e}^{\text{a}}    \label{eq:central_eq}
\end{align}

Next, we introduce the Estimation Set $\mathcal{E}$, which is used to estimate the language-specific components $\textbf{e}^{s}$.

\begin{definition}[Estimation Set]
    The Estimation set $\mathcal{E}$ is defined as a collection of $n$ code snippets $\{ c_{1}^{(l)}, \dots, c_{n}^{(l)}   \}$ from each programming language $ l \in \{1, \dots, \ell\}.$  Importantly, the code snippets in this set need not be direct translations of one another. 
\end{definition}

Now for given  model $\mathcal{M}$, we define embedding matrix $\textbf{E}_{l} \in \mathbb{R}^{n \times d}$ for each language $\ell$ as $\textbf{E}_{l} = \left[ \mathcal{M} (c_{1}^{(l)}), \dots, \mathcal{M}(c_{n}^{(l)}) \right].$ 

In the subsequent sections, we explore a variety of methods designed to remove language-specific information. This analysis is conducted from the unified perspective of Equation \ref{eq:central_eq}, which serves as the fundamental framework for disentangling language-specific and language-agnostic components within code embeddings. Additionally, we explicitly outline the assumptions that underpin each of these methods.

\begin{algorithm*}[t]
\tiny
\SetAlgoLined
\DontPrintSemicolon
\SetKwInput{KwRequire}{Require}
\KwIn{code embedding $\textbf{e} \in \mathbb{R}^{d}$, programming language $l \in \{1, \dots, \ell\}$ of embedding $\textbf{e}$ and  embedding matrices ${\textbf{E}_{1}, \dots, \textbf{E}_{\ell}}$, where $\textbf{E}_i \in \mathbb{R}^{n \times d}$, rank $r$ of the syntactic subspace in the case of LRD and CS-LRD. }
\KwOut{Language agnostic code embedding $\textbf{e}_{a} \in \mathbb{R}^{d}.$ }

\BlankLine
\SetKwFunction{FMain}{CS-LRD}
\SetKwFunction{FCentering}{centering}
\SetKwFunction{FLrd}{LRD}
\SetKwProg{Fn}{def}{}{}
\noindent
\begin{minipage}[t]{0.22\textwidth}
\Fn{\FCentering{$\textbf{\textup{M}}$}}{
    $\textbf{m}_{l} = \textbf{M}[:, l]$ \;
    \Return $\textbf{m}_{l} \in \mathbb{R}^{d}$ 
}
\end{minipage}%
\begin{minipage}[t]{0.30\textwidth}
\Fn{\FLrd{$\textbf{\textup{E}}$, $r$}}{
$\textbf{U}_r,\boldsymbol{\Sigma}_r, \textbf{V}_r = \textsc{topk-svd} (\textbf{M}, r)$ \;
\Return $\textbf{V}_{r} \in \mathbb{R}^{d \times r}$
}
\end{minipage}%
\begin{minipage}[t]{0.48\textwidth}
\Fn{\FMain{$\textbf{\textup{M}}$, $r$}}{
$\widehat{\textbf{m}}_{c} = \frac{1}{d} \textbf{M} \cdot \mathbb{1}_{\ell}.$ \;
$\widehat{\textbf{U}}_{r},\widehat{\boldsymbol{\Sigma}}_{r}, \widehat{\textbf{V}}_{r} = \textsc{topk-svd}(\textbf{M} - \widehat{\textbf{m}}_{c} \cdot \mathbb{1}^{T}_{\ell},r )$ \;
$\widehat{\textbf{M}}_{s} = \widehat{\textbf{U}}_{r}, \widehat{\boldsymbol{\Gamma}}_{s} = \widehat{\textbf{V}}^{T}_{r} \cdot \widehat{\boldsymbol{\Sigma}}_{r}.$ \;
$\widetilde{\textbf{M}}_{s} = \textsc{pseudo-inverse}(\widehat{\textbf{m}}_{c} \cdot \mathbb{1}^{T}_{\ell} +  \widehat{\textbf{M}}_{s} \cdot \widehat{\boldsymbol{\Gamma}}_{s}^{T} ).$ \; 
$\textbf{m}_{c} =  \widetilde{\textbf{M}}_{s} \cdot \mathbb{1}_{\ell}\textfractionsolidus \lVert \widetilde{\textbf{M}}_{s} \cdot \mathbb{1}_{\ell}\rVert_{2}^2 $ \;
$\textbf{U}_{r},\boldsymbol{\Sigma}_{r}, \textbf{V}_{r} = \textsc{topk-svd}(\widetilde{\textbf{M}}_{s} - \textbf{m}_{c} \cdot \mathbb{1}^{T},r ) ; $ \;
$\textbf{M}_{s} = \textbf{U}_{r}, \boldsymbol{\Gamma} = \textbf{V}^{T} \cdot \boldsymbol{\Sigma}_{r}.$  \;
\Return $\textbf{M}_{s} \in \mathbb{R}^{d \times r}$
}
\end{minipage}

$\textbf{M}  = [ \textsc{mean}(\textbf{E}_{1}), \dots, \textsc{mean}(\textbf{E}_{\ell}) ]  \in \mathbb{R}^{d \times \ell}$ \;
\eIf{\texttt{\textup{estimation-method}} == \texttt{\textup{centering}}}{
    $\textbf{e}^{s} = $ \FCentering{$\textbf{\textup{M}}.$} \;
}{
    \uElseIf{\texttt{\textup{estimation-method}} == \texttt{\textup{LRD}}}{
        $\textbf{P} =  \texttt{\textup{LRD}}(\textbf{E}_{l}, r).$ \;
    }
    \Else{
         $\textbf{P} = \texttt{\textup{CS-LRD}}(\textbf{M}, r).$ \;
    }
    $\textbf{e}^{s} = \textbf{e} - \textbf{P} \cdot \textbf{P}^{T} \cdot  \textbf{e}.$ \;
}

$\textbf{e}^{a} = \textbf{e} - \textbf{e}^{s}$

\Return $\textbf{e}^{a}$
\caption{Language Agnostic Code Embeddings}
\label{alg:lace}
\end{algorithm*}

\subsection{Centering}

The first method we explore is centering \cite{libovicky2020language}, which is grounded in the following key assumption:

\begin{Assumption} \label{assumption:centering}
      Given an programming language $l$, centering method makes an assumption that language specific components $\textbf{e}^{s}$ is same for all embeddings in that programming language $\ell$.
\end{Assumption}

Under Assumption \ref{assumption:centering}, the mean of the embeddings for a programming language $\ell$ can  be expressed as:

\begin{align*}
    \textbf{m}_{l} & = \frac{1}{n} \sum_{i=1}^{n} \textbf{e}_{i} = \frac{1}{n } \sum_{i=1}^n \left(\textbf{e}^{s}_{i} + \textbf{e}^{a}_{i} \right) \\
    &\stackrel{(\ref{eq:central_eq})}{=}  \frac{1}{n } \sum_{i=1}^n \left(\textbf{e}^{s} + \textbf{e}^{a}_{i} \right) = \textbf{e}^{s} + \underbrace{\frac{1}{n} \sum_{i=1}^{n} \textbf{e}_{i}^{a}}_{\textup{small for large } n} \approx \textbf{e}^{s}.
\end{align*}


From the above, it is evident that for a large enough value of $n$, the language-specific syntax embedding for a given programming language can be approximately estimated as the mean of the embeddings in that language. This method is summarized in Algorithm \ref{alg:lace} in \textsf{centering}.

\subsection{Low Rank Decomposition}
A significant concern with the centering method, as outlined in Assumption \ref{assumption:centering}, is its presumption that the syntax embedding for each code is the same and is independent of the given code content. Instead, it is only dependent on the programming language. To address this limitation, Low Rank Decomposition (LRD) \citep{schmidt1907theorie, yang2021simple} introduces distinct syntax subspaces for each programming language, operating under the following assumptions:

\begin{Assumption}
    The low rank decomposition method is based on the following assumptions:

    \begin{enumerate}

        \item The language-specific syntax embedding varies for each individual embedding.
         
        \item The language-specific syntax embedding, \(\textbf{e}^{\textup{s}}\), is orthogonal to the language-agnostic semantic embedding, \(\textbf{e}^{\textup{a}}\), i.e., \(\textbf{e}^{\textup{s}} \perp \textbf{e}^{\textup{a}}\).

        \item For each programming language, there exists a low-rank subspace of rank \(r\) that captures the syntactic essence of the embeddings.
    \end{enumerate}
\end{Assumption}

Based on the above assumptions, we determine the syntactic subspace of language $l$  of rank $r$, denoted as $\textbf{E}_{l}^{r} \in \mathbb{R}^{n \times d} $, as follows:

\begin{align}  \label{eq:lrd}
\begin{aligned} 
    \min_{ \textbf{E}_{l}^{r} \in \mathbb{R}^{n \times d}}  & \norm{\textbf{E}_{l} - \textbf{E}_{l}^{r} }_{\textup{F}}^2 \\
    &\text{\textup{s.t.}} \, \textsc{rank}(\textbf{E}_{l}^{r}) \leq r.
\end{aligned}
\end{align}

Equation \ref{eq:lrd} can be solved using \textsc{topk-svd} with 
$ k = r $
where $ \mathbf{E}_{l}^{r} = \mathbf{U}_{r} \mathbf{\Sigma}_{r} \mathbf{V}_{r}^{T}.$ The projection of the embedding $\mathbf{e}$ onto the \textsc{rowspace} $(\mathbf{E}_{l}^{r}) $is given by 
$ \mathbf{V}_{r} \mathbf{V}_{r}^{T} \mathbf{e}.$ The language-agnostic embedding is then obtained by removing this component:
$ \mathbf{e}^{a} = \mathbf{e} - \mathbf{V}_{r} \mathbf{V}_{r}^{T} \mathbf{e}.$ This method is summarized in Algorithm \ref{alg:lace} in \textsf{LRD}.



\subsection{Common Specific Low Rank Decomposition} 

The Common Specific Low Rank Decomposition \cite{piratla2020efficient, xie2022discovering} is a variant of low rank decomposition. Given different data domains, this method aims to learn both a common subspace shared across domains and a specific subspace unique to each domain.

\begin{Assumption}
The Common Specific Low Rank Decomposition is grounded on the following assumptions:
\begin{enumerate}
    \item The language-specific syntax embedding varies for each individual embedding.
    
    \item The language-specific syntax embedding, \(\textbf{e}^{\textup{s}}\), is orthogonal to the language-agnostic semantic embedding, \(\textbf{e}^{\textup{a}}\). In other words, \(\textbf{e}^{\textup{s}} \perp \textbf{e}^{\textup{a}}\).
    
    \item There exists a unified syntax subspace, consistent across all programming languages, that encapsulates the syntactic attributes of the code embedding.
\end{enumerate}    
\end{Assumption}

A key distinction is that while the syntax subspace in the traditional LRD is determined for each language individually, the CS-LRD method derives a singular, unified syntax subspace that encompasses all the considered programming languages.  It is formulated as:

\begin{align}  \label{eq:cs-lrd}
\begin{aligned} 
    \min_{\textbf{\textup{m}}_{c} \in \mathbb{R}^{d}, \textbf{\textup{M}}_{s} \in \mathbb{R}^{d \times r}, \boldsymbol{\Gamma}_{s} \in \mathbb{R}^{d \times \ell} }  & \norm{\textbf{\textup{M}} - \textbf{\textup{m}}_{c} \cdot \mathbb{1}^{T}_{\ell} - \textbf{\textup{M}}_{s}  \cdot \boldsymbol{\Gamma}_{s}^{T}   }_{\textup{F}}^2 \\
    &\text{\textup{s.t.}} \, \,\textbf{\textup{m}}_{c} \perp \textsc{colspan}(\textbf{\textup{M}}_{s}).
\end{aligned}
\end{align}%
where $\mathbf{M} = [\mathbf{m}_1, \dots, \mathbf{m}_{\ell}]$$,$ and $\mathbf{m}_1, \dots, \mathbf{m}_{\ell}$ are the mean embeddings of the $\{1, \dots, \ell\}$ programming languages. The matrix $\mathbf{M}_{s}$, which captures the common syntactic subspace, can be obtained by the \textsc{CS-LRD} function in Algorithm \ref{alg:lace}. Similar to LRD, the language-agnostic embedding is obtained by removing the projection of $\mathbf{e}$ on $\textbf{M}_{s}$, i.e., $ \mathbf{e}^{a} = \mathbf{e} - \mathbf{M}_{s} \mathbf{M}_{s}^{T} \mathbf{e}. $


\input{plots/lang_id/lang_id_classification}

\section{Experiments}

\noindent \textbf{Setup:}
We examine three tasks and analyze the performance before and after removing language-specific components:
(i) Probing - This task involves identifying languages using a linear classifier.
(ii) Code2Code search - Given a piece of code in language $L_1$, the objective is to retrieve the most semantically relevant code in another language $L_2$.
(iii) Text2Code search - The aim is to identify code that corresponds to a provided natural language query.

The first task assesses whether the procedures in Algorithm \ref{alg:lace} effectively eliminate language-specific (syntax) components. The second and third tasks determine if language-agnostic (semantic) components are preserved.

\noindent \textbf{Datasets:} We utilize programs from the Stack dataset \cite{kocetkov2022stack} to estimate language-specific components. For the Code2Code search, we employ XLCoST \cite{zhu2022xlcost}, and for the Text2Code search, we use CSN \cite{husain2019codesearchnet}.

\noindent \textbf{Models:} We consider five models, including encoder-only and encoder-decoder models: CodeBERT \cite{feng2020codebert}, GraphCodeBERT \cite{guo2020graphcodebert}, UnixCoder \cite{guo2022unixcoder}, StarEncoder \cite{li2023starcoder}, and CodeT5+ \cite{wang2023codet5+}.

\noindent \textbf{Embeddings:} For models like CodeBERT, GraphCodeBERT, UnixCoder, and StarEncoder, there isn't a standard method to obtain embeddings. In our retrieval tasks, we use mean embeddings, derived from the mean of the last hidden states. We conduct an ablation study to explore other embedding extraction methods in Section \ref{subsec:experiments_abalation_alignment_pooling}. For CodeT5+, only the pooler embedding is recomended and is given as output, and this is what we employ in our experiments.

\noindent \textbf{Retrieval Metrics:} For the Retrieval Task, we use Mean Reciprocal Rank (MRR) as our evaluation metric. MRR is calculated as $\textup{MRR} = \frac{1}{n} \sum_{i=1}^{n} \frac{1}{\text{rank}(c_i)} \times 100$, where $n$ represents the total number of queries, and $\text{rank}(c_i)$ denotes the rank of the correct answer for the $i$-th query in the retrieval results. Higher MRR values indicate better performance.

\subsection{Probing}

We evaluate the syntactic component of embeddings by employing a \emph{linear} classifier for the task of language identification, both pre and post transformations. From the Stack dataset \cite{kocetkov2022stack}, we allocate 10,000 code instances for estimating language components. For training, we use 24,000 code snippets from each language, and for validation, we utilize 6,000 codes from each respective language. The testing is performed on 10,000 codes for each language. The outcomes are depicted in Figure \ref{fig:language_identification}. Before transformation, the linear classifier yields high accuracy on the embeddings. However, after the removal of language-specific components, the accuracy declines sharply, experiencing a drop of at least 70\% across all models. In particular, for the CodeT5+ model, the accuracy approaches random performance. Moreover, in the context of CS-LRD, there's an interesting relationship between the rank \( r \) and performance. As \( r \) increases, the classifier's performance diminishes. It's worth noting that this behavior is not observed with the LRD.  

we also visualize PCA of CodeT5+ embeddings and show it in Figure \ref{fig:original_pca} which shows embeddings are clustered by language. But after removing language-specific components we see that in Figure \ref{fig:centering_pca} to Figure \ref{fig:cs_lrd_pca} there are no longer language clusters.

\subsection{Code2Code Search}
\label{subsec:experiments_code2codesearch}

Given a query $Q$ in a source language $\mathcal{S}$, the objective is to extract a code snippet with semantic similarity from a specific database. Depending on the language composition of the database, we consider three different variations:

\begin{itemize}
\item \textbf{Monolingual Database:} In this conventional setting, the database consists entirely of programs written in a single target language $\mathcal{T}$, which is distinct from the source language $\mathcal{S}$. 
\item \textbf{Source-Excluded Multilingual Database:} In this variation, the database is composed of programs in multiple languages $\mathcal{T}_{1}, \dots, \mathcal{T}_{n}$, where $\mathcal{T}_{i}$ differs from the source language $\mathcal{S}$.  
\item \textbf{Source-Included Multilingual Database:} This final variation includes the source language $\mathcal{S}$ within its spectrum of target languages. We evaluate the language bias of models \cite{yang2021simple}, wherein a code from the source language $\mathcal{S}$ is ranked higher than codes that are more semantically similar but from different languages.
\end{itemize}

For this task, we use the XLCoST dataset \cite{zhu2022xlcost}, which contains parallel translations of seven programming languages: C, C\#, C++, Java, JavaScript, PHP, and Python. However, it's important to note that CodeBERT and GraphCodeBERT do not support C, C++, and C\#. Therefore, we only consider Java, JavaScript, PHP, and Python for these models, while all seven languages are included for all other models.

\input{plots/c2c/c2c_zeroshot}
\input{plots/c2c/c2c_zeroshot_r_value}
\begin{table*}[t!]
\centering
\resizebox{0.8\textwidth}{!}{%
\input{tables/c2c/zeroshot/c2c_codet5+_pooler}
}

\vspace{0.4cm}\caption{MRR averaged across all target languages for zero-shot Code2Code search using CodeT5+ \cite{wang2023codet5+}.}
\label{tab:c2c_zeroshot_codet5p}
\end{table*}
We present the change in the Mean Reciprocal Rank (MRR) before and after the removal of the language component in Figure \ref{fig:c2c_zeroshot}.  This change is averaged over all pairwise language retrieval tasks, amounting to \(6 \times 7 = 42\) tasks in total. Additionally, for the CodeT5+ model, we offer a detailed breakdown of the MRR for each source language. The retrieval results are averaged across the six target languages and are tabulated in Table \ref{tab:c2c_zeroshot_codet5p}.


\noindent \textbf{Discussion:}  Significant improvements are observed before and after removing the language component, with an absolute increase in MRR ranging up to +17. We discuss a couple of factors below.

\begin{enumerate}
\item \textbf{Database Configuration:} Models exhibit substantial language bias, leading to a drastic drop in performance in the 'Source Included Multilingual' setup, with a reduction of -59.62\% from 89.51 to 29.89.

\item \textbf{Centering Effects:} In three out of five cases, centering has a detrimental impact on performance. This aligns with the notion that centering may mix syntax and semantic signals, potentially removing semantic meaning as well \cite{yang2021simple, xie2022discovering}.

\item \textbf{UniXcoder Exception:} Notably, UniXcoder explicitly aligns representations from different programming languages during pretraining itself using a task involving cross-modal generation \cite{guo2022unixcoder}. Consequently, none of the methods provide any improvement in this case.  

\item \textbf{CS-LRD Superiority:} In most cases, CS-LRD outperforms both centering and LRD. This is attributed to the joint learning of the syntax subspace across different programming languages in CS-LRD.

\item \textbf{Effect of Rank in LRD and CS-LRD:} We examine the impact of the rank of the subspace $r$ in Figure \ref{fig:c2c_zero_shot_r_value} for both LRD (Top Row) and CS-LRD (Bottom Row). Increasing $r$ consistently enhances MRR in CS-LRD, while no such behavior is observed in LRD, which is less stable compared to CS-LRD.

\end{enumerate}

\input{plots/t2c/t2c_zeroshot}
\input{plots/t2c/t2c_zeroshot_r_value}
\begin{table*}[t!]
\centering
\resizebox{0.7\textwidth}{!}{%
\input{tables/t2c/zeroshot/t2c_unixcoder_mean}
}
\vspace{0.4cm}\caption{MRR for zero-shot Text2Code search using Unixcoder \cite{guo2022unixcoder} .}
\label{tab:t2c_zeroshot_unixcoder}
\end{table*}


\input{plots/c2c_zeroshot_abalation_alignment}
\input{plots/c2c_zeroshot_abalation_pooling}

\subsection{Text2Code Search}
\label{subsec:experiments_text2codesearch}

In this section, we delve into Text2Code search, a task where the objective is to find code that corresponds to a given natural language query (in English).  We explore two distinct settings for Text2Code search:

\begin{itemize}
    \item \textbf{Monolingual Database:} In this setting, we construct the retrieval database using data from a single programming language.
    \item \textbf{Multilingual Database:} In contrast, for this setting, we include data from all programming languages in the retrieval database. The goal is to locate the correct code snippet that matches the query, regardless of the programming language.
\end{itemize}

For this task, we utilize the CodeSearchNet dataset \cite{husain2019codesearchnet}, which contains data in six programming languages: Go, Ruby, Java, JavaScript, PHP, and Python. Retrieval database consists of codes in both val and test. 

We present the change in Mean Reciprocal Rank (MRR) before and after removing the language component in Figure \ref{fig:t2c_zeroshot}. Full view for Unixcoder can be found at  Table \ref{tab:t2c_zeroshot_unixcoder}. 

\noindent \textbf{Discussion:} Sizable improvements are observed before and after removing language component, with an absolute increase in MRR ranging upto +8.  We discuss a couple of factors below.


\begin{enumerate}

\item \textbf{CodeT5+ Exception:} CodeT5+ includes contrastive tuning as one of its pretraining tasks \cite{wang2023codet5+} for text-to-code, which explicitly aligns English with programming languages. Hence, we don't observe any improvement.

\item \textbf{Centering Superiority:} Unlike in Code2Code search, in Text2Code search centering outperforms both LRD and CS-LRD.

\item \textbf{Effect of Rank in LRD and CS-LRD:} We study the influence of the subspace rank r as depicted in Figure \ref{fig:t2c_zero_shot_r_value}. The top row illustrates the effect for LRD, while the bottom row represents CS-LRD. For both CS-LRD and LRD, increasing r either consistently improves MRR or remains stable. However, for CodeT5+, there is a consistent decrease.

\item \textbf{Effect of Projecting out English:} We conduct retrieval in two distinct ways. In the first method, we remove language components solely from programming languages, leaving the query unaffected (no English component is removed). In the second method, we transform the query by removing the English language component from it. The results are depicted in Figure \ref{fig:t2c_zero_shot_r_value}. We find that projecting out the English language components is crucial to observe an increase in MRR.

 \end{enumerate}

\subsection{Ablation Study}
\label{subsec:experiments_abalation}


In this section, we conduct various ablation studies focusing on the estimation set's size and the effects of different kinds of embeddings.

\subsubsection{Effect of Estimation Set Size}
\label{subsec:experiments_abalation_alignment_size}

In this section, we investigate the impact of estimation set size on language estimation and its influence on Mean Reciprocal Rank (MRR) in zero-shot Code2Code search. We randomly sample $\{100, 500, 1000, 5000, 10000, 25000\}$ examples from the original pool of 50,000 samples from the Stack dataset for each language used in Section \ref{subsec:experiments_code2codesearch}. Subsequently, we conduct retrievals with the language components removed based on these samples. This study is repeated five times for each sample size, and we calculate the MRR change. The results can be seen in Figure \ref{fig:c2c_zeroshot_abalation_alignment}. In this figure, the top row represents Centering, the middle row showcases LRD, and the bottom row depicts CS-LRD.

The results highlight a significant change in MRR, even with estimation sets containing as few as 100 samples per language. However, some variance is observed in certain instances. This variance diminishes considerably once the estimation set expands to 1000 samples, resulting in a steadier MRR shift. Interestingly, the variance is typically greater for Centering and LRD compared to CS-LRD. This study also reveals that specific examples in the estimation set don't play as significant a role as the overall size of the estimation set.

\subsubsection{Mean embedding vs [CLS] embedding vs Pooler output}

\label{subsec:experiments_abalation_alignment_pooling}

In this section, we examine various kinds of embeddings and analyze the effects of removing language components from them for zero-shot code2code search. As noted in Sections \ref{subsec:experiments_code2codesearch} and \ref{subsec:experiments_text2codesearch}, we utilized mean embeddings for CodeBERT, GraphCodeBERT, UnixCoder, and StarEncoder. However, other embedding types are also commonly employed in practice.

To clarify, let \( c \) be a code snippet. The function \(\textsf{encoder}(c)\) produces the last hidden state with the shape \( \mathbb{R}^{t \times d} \), where \( t \) denotes the number of tokens in the code. There are several methods to obtain a single \( \mathbb{R}^{d} \) representation from the encoder's output. These methods are defined as follows:

\begin{align*}
    \text{mean-embedding}(c) &\triangleq \textsf{encoder}(c).\text{mean(0)}\\
    \text{cls-embedding}(c)  &\triangleq \textsf{encoder}(c)[0]\\
    \text{pooler-embedding}(c) &\triangleq \textsf{pooler}(\textsf{encoder}(x)[0])
\end{align*}

Here, \texttt{pooler} is an MLP layer positioned atop the encoder, and its output is directed to the language modeling head.

Results are displayed in Figure \ref{fig:c2c_zeroshot_abalation_pooling}. We observe that the improvements aren't limited to mean-embedding; they also extend to cls-embedding and pooler-embedding. Specifically, when using CS-LRD with CodeBERT's cls-embedding, there's a significant increase of +26.22. Similarly, StarEncoder's pooler embedding sees a +14.87 improvement with CS-LRD.  Notably, mean-embedding remains superior to other embedding variants regardless of the presence or absence of language information.

\section{Related Work}

\noindent \textbf{Cross Lingual properties of Natural Language models}: We are the first to investigate the cross-lingual properties of pretrained multilingual code language models, examining them through the lens of both language-specific (syntax) and language-agnostic (semantic) attributes. Our research is motivated by a rich body of work that probes similar behavior in multilingual natural language models \cite{schuster2019cross, libovicky2020language, kulshreshtha2020cross, yang2021simple, xie2022discovering, chang2022geometry}. While these studies predominantly concentrate on models trained for natural languages, our emphasis lies on those designed for programming languages.

\noindent \textbf{Code Representation Learning}:  The monumental success of BERT \cite{devlin-etal-2019-bert} and T5 \cite{raffel2020exploring} in natural language understanding has sparked significant interest in adapting similar architectures for programming languages. This interest has given rise to models like CodeBERT \cite{feng2020codebert}, CodeTransformer \cite{zugner2020language}, GraphCodeBERT \cite{guo2020graphcodebert}, ContraCode \cite{jain-etal-2021-contrastive}, SynCoBERT \cite{wang2021syncobert}, UniXCoder \cite{guo2022unixcoder}, and PLBART \cite{ahmad2021unified}. Some of these works \cite{zugner2020language, guo2020graphcodebert} explore code-specific pretraining tasks, utilizing both language-specific features (e.g., Program Analysis Edges) and language-agnostic features (e.g., Abstract Syntax Trees) to improve the performance of multilingual code models. In contrast, our work focuses on examining the representations of pretrained multilingual code language models.

\noindent \textbf{LMs for Code Generation}: In recent years, there have been many language models (LMs) for code trained with various architectures, sizes, and data mixtures, inspired by the huge success of GPT \cite{radford2019language, brown2020language}. Some of these include Codex \cite{chen2021evaluating}, CodeGeeX \cite{zheng2023codegeex}, SantaCoder \cite{allal2023santacoder}, PolyCoder \cite{xu2022systematic}, CodeGen \cite{nijkamp2022codegen}, StarCoder \cite{li2023starcoder}, WizardCoder \cite{luo2023wizardcoder}, and Code Llama \cite{roziere2023code}. In this work, instead of focusing on code generation, we concentrate on code representations.

\section{Discussion}

In our study of multilingual code models, we find that these embeddings can be decomposed into two main components: language-specific and language-agnostic. Through extensive experimentation, we conclude that when representations are not aligned during pre-training, the removal of the language-specific component, utilizing only the language-agnostic component, significantly enhances performance in retrieval tasks.

\section*{Acknowledgements}
A. Gu is supported by the National Science Foundation (NSF) Graduate Research Fellowship under Grant No. 2141064. We sincerely thank Armando Solar-Lezama for comments and feedback on the project.

\bibliography{custom}

\appendix
\input{plots/c2c/c2c_finetuned}
\input{plots/t2c/t2c_finetuned}

\section{Code2Code search results}

We provide more detailed results for the Code2Code search, where we calculate the mean over the target language and present the MRR (Mean Reciprocal Rank) for each source language. This is similar to what is shown in Table \ref{tab:c2c_zeroshot_codet5p} for CodeT5+. For CodeBERT, the details can be found in Table \ref{tab:c2c_zeroshot_codebert}. For GraphCodeBERT, see Table \ref{tab:c2c_zeroshot_graphcodebert}. Results for UnixCoder are in Table \ref{tab:c2c_zeroshot_unixcoder}, and for StarEncoder, refer to Table \ref{tab:c2c_zeroshot_starencoder}.

\section{Text2Code search results}

We provide more detailed results for the Text2Code search, similar to the information shown in Table \ref{tab:t2c_zeroshot_unixcoder} for UnixCoder. In Table \ref{tab:t2c_zeroshot}, we display results for all four models: CodeBERT, GraphCodeBERT, StarEncoder, and CodeT5+.

\begin{table*}[t!]
\centering
\resizebox{\textwidth}{!}{%
\input{tables/c2c/zeroshot/c2c_codebert_mean}
\input{tables/c2c/zeroshot/c2c_codebert_cls}
}
\vspace{0.4cm}
\resizebox{0.5\textwidth}{!}{%
\input{tables/c2c/zeroshot/c2c_codebert_pooler}
}
\vspace{0.4cm}\caption{Mean Reciprocal Rank (MRR) averaged across all target languages for zero-shot Code2Code search using CodeBERT \cite{feng2020codebert}.}
\label{tab:c2c_zeroshot_codebert}
\end{table*}

\begin{table*}[]
\centering
\resizebox{\textwidth}{!}{%
\input{tables/c2c/zeroshot/c2c_graphcodebert_mean}
\input{tables/c2c/zeroshot/c2c_graphcodebert_cls}
}
\vspace{0.4cm}
\resizebox{0.5\textwidth}{!}{%
\input{tables/c2c/zeroshot/c2c_graphcodebert_pooler}
}
\vspace{0.4cm}\caption{Mean Reciprocal Rank (MRR) averaged across all target languages for zero-shot Code2Code search using GraphCodeBERT \cite{guo2020graphcodebert}.}
\label{tab:c2c_zeroshot_graphcodebert}
\end{table*}

\begin{table*}[t!]
\centering
\resizebox{\textwidth}{!}{%
\input{tables/c2c/zeroshot/c2c_starencoder_mean}
\input{tables/c2c/zeroshot/c2c_starencoder_cls}
}
\vspace{0.4cm}
\resizebox{0.5\textwidth}{!}{%
\input{tables/c2c/zeroshot/c2c_starencoder_pooler}
}\vspace{0.4cm}\caption{Mean Reciprocal Rank (MRR) averaged across all target languages for zero-shot Code2Code search using StarEncoder \cite{li2023starcoder}.}
\label{tab:c2c_zeroshot_starencoder}
\end{table*}

\begin{table*}[t!]
\centering
\resizebox{\textwidth}{!}{%
\input{tables/c2c/zeroshot/c2c_unixcoder_mean}
\input{tables/c2c/zeroshot/c2c_unixcoder_cls}
}\vspace{0.4cm}
\resizebox{0.5\textwidth}{!}{%
\input{tables/c2c/zeroshot/c2c_unixcoder_pooler}
}\vspace{0.4cm}\caption{Mean Reciprocal Rank (MRR) averaged across all target languages for zero-shot Code2Code search using UnixCoder \cite{guo2022unixcoder}.}
\label{tab:c2c_zeroshot_unixcoder}
\end{table*}

\begin{table*}[t!]
\centering
\resizebox{\textwidth}{!}{%
\input{tables/t2c/zeroshot/t2c_codebert_mean}
\input{tables/t2c/zeroshot/t2c_graphcodebert_mean}
}
\vspace{0.4cm}
\resizebox{\textwidth}{!}{%
\input{tables/t2c/zeroshot/t2c_starencoder_mean}
\input{tables/t2c/zeroshot/t2c_codet5+_pooler}
}
\caption{Mean Reciprocal Rank (MRR) for zero-shot Text2Code search using CodeBERT \cite{feng2020codebert}, GraphCodeBERT \cite{guo2020graphcodebert}, StarEncoder \cite{li2023starcoder}, CodeT5+ \cite{wang2023codet5+}. }
\label{tab:t2c_zeroshot}
\end{table*}

\section{Effect on Contrastive Finetuned models}

In this section, we fine-tune the models using contrastive loss \cite{oord2018representation} for three epochs, with a batch size of eight, employing the AdamW optimizer with a linear scheduler and 500 warm-up steps. For both Code2Code search and Text2Code search, we ensure that each batch includes translation pairs from multiple languages through random sampling. This form of multilingual contrastive learning encourages representations to be aligned across programming languages. The results for Code2Code search can be viewed in Figure \ref{fig:c2c_finetuned}, and for Text2Code search in Figure \ref{fig:t2c_finetuned}. Similar to what we saw in Section \ref{subsec:experiments_code2codesearch} and Section \ref{subsec:experiments_text2codesearch}, there is no significant benefit in removing language components when representations are already aligned.



\end{document}

%% file: plots/lang_id/lang_id_classification.tex
\begin{figure*}[ht!]
  \centering
  \begin{subfigure}[b]{0.19\linewidth}
    \centering
    \includegraphics[width=\linewidth]{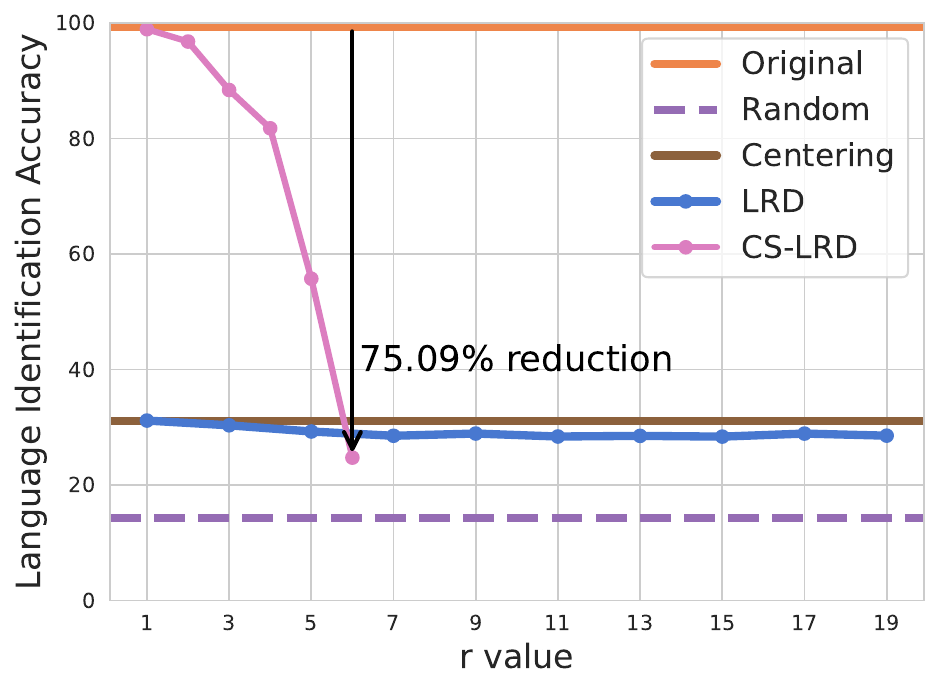}
    \subcaption{CodeBERT}
    \label{fig:lang_id_classification_codebert}
  \end{subfigure}
  \begin{subfigure}[b]{0.19\linewidth}
    \centering
    \includegraphics[width=\linewidth]{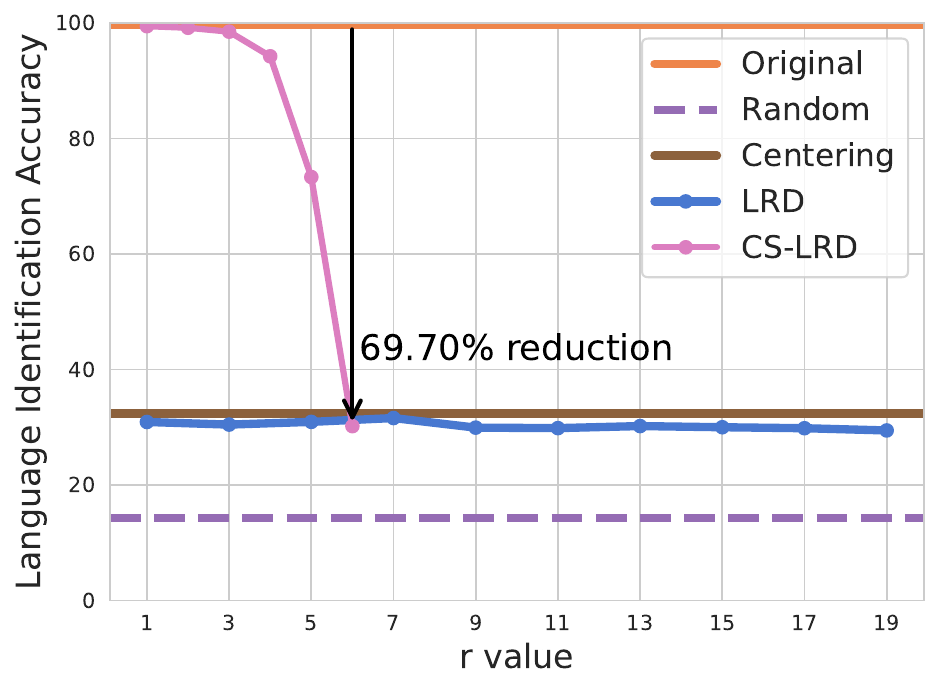}
    \subcaption{GraphCodeBERT}
    \label{fig:lang_id_classification_graphcodebert}
  \end{subfigure}
  \begin{subfigure}[b]{0.19\linewidth}
    \centering
    \includegraphics[width=\linewidth]{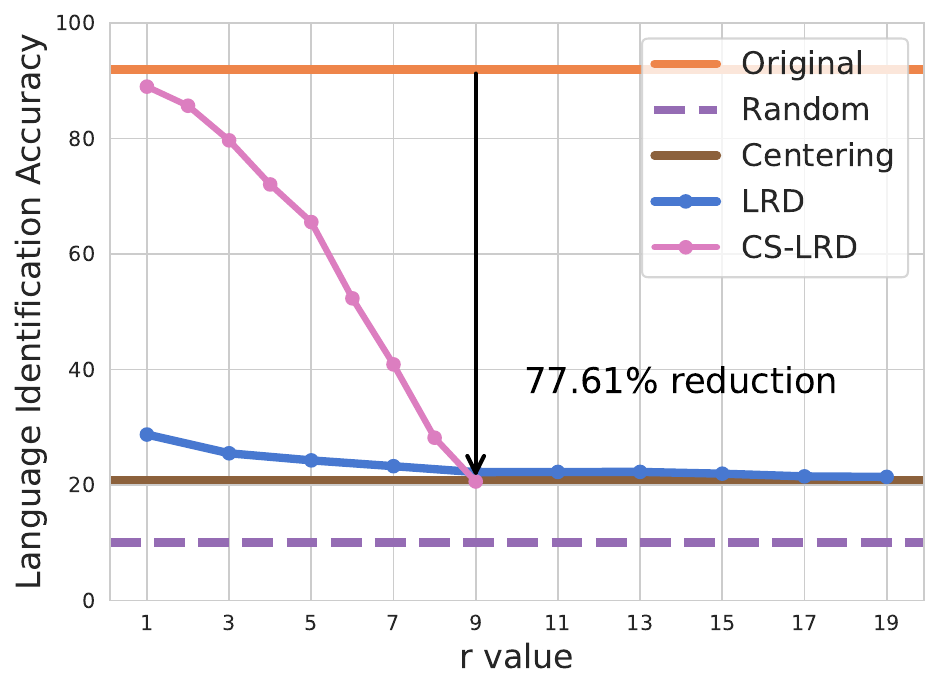}
    \subcaption{UnixCoder}
    \label{fig:lang_id_classification_unixcoder}
  \end{subfigure}
  \begin{subfigure}[b]{0.19\linewidth}
    \centering
    \includegraphics[width=\linewidth]{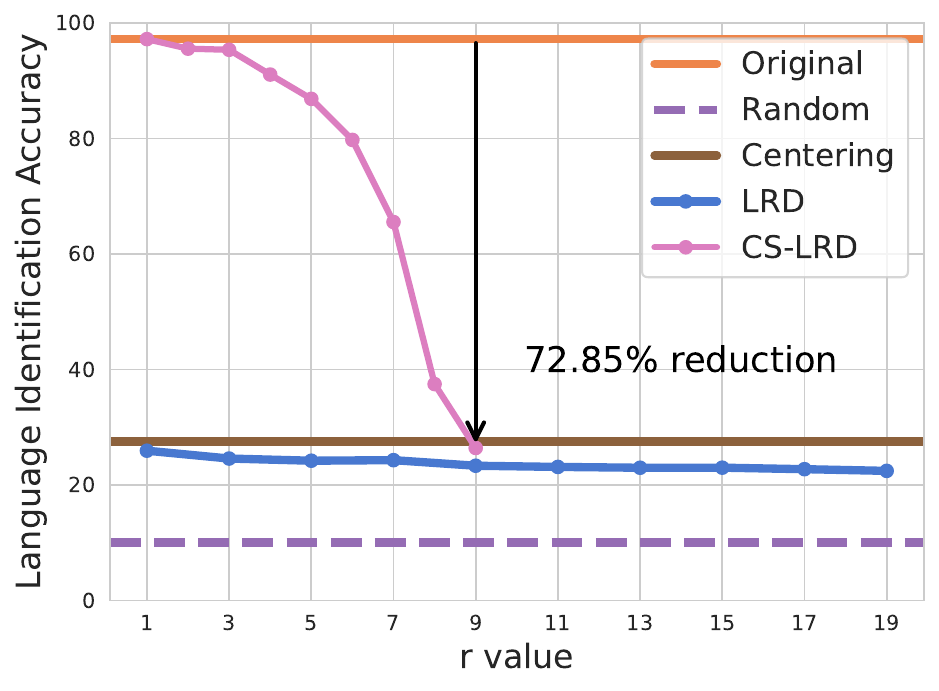}
    \subcaption{StarEncoder}
    \label{fig:lang_id_classification_starencoder}
  \end{subfigure}
  \begin{subfigure}[b]{0.19\linewidth}
    \centering
    \includegraphics[width=\linewidth]{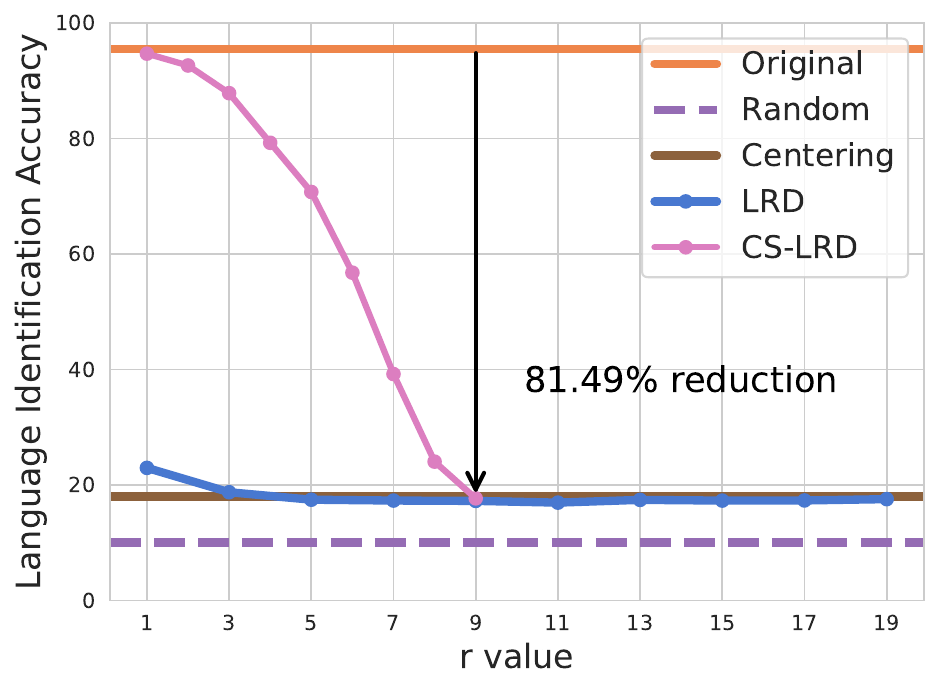}
    \subcaption{CodeT5+}
    \label{fig:lang_id_classification_codet5+}
  \end{subfigure} \\[0.16in]  
  \begin{subfigure}[b]{0.24\linewidth}
    \centering
    \includegraphics[width=\linewidth]{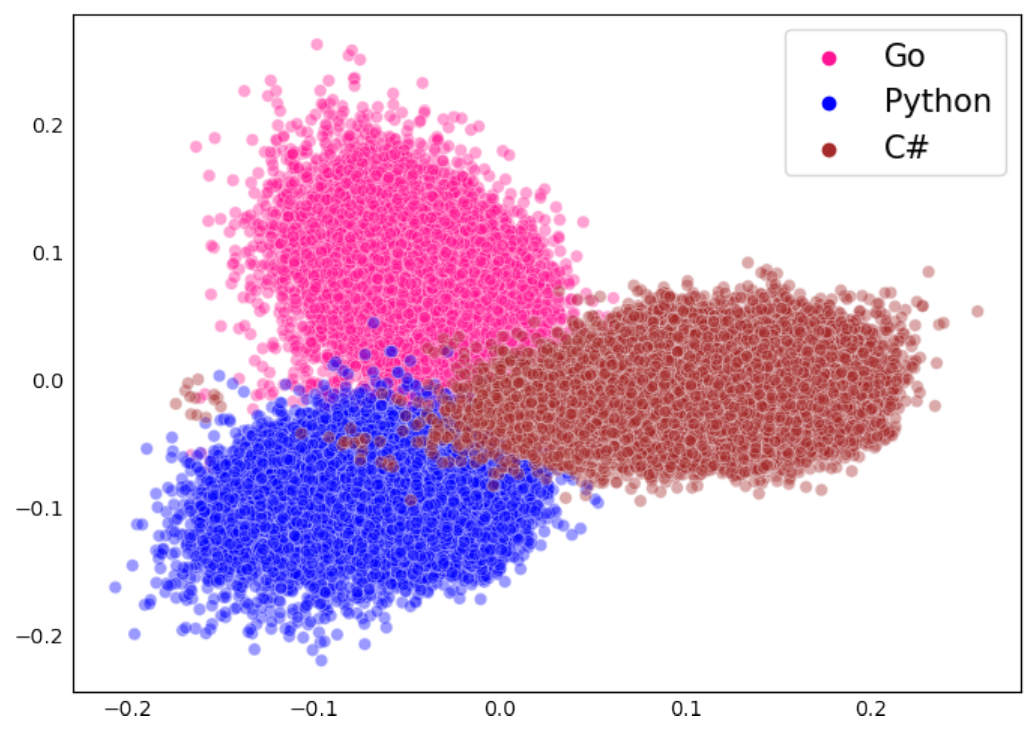}
    \subcaption{Original}
    \label{fig:original_pca}
  \end{subfigure}
  \begin{subfigure}[b]{0.24\linewidth}
    \centering
    \includegraphics[width=\linewidth]{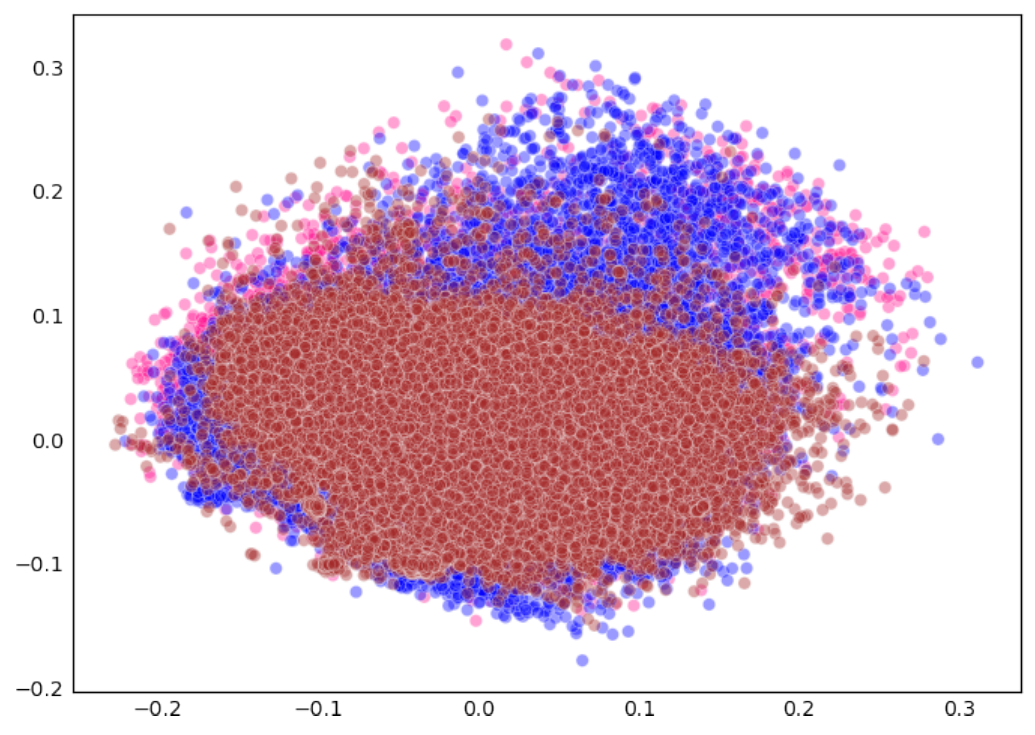}
    \subcaption{Centering}
    \label{fig:centering_pca}
  \end{subfigure}
  \begin{subfigure}[b]{0.24\linewidth}
    \centering
    \includegraphics[width=\linewidth]{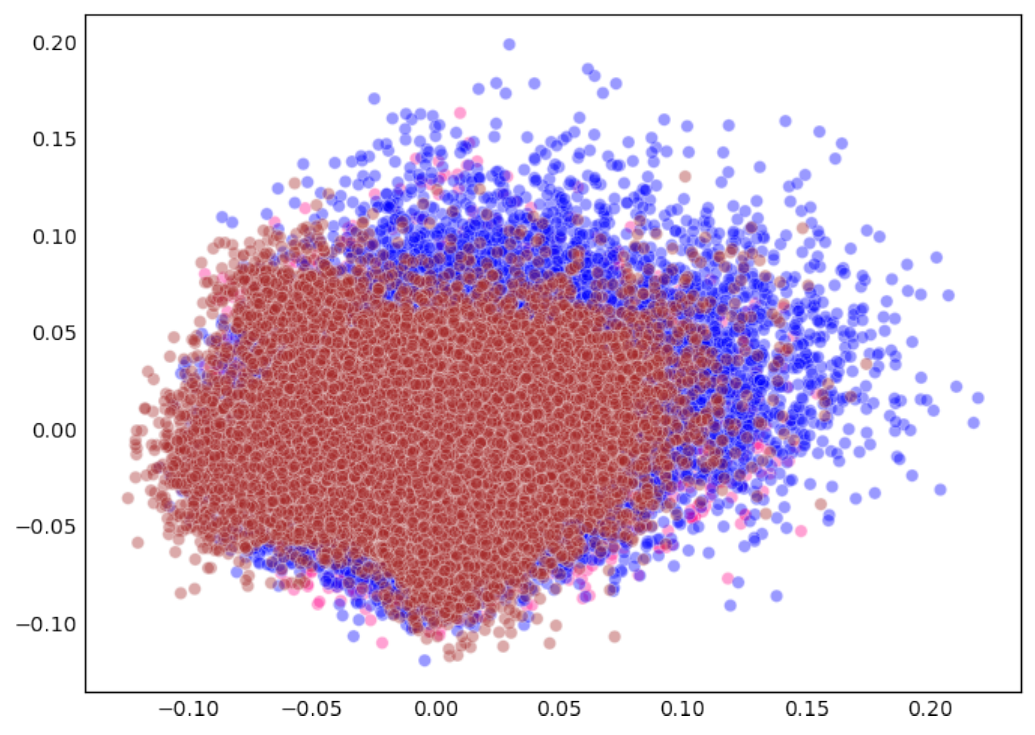}
    \subcaption{LRD}
    \label{fig:lrd_pca}
  \end{subfigure}
  \begin{subfigure}[b]{0.24\linewidth}
    \centering
    \includegraphics[width=\linewidth]{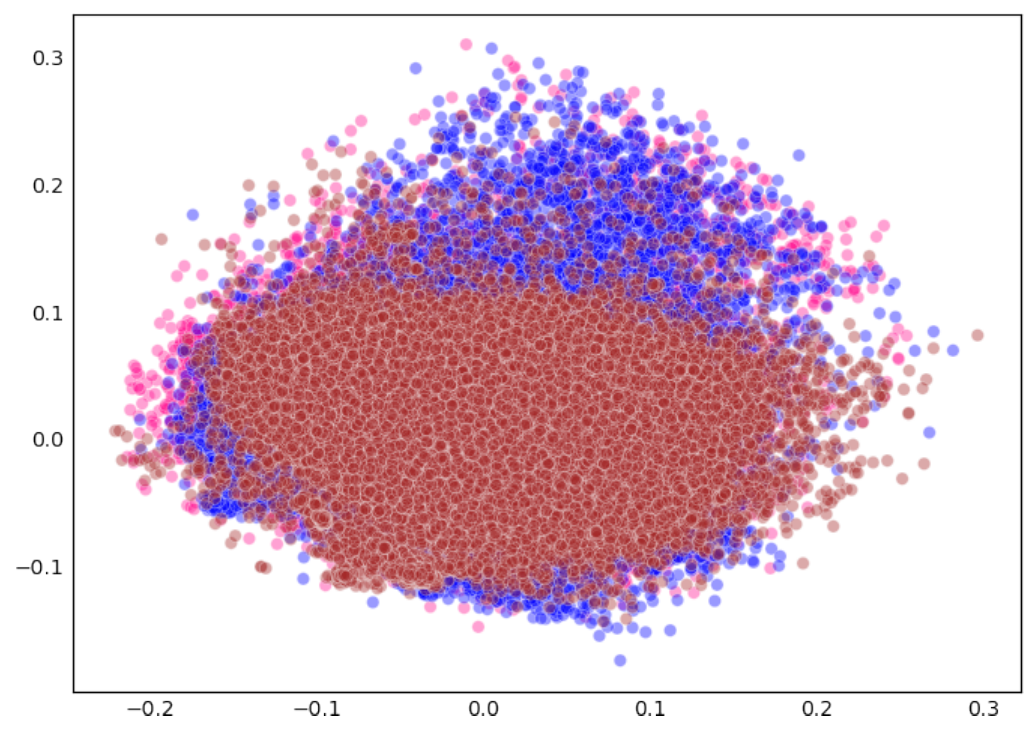}
    \subcaption{CS-LRD}
    \label{fig:cs_lrd_pca}
  \end{subfigure}  
\caption{The top row illustrates the impact on language identification accuracy before and after removing language-specific components. Meanwhile, the bottom row displays the PCA of Code T5+ embeddings for the languages: Go, Python, and C\#.}
  \label{fig:language_identification}
\end{figure*}

%% file: plots/c2c/c2c_zeroshot.tex
\begin{figure*}[htbp!]
  \centering

  \begin{subfigure}[b]{0.19\linewidth}
    \centering
    \includegraphics[width=\linewidth]{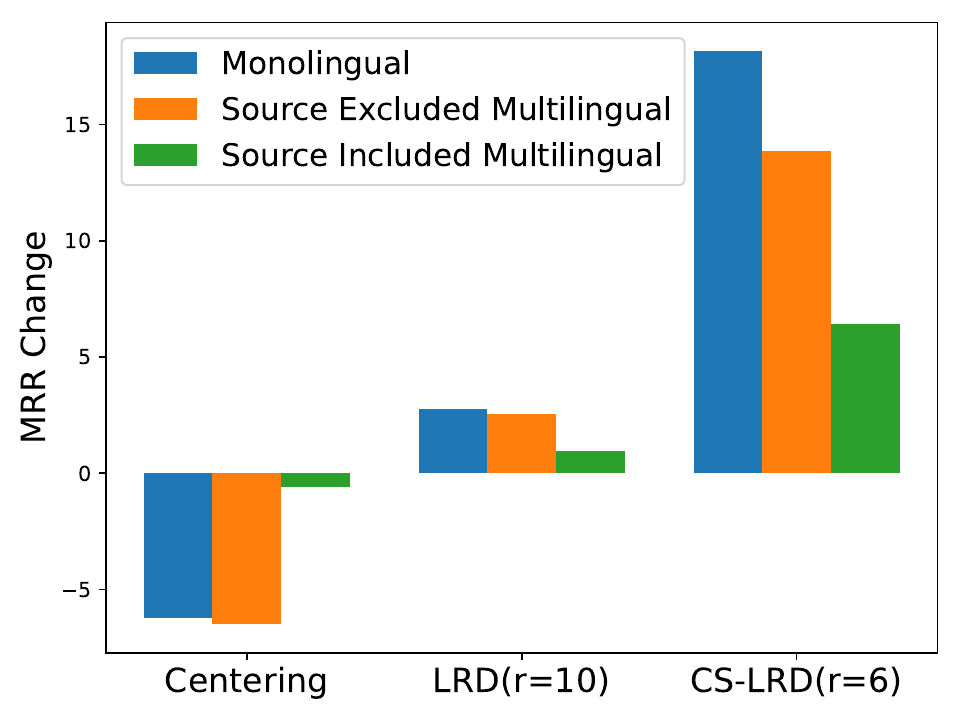}
    \subcaption{CodeBERT}
    \label{fig:c2c_zeroshot_codebert_mean}
  \end{subfigure}
  \begin{subfigure}[b]{0.19\linewidth}
    \centering
    \includegraphics[width=\linewidth]{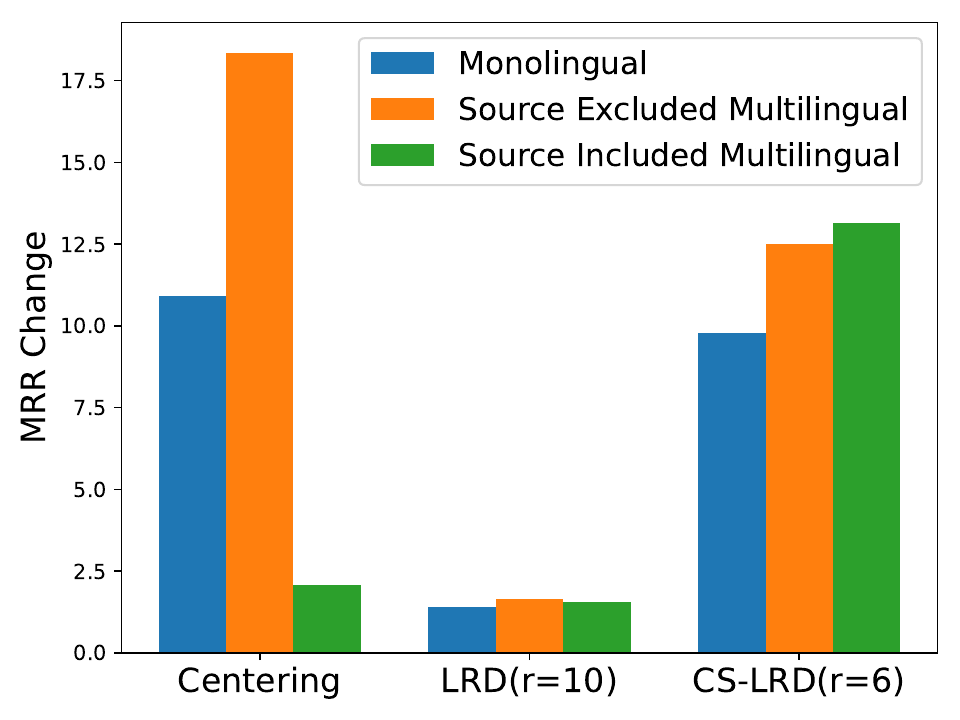}
    \subcaption{GraphCodeBERT}
    \label{fig:c2c_zeroshot_graphcodebert_mean}
  \end{subfigure}
  \begin{subfigure}[b]{0.19\linewidth}
    \centering
    \includegraphics[width=\linewidth]{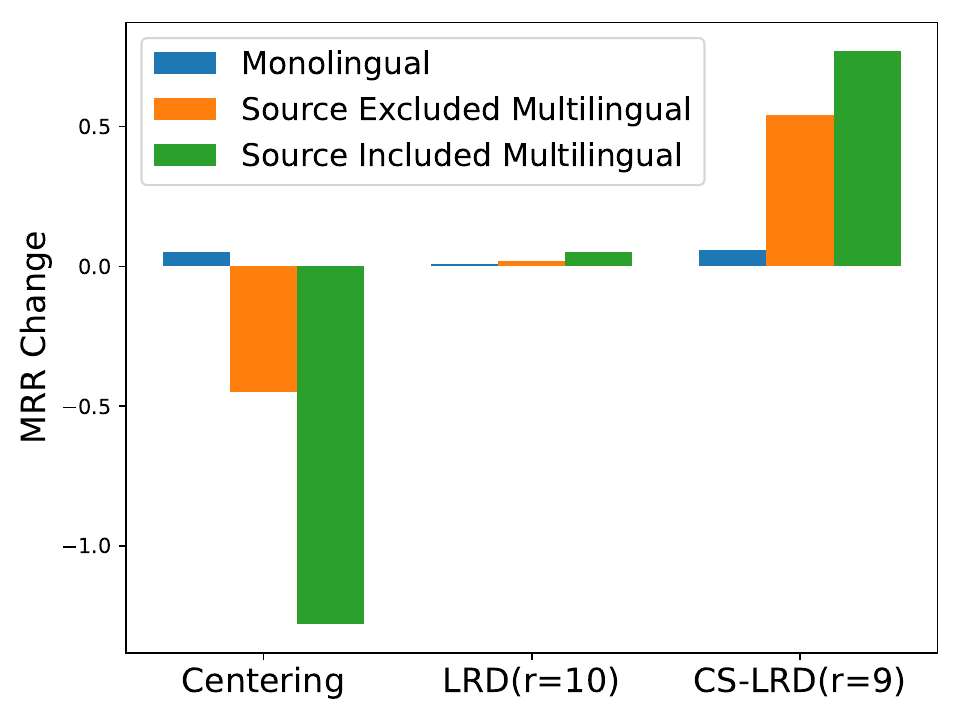}
    \subcaption{UnixCoder}
    \label{fig:c2c_zeroshot_unixcoder_mean}
  \end{subfigure}
  \begin{subfigure}[b]{0.19\linewidth}
    \centering
    \includegraphics[width=\linewidth]{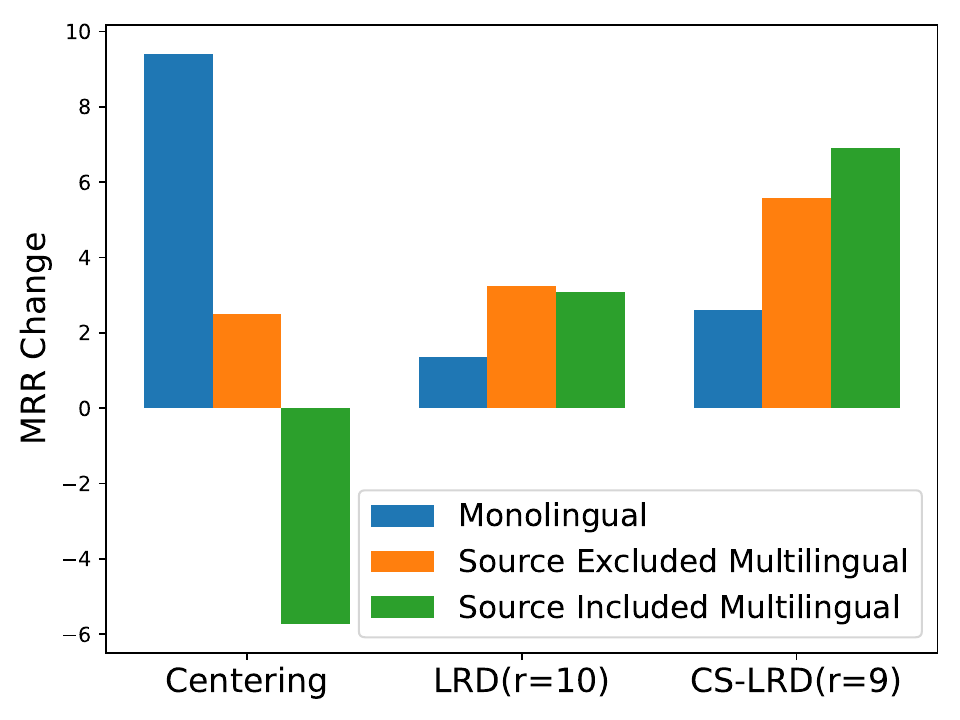}
    \subcaption{StarEncoder}
    \label{fig:c2c_zeroshot_starencoder_mean}
  \end{subfigure}
  \begin{subfigure}[b]{0.19\linewidth}
    \centering
    \includegraphics[width=\linewidth]{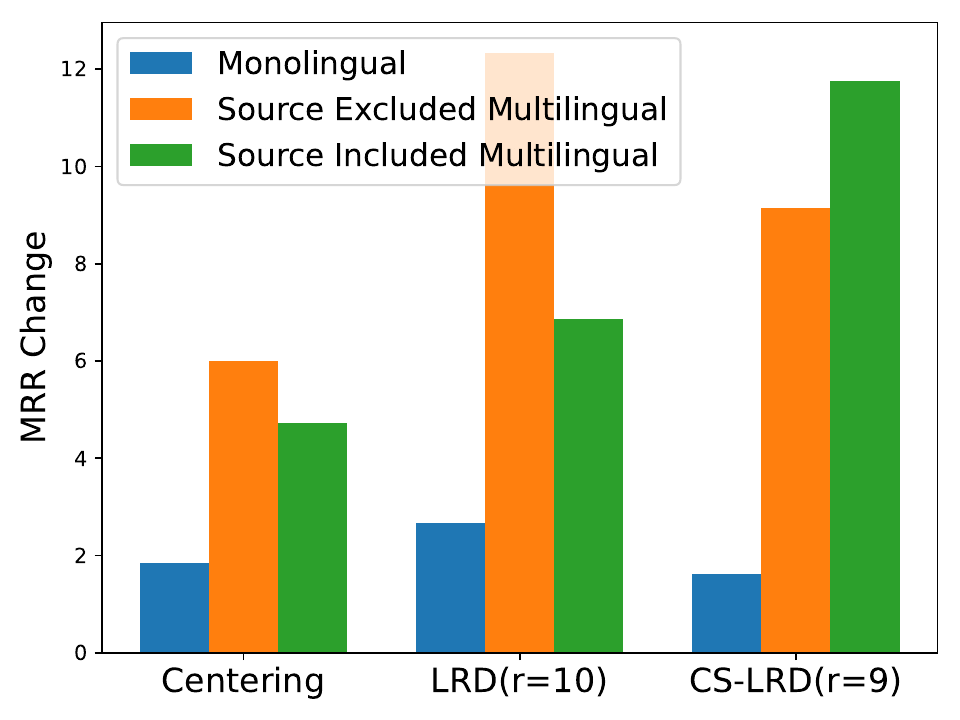}
    \subcaption{CodeT5+}
    \label{fig:c2c_zeroshot_codet5+_pooler}
  \end{subfigure}
  \caption{Absolute change in MRR after removing language components in zero-shot Code2Code search.}
  \label{fig:c2c_zeroshot}
\end{figure*}

%% file: plots/c2c/c2c_zeroshot_r_value.tex
\begin{figure*}[htb!]
  \centering

  \begin{subfigure}[b]{0.19\linewidth}
    \centering
    \includegraphics[width=\linewidth]{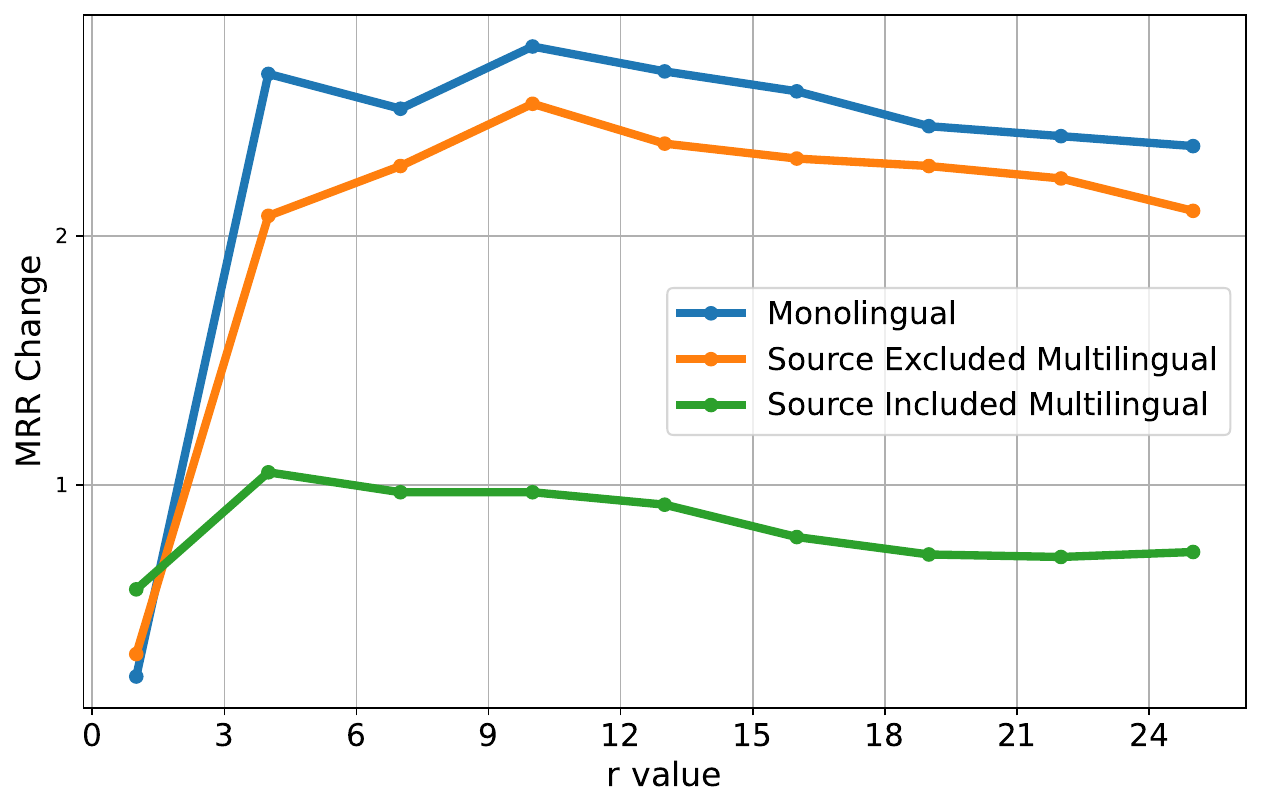}
    \subcaption{CodeBERT}
    \label{fig:c2c_zero_shot_lrd_plot_codebert_mean}
  \end{subfigure}
  \begin{subfigure}[b]{0.19\linewidth}
    \centering
    \includegraphics[width=\linewidth]{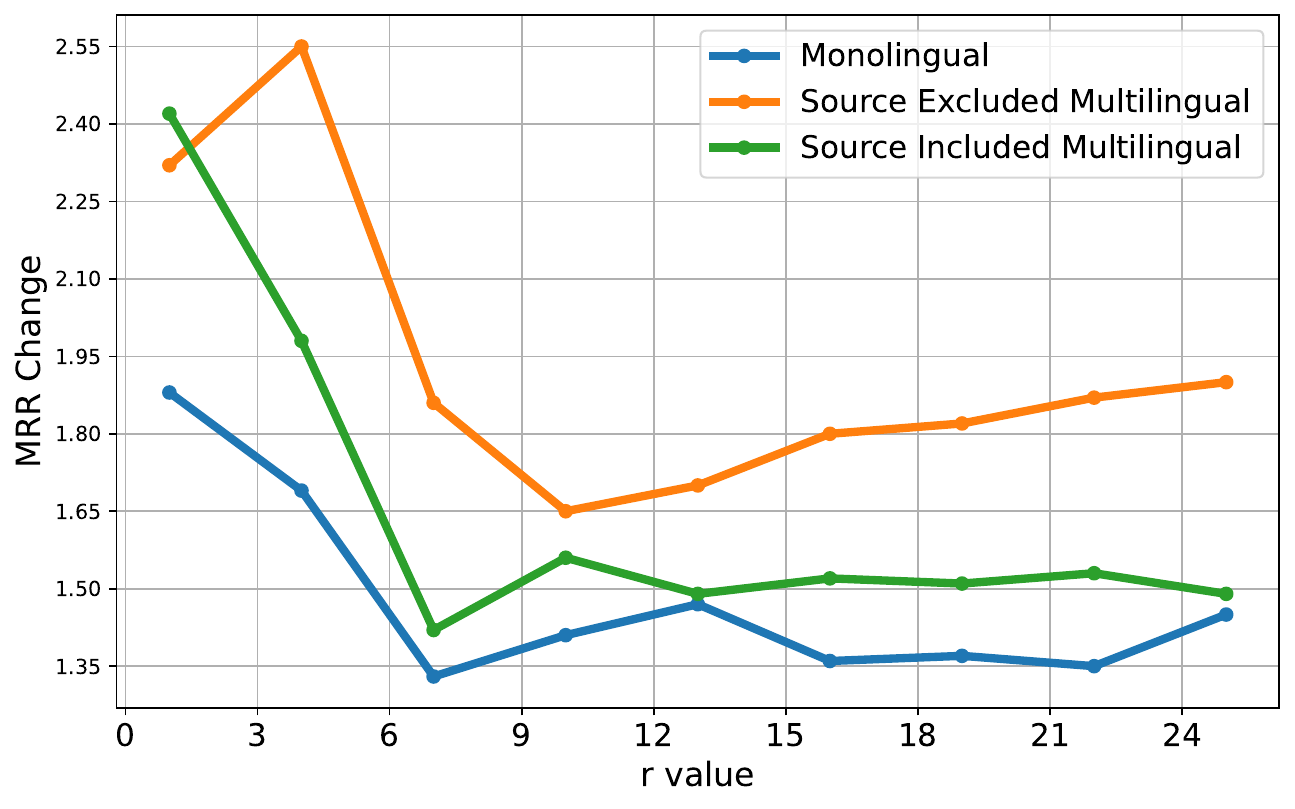}
    \subcaption{GraphCodeBERT}
    \label{fig:c2c_zero_shot_lrd_plot_graphcodebert_mean}
  \end{subfigure}
  \begin{subfigure}[b]{0.19\linewidth}
    \centering
    \includegraphics[width=\linewidth]{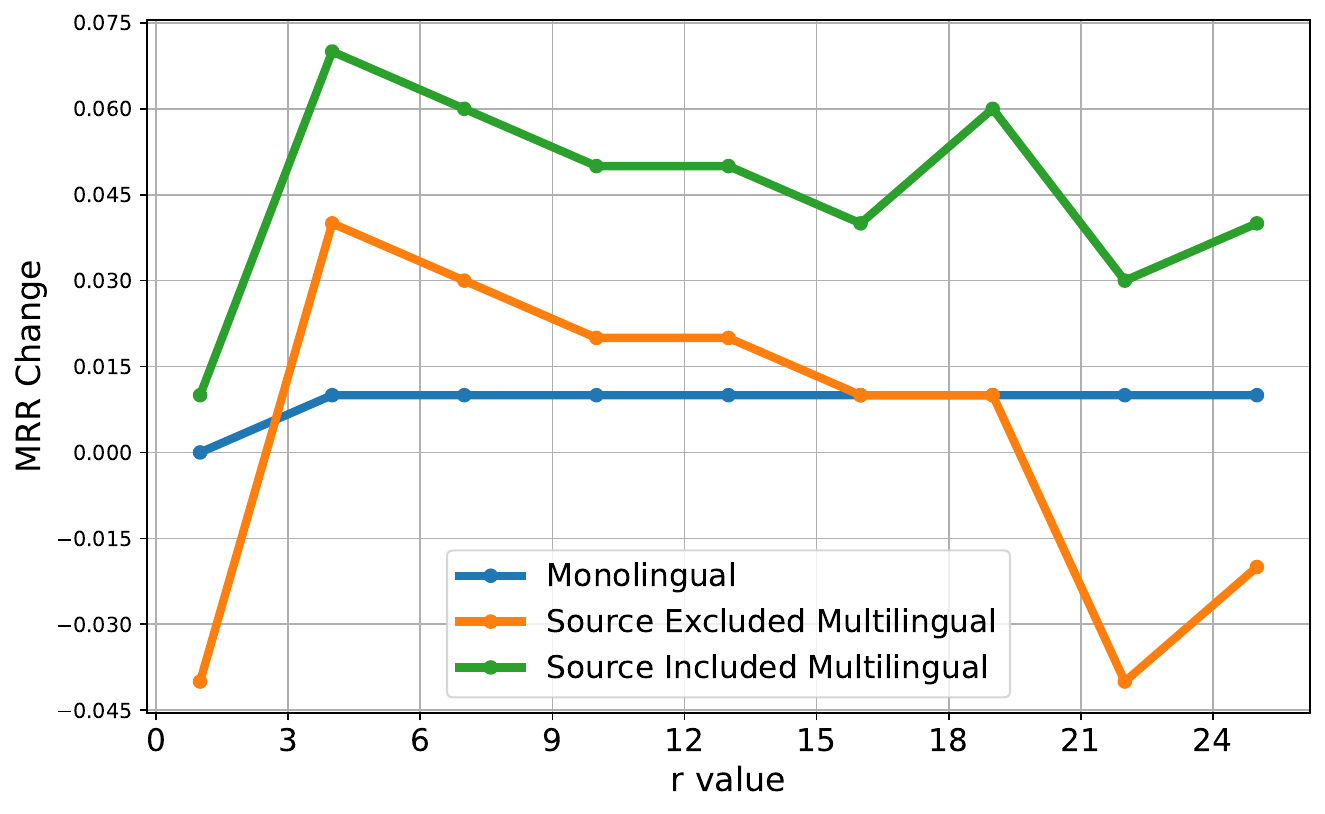}
    \subcaption{Unixcoder}
    \label{fig:c2c_zero_shot_lrd_plot_unixcoder_mean}
  \end{subfigure}
  \begin{subfigure}[b]{0.19\linewidth}
    \centering
    \includegraphics[width=\linewidth]{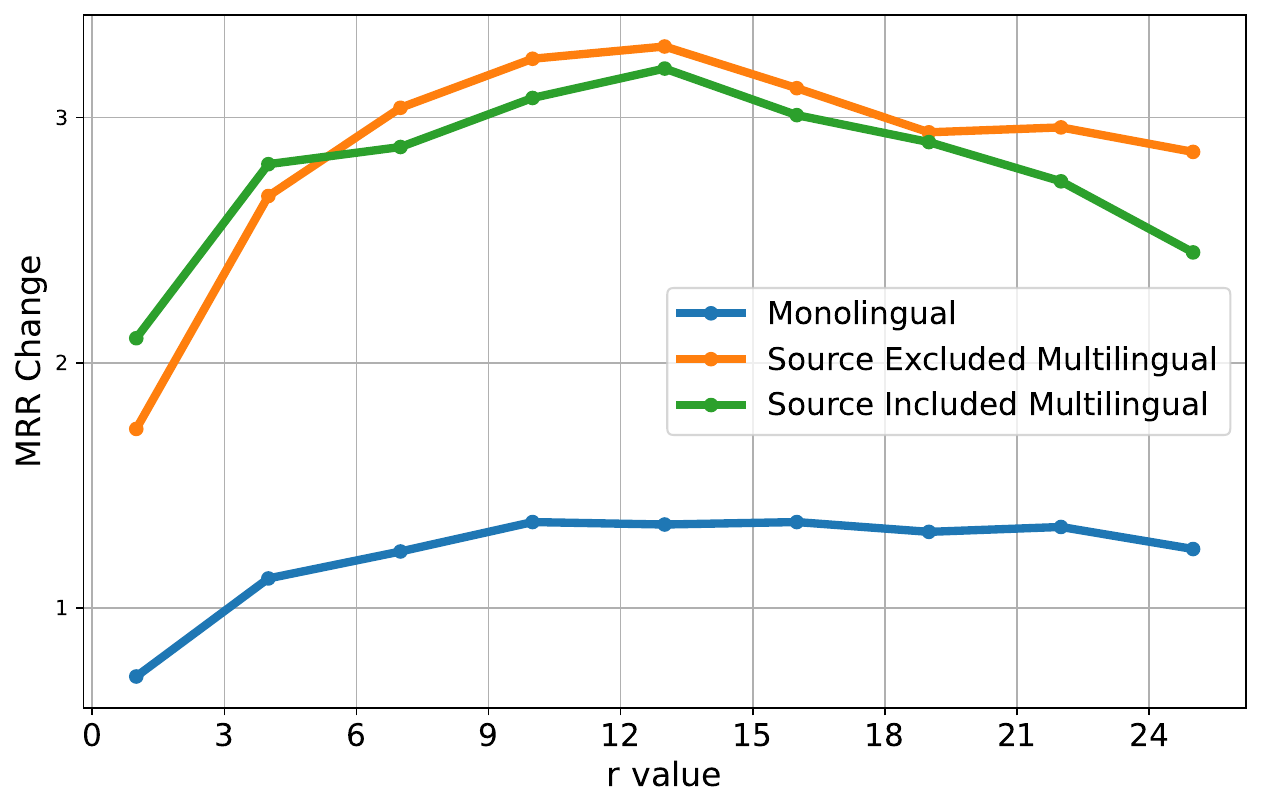}
    \subcaption{StarEncoder}
    \label{fig:c2c_zero_shot_lrd_plot_starencoder_mean}
  \end{subfigure}
  \begin{subfigure}[b]{0.19\linewidth}
    \centering
    \includegraphics[width=\linewidth]{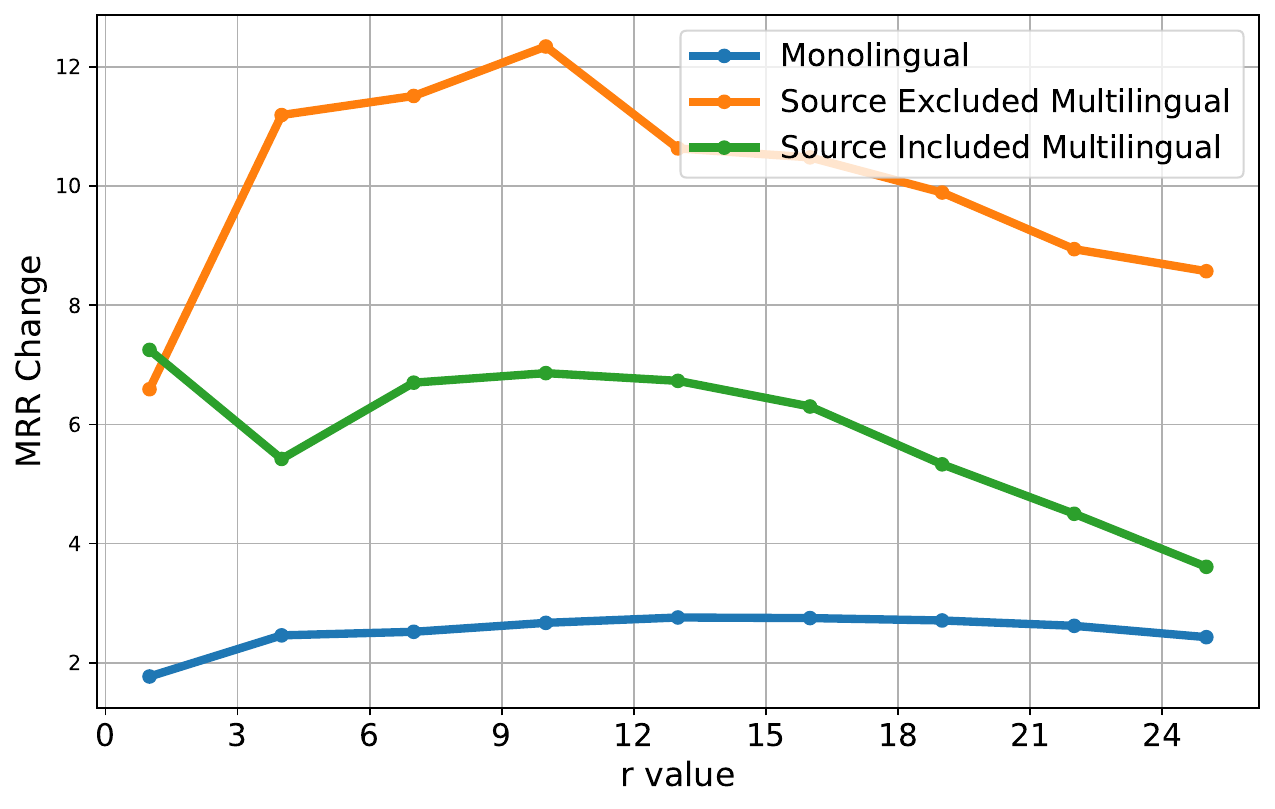}
    \subcaption{CodeT5+}
    \label{fig:c2c_zero_shot_lrd_plot_codet5+_pooler}
  \end{subfigure}

  \begin{subfigure}[b]{0.19\linewidth}
    \centering
    \includegraphics[width=\linewidth]{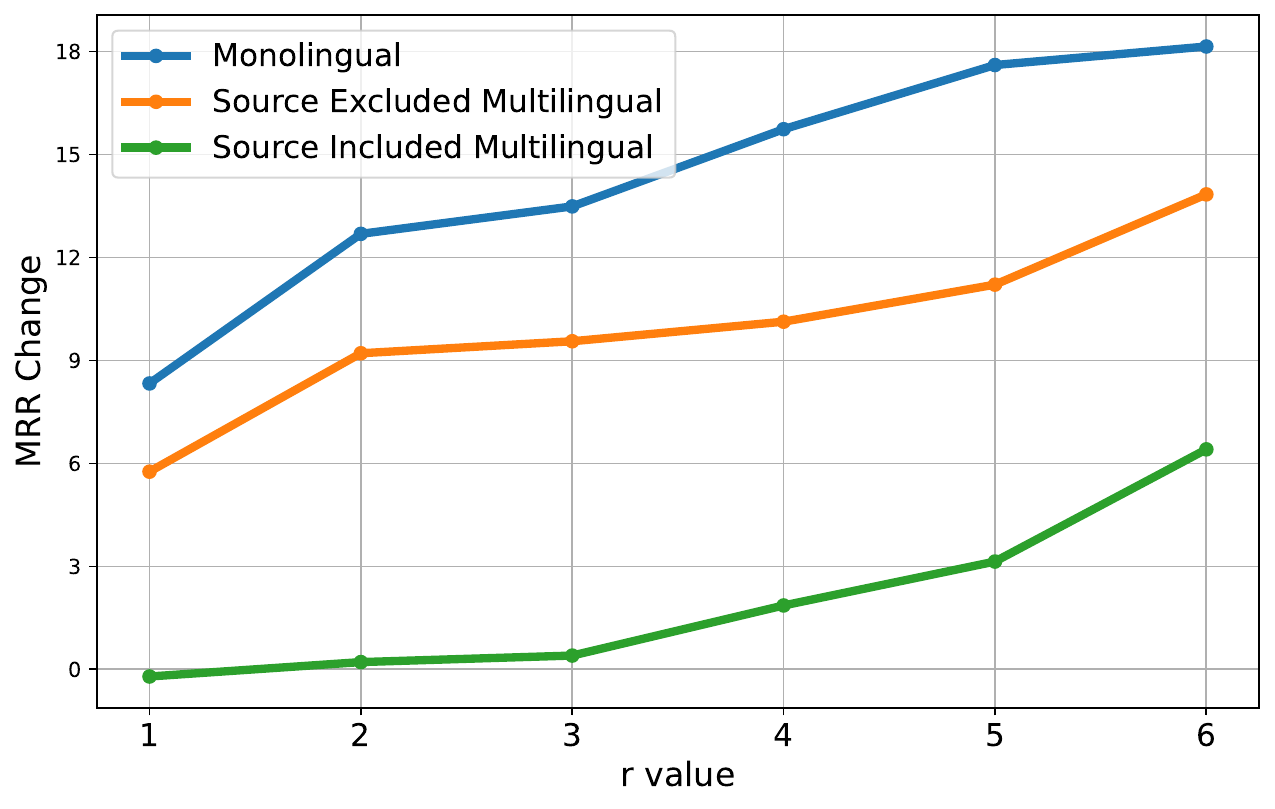}
    \subcaption{CodeBERT}
    \label{fig:c2c_zero_shot_cs_lrd_plot_codebert_mean}
  \end{subfigure}
  \begin{subfigure}[b]{0.19\linewidth}
    \centering
    \includegraphics[width=\linewidth]{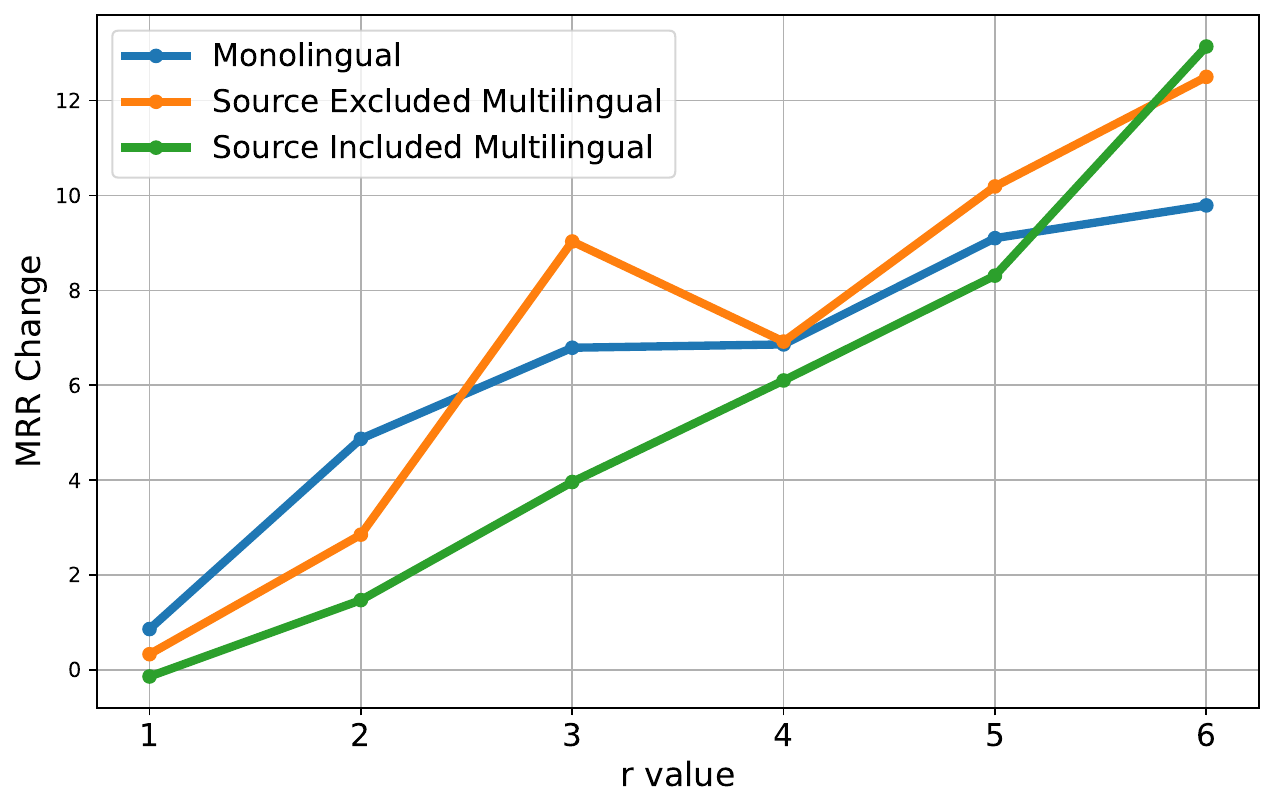}
    \subcaption{GraphCodeBERT}
    \label{fig:c2c_zero_shot_cs_lrd_plot_graphcodebert_mean}
  \end{subfigure}
  \begin{subfigure}[b]{0.19\linewidth}
    \centering
    \includegraphics[width=\linewidth]{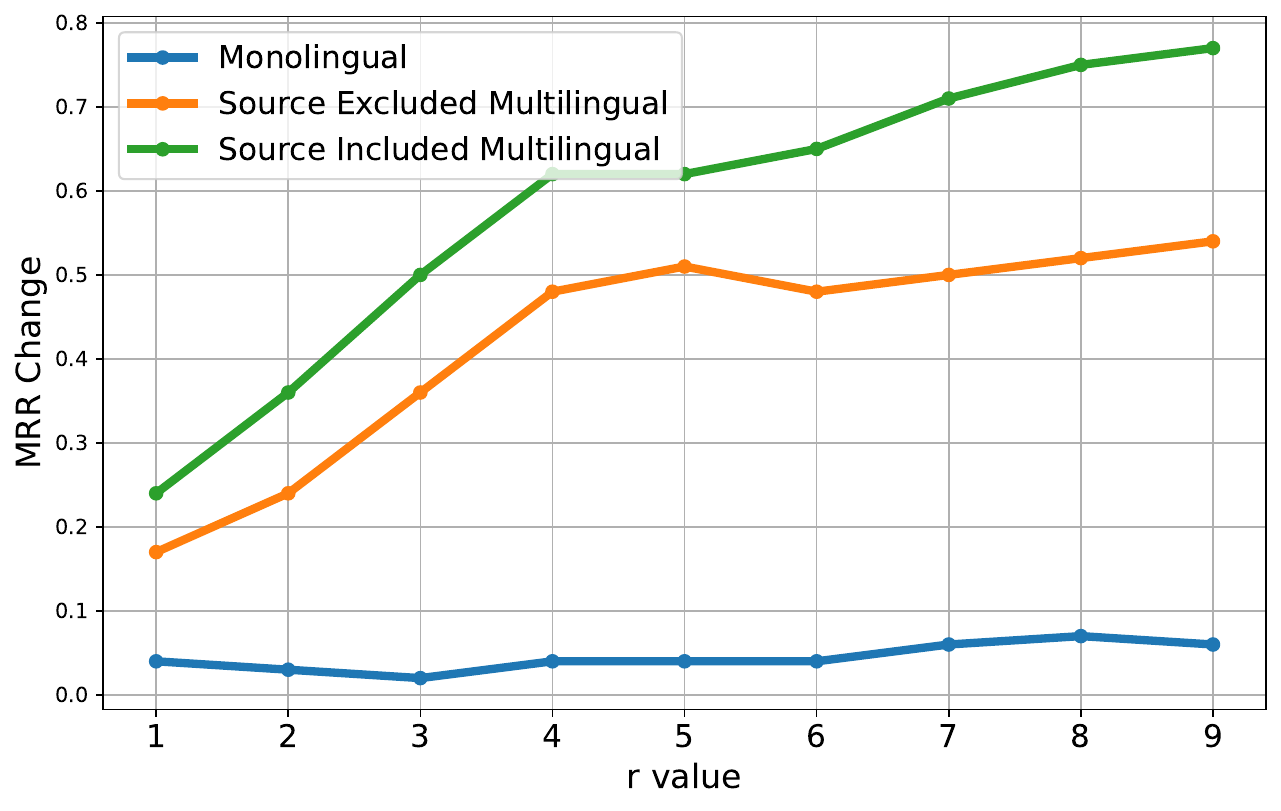}
    \subcaption{Unixcoder}
    \label{fig:c2c_zero_shot_cs_lrd_plot_unixcoder_mean}
  \end{subfigure}
  \begin{subfigure}[b]{0.19\linewidth}
    \centering
    \includegraphics[width=\linewidth]{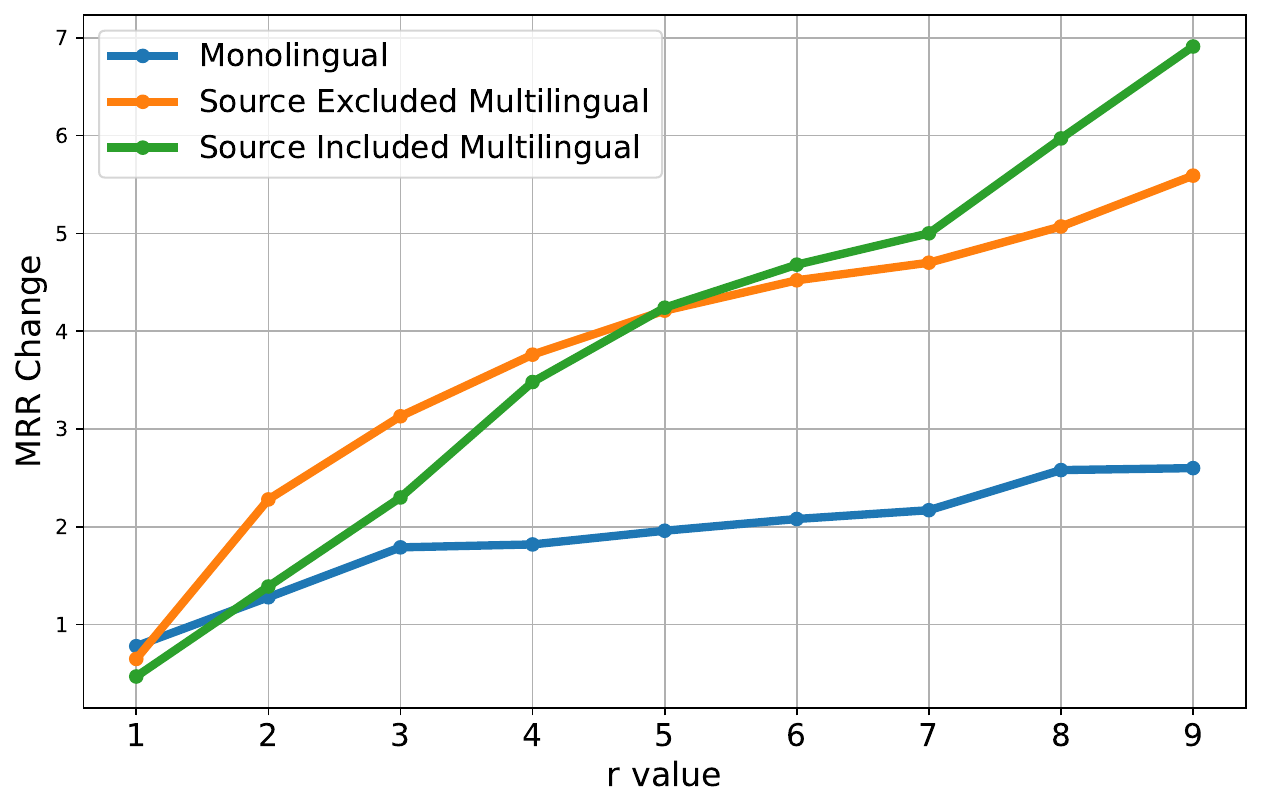}
    \subcaption{StarEncoder}
    \label{fig:c2c_zero_shot_cs_lrd_plot_starencoder_mean}
  \end{subfigure}
  \begin{subfigure}[b]{0.19\linewidth}
    \centering
    \includegraphics[width=\linewidth]{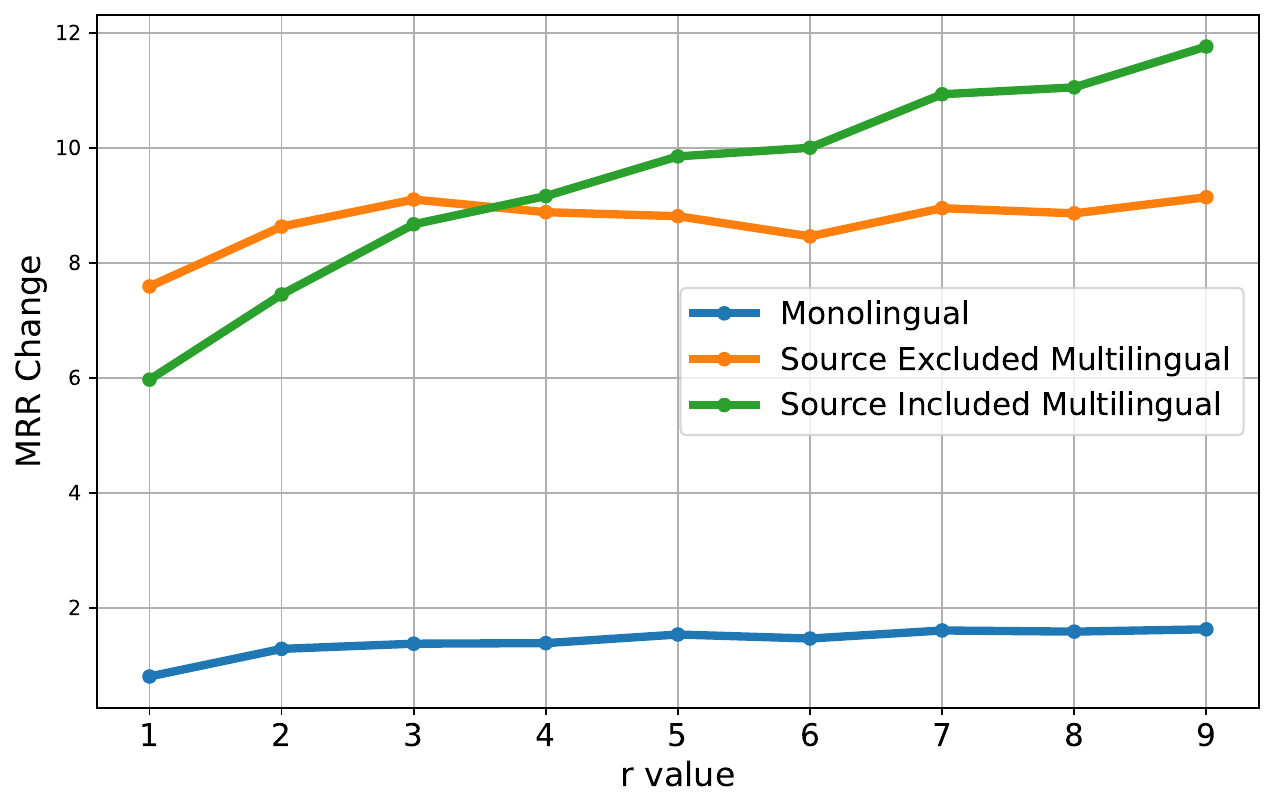}
    \subcaption{CodeT5+}
    \label{fig:c2c_zero_shot_cs_lrd_plot_codet5+_pooler}
  \end{subfigure}
  \caption{Effect of the rank $(r)$ of the language subspace on MRR change in zero-shot Code2Code search. The top row shows it for LRD, and the bottom row for CS-LRD.}
  \label{fig:c2c_zero_shot_r_value}
\end{figure*}

%% file: tables/c2c/zeroshot/c2c_codet5+_pooler.tex
\begin{tabular}{c|c|cccccccc}
\hline
CodeT5+                                       &             & C     & C\#   & C++   & Java  & Javascript & PHP   & Python & Avg.                    \\ \hline
\multirow{4}{*}{Monolingual}                  & Original    & 86.19 & 88.12 & 90.93 & 89.95 & 90.63      & 89.46 & 91.32  & 89.51                   \\
                                              & Centering   & 87.59 & 91.10 & 93.25 & 92.23 & 91.70      & 90.97 & 92.62  & 91.35 (\textbf{+1.84})  \\
                                              & LRD(r=10)   & 88.87 & 91.77 & 94.85 & 93.20 & 91.84      & 91.85 & 92.90  & 92.18 (\textbf{+2.67})  \\
                                              & CS-LRD(r=9) & 87.37 & 90.50 & 93.12 & 91.78 & 92.09      & 90.41 & 92.69  & 91.14 (\textbf{+1.63})  \\ \hline
\multirow{4}{*}{Source Excluded Multilingual} & Original    & 43.73 & 24.71 & 59.51 & 31.19 & 57.43      & 61.12 & 65.26  & 48.99                   \\
                                              & Centering   & 49.98 & 28.93 & 74.73 & 36.05 & 65.28      & 61.67 & 68.32  & 54.99 (\textbf{+6.00})  \\
                                              & LRD(r=10)   & 56.24 & 39.73 & 77.16 & 44.79 & 69.90      & 67.05 & 74.45  & 61.33 (\textbf{+12.34}) \\
                                              & CS-LRD(r=9) & 57.04 & 31.13 & 76.89 & 37.75 & 66.99      & 65.10 & 72.03  & 58.13 (\textbf{+9.14})  \\ \hline
\multirow{4}{*}{Source Included Multilingual} & Original    & 34.35 & 16.94 & 17.69 & 23.99 & 32.76      & 38.09 & 45.43  & 29.89                   \\
                                              & Centering   & 37.03 & 16.28 & 33.23 & 21.89 & 40.14      & 40.97 & 52.81  & 34.62 (\textbf{+4.73})  \\
                                              & LRD(r=10)   & 45.88 & 11.37 & 41.55 & 23.47 & 40.37      & 39.96 & 54.63  & 36.75 (\textbf{+6.86})  \\
                                              & CS-LRD(r=9) & 47.07 & 20.47 & 36.87 & 29.73 & 47.28      & 50.06 & 60.04  & 41.65 (\textbf{+11.76}) \\ \hline
\end{tabular}%

%% file: plots/t2c/t2c_zeroshot.tex
\begin{figure*}[htb!]
  \centering

  \begin{subfigure}[b]{0.19\linewidth}
    \centering
    \includegraphics[width=\linewidth]{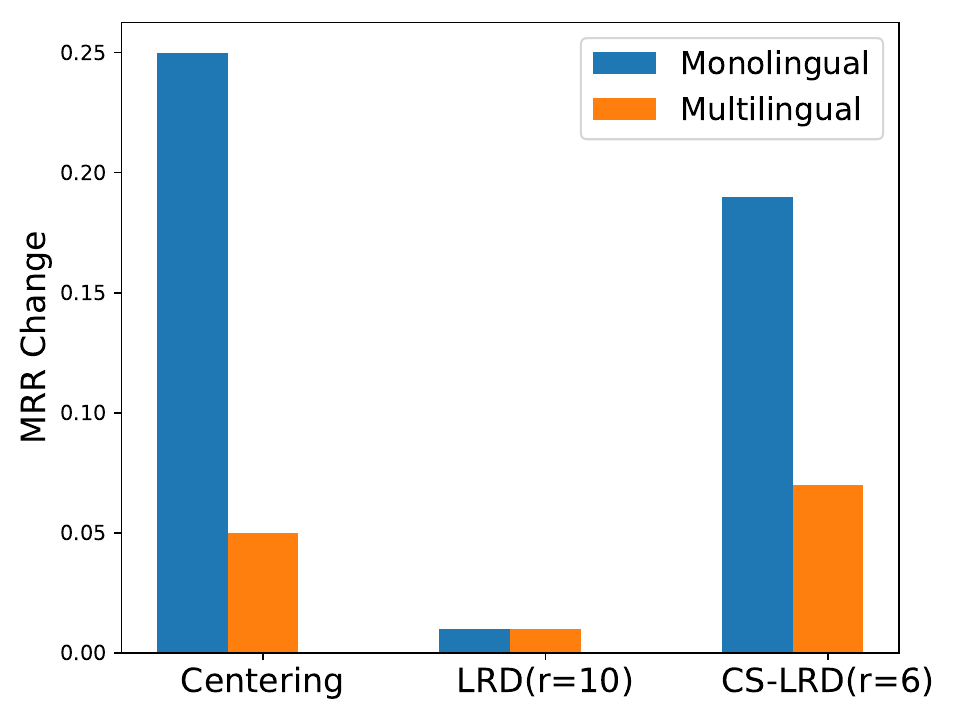}
    \subcaption{CodeBERT}
    \label{fig:t2c_zeroshot_codebert_mean}
  \end{subfigure}
  \begin{subfigure}[b]{0.19\linewidth}
    \centering
    \includegraphics[width=\linewidth]{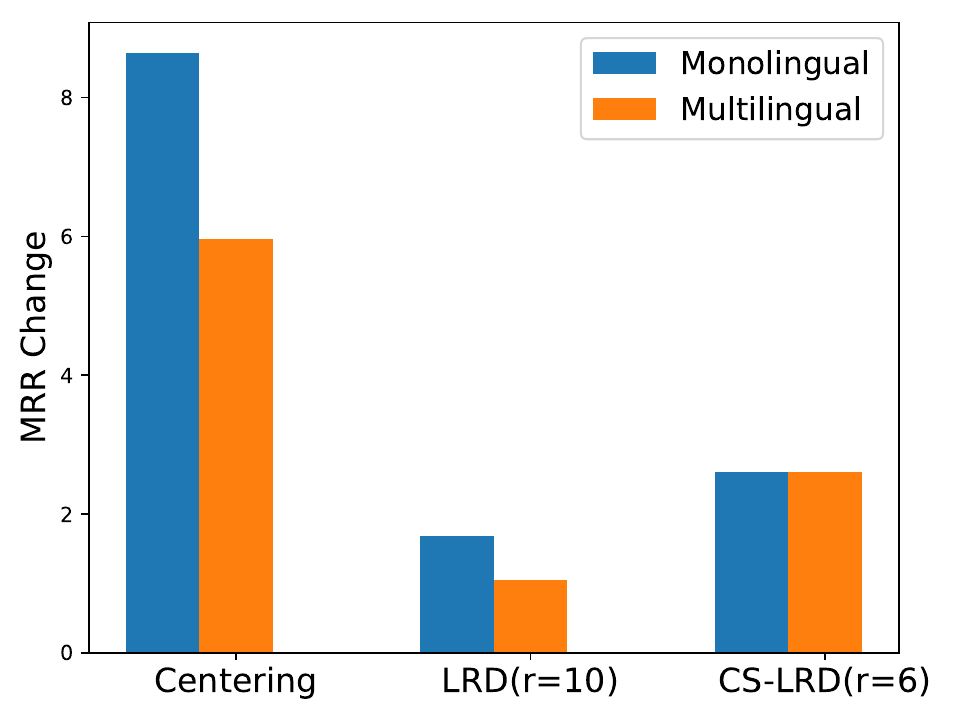}
    \subcaption{GraphCodeBERT}
    \label{fig:t2c_zeroshot_graphcodebert_mean}
  \end{subfigure}
  \begin{subfigure}[b]{0.19\linewidth}
    \centering
    \includegraphics[width=\linewidth]{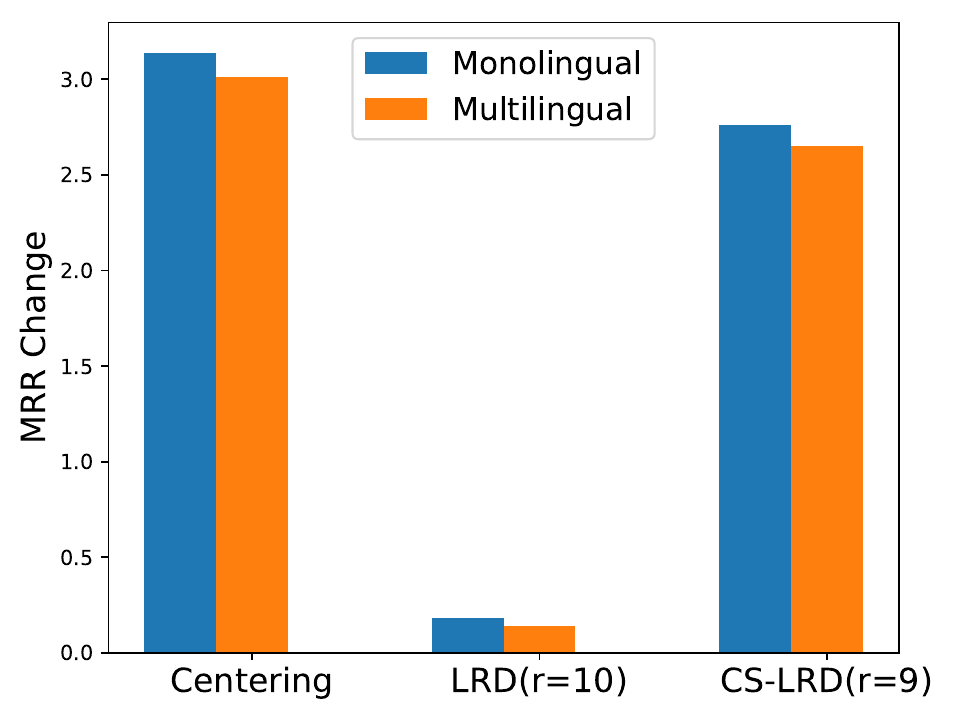}
    \subcaption{UnixCoder}
    \label{fig:t2c_zeroshot_unixcoder_mean}
  \end{subfigure}
  \begin{subfigure}[b]{0.19\linewidth}
    \centering
    \includegraphics[width=\linewidth]{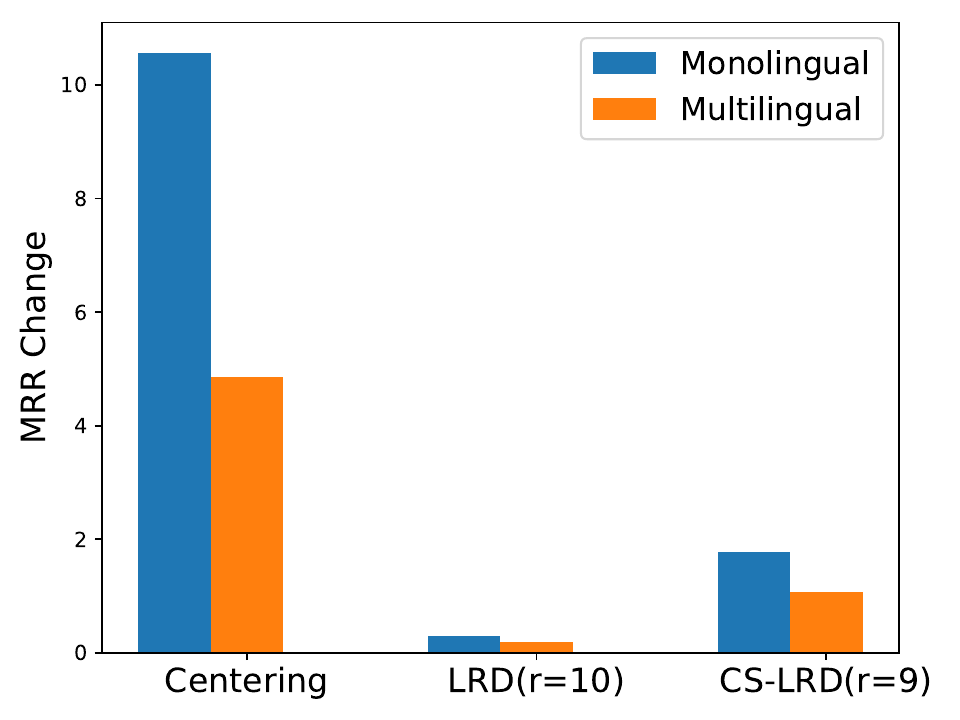}
    \subcaption{StarEncoder}
    \label{fig:t2c_zeroshot_starencoder_mean}
  \end{subfigure}
  \begin{subfigure}[b]{0.19\linewidth}
    \centering
    \includegraphics[width=\linewidth]{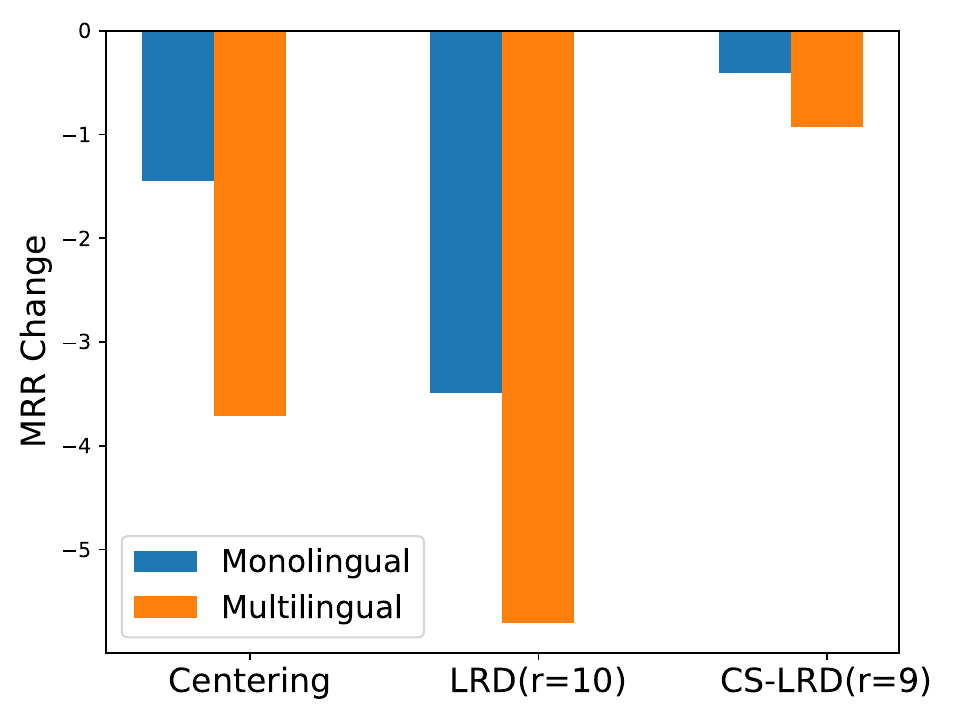}
    \subcaption{CodeT5+}
    \label{fig:t2c_zeroshot_codet5+_pooler}
  \end{subfigure}
  \caption{Absolute change in MRR after removing language components in zero-shot Text2Code search.}
  \label{fig:t2c_zeroshot}
\end{figure*}

%% file: plots/t2c/t2c_zeroshot_r_value.tex
\begin{figure*}[htb!]
  \centering

  \begin{subfigure}[b]{0.19\linewidth}
    \centering
    \includegraphics[width=\linewidth]{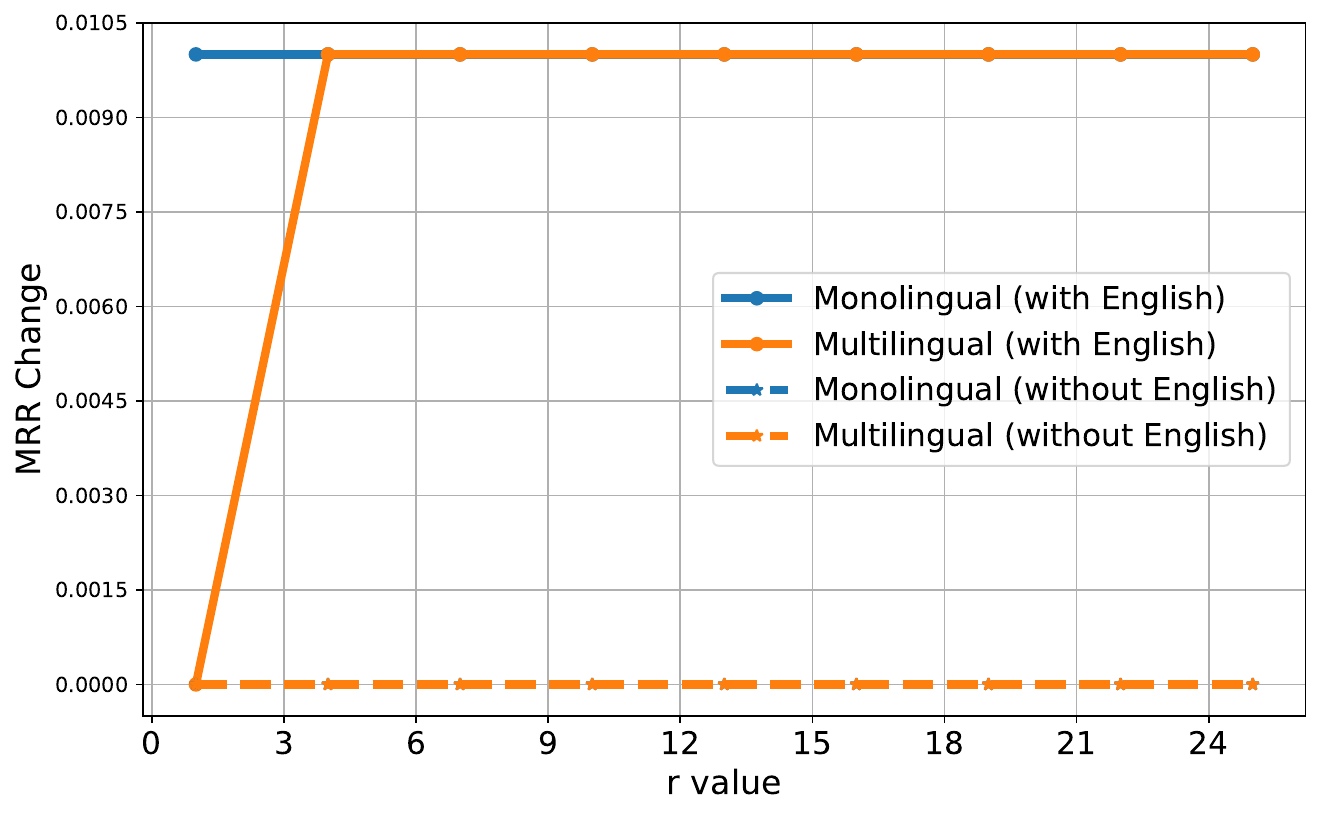}
    \subcaption{CodeBERT}
    \label{fig:t2c_zero_shot_lrd_plot_codebert_mean}
  \end{subfigure}
  \begin{subfigure}[b]{0.19\linewidth}
    \centering
    \includegraphics[width=\linewidth]{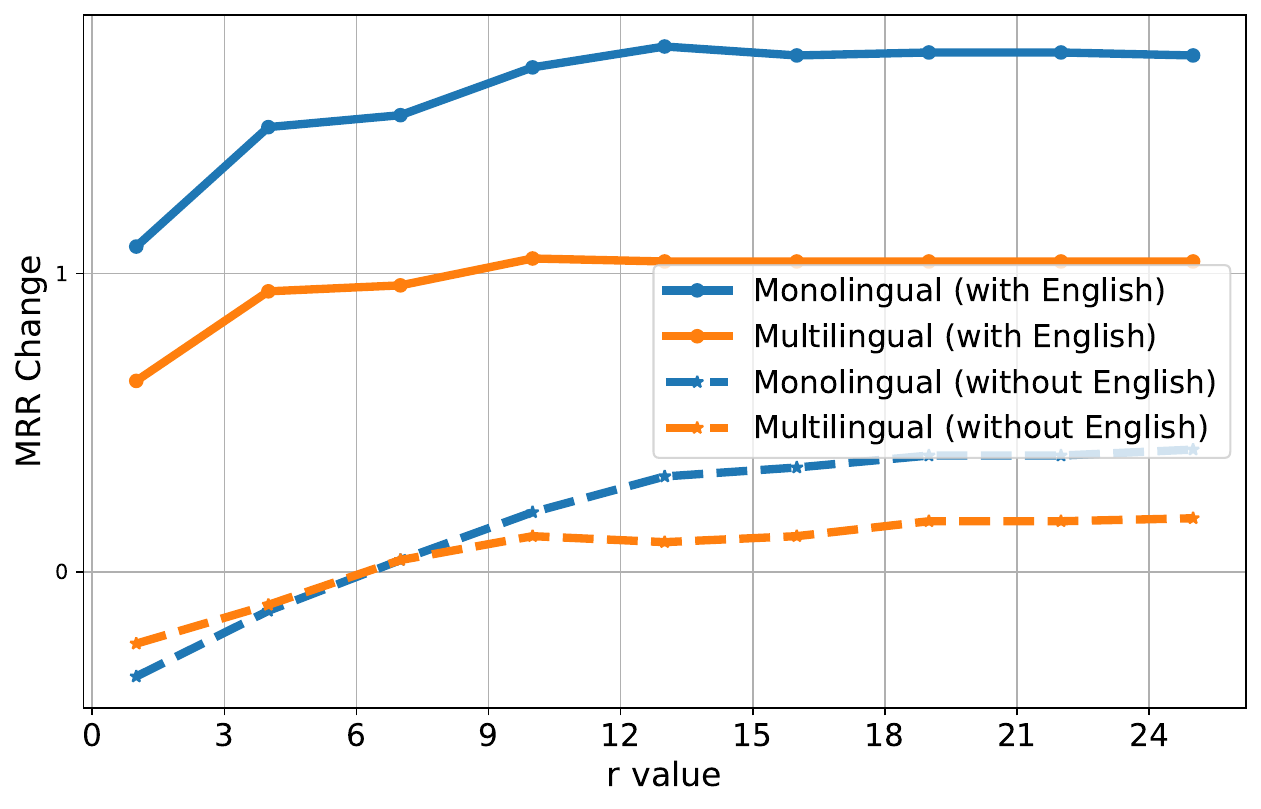}
    \subcaption{GraphCodeBERT}
    \label{fig:t2c_zero_shot_lrd_plot_graphcodebert_mean}
  \end{subfigure}
  \begin{subfigure}[b]{0.19\linewidth}
    \centering
    \includegraphics[width=\linewidth]{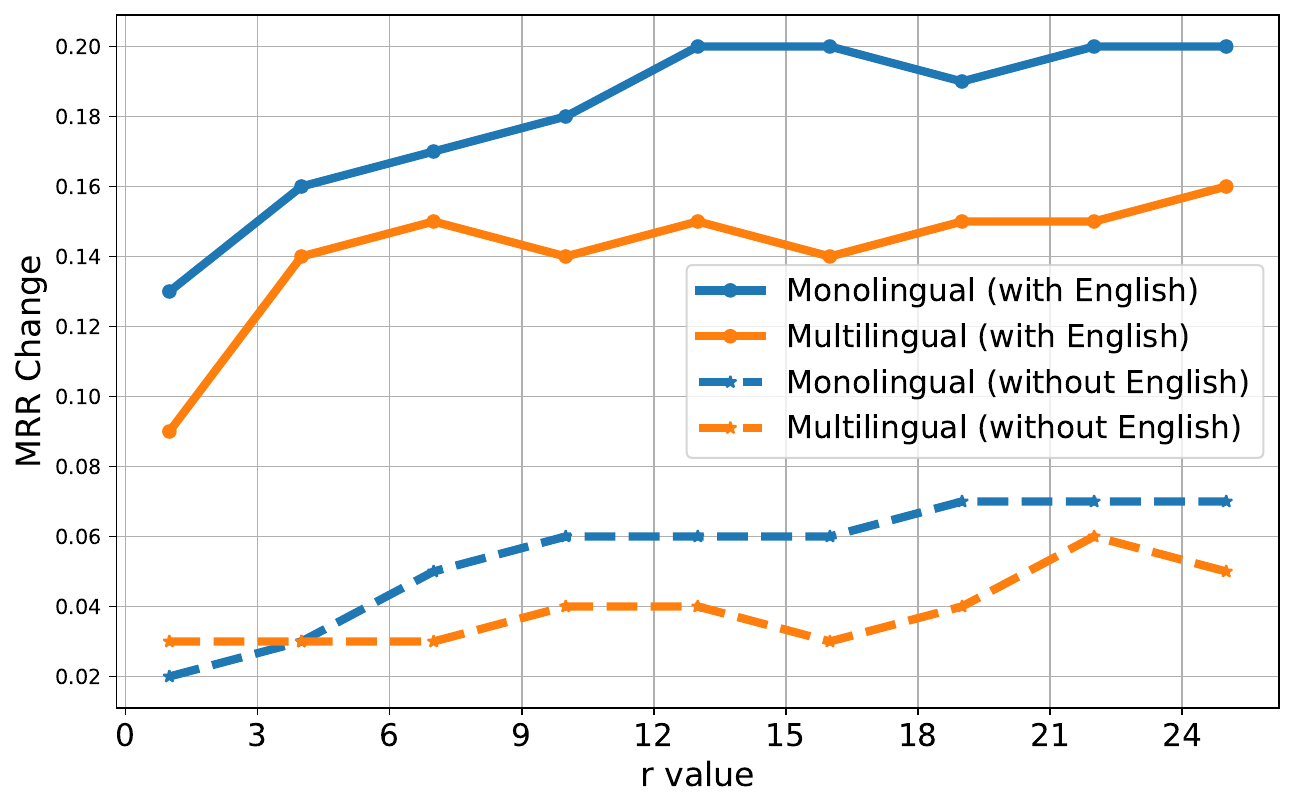}
    \subcaption{UnixCoder}
    \label{fig:t2c_zero_shot_lrd_plot_unixcoder_mean}
  \end{subfigure}
  \begin{subfigure}[b]{0.19\linewidth}
    \centering
    \includegraphics[width=\linewidth]{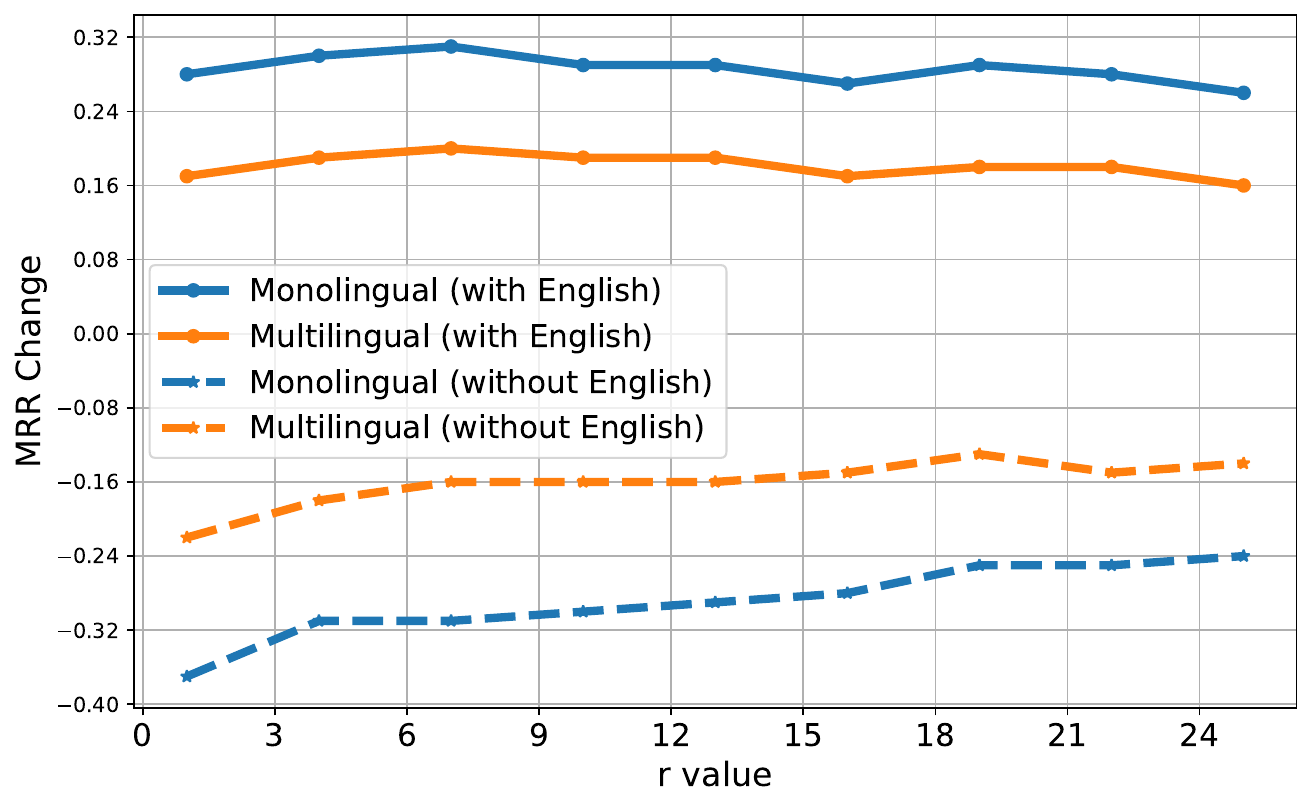}
    \subcaption{StarEncoder}
    \label{fig:t2c_zero_shot_lrd_plot_starencoder_mean}
  \end{subfigure}
  \begin{subfigure}[b]{0.19\linewidth}
    \centering
    \includegraphics[width=\linewidth]{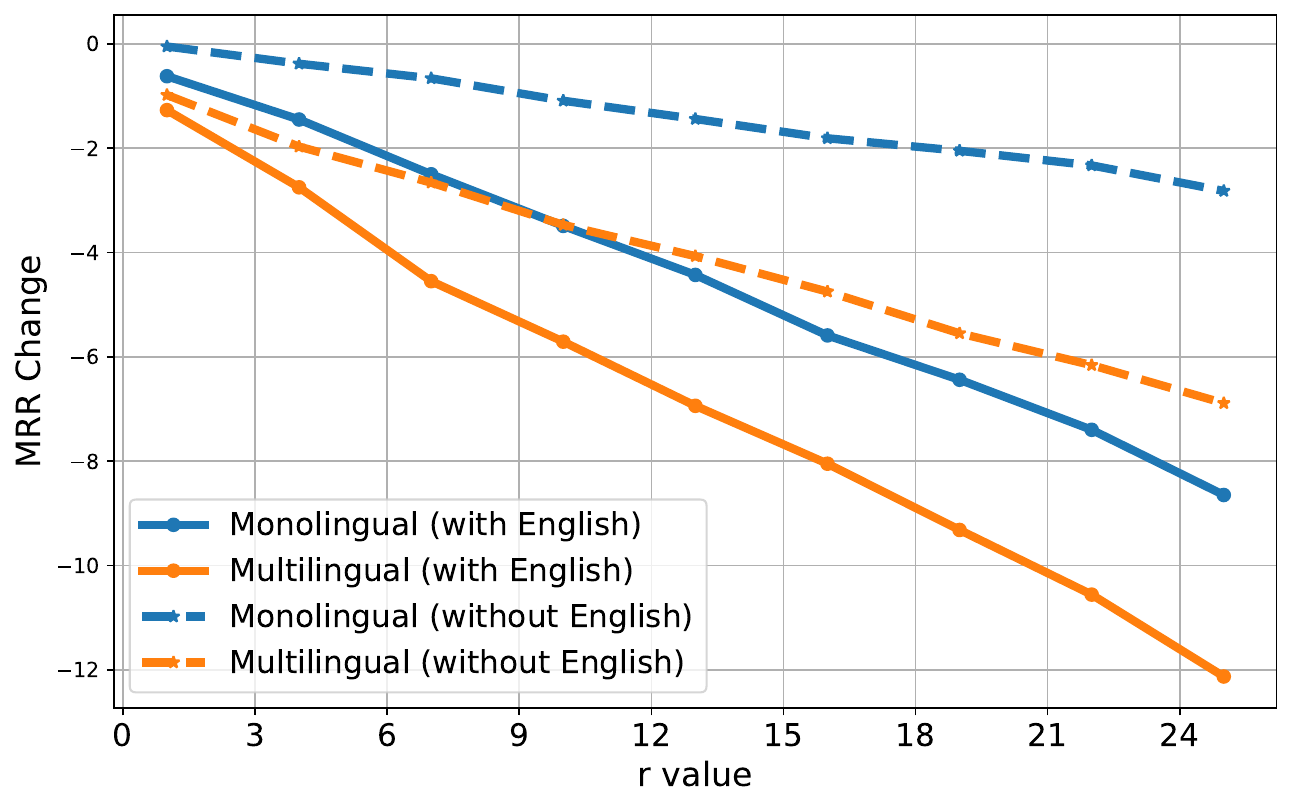}
    \subcaption{CodeT5+}
    \label{fig:t2c_zero_shot_lrd_plot_codet5+_pooler}
  \end{subfigure}

  \begin{subfigure}[b]{0.19\linewidth}
    \centering
    \includegraphics[width=\linewidth]{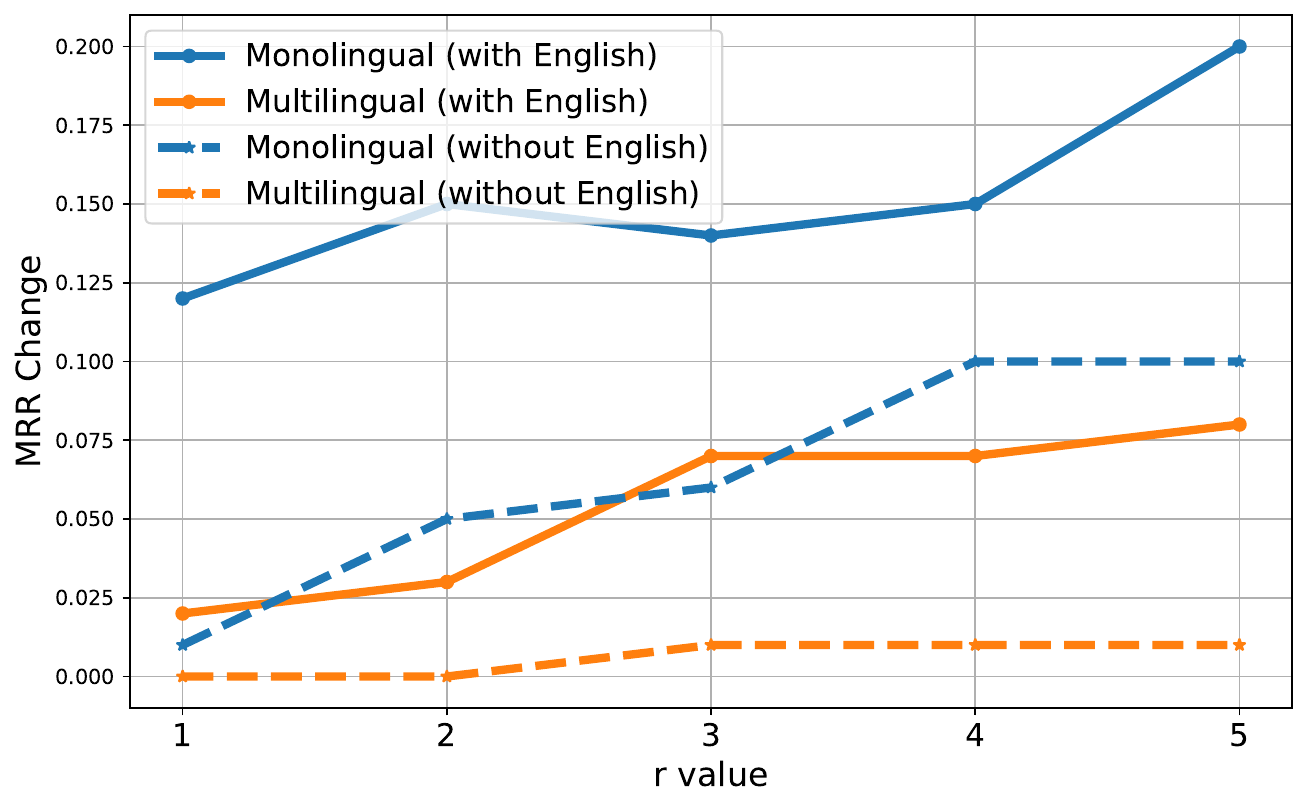}
    \subcaption{CodeBERT}
    \label{fig:t2c_zero_shot_cs_lrd_plot_codebert_mean}
  \end{subfigure}
  \begin{subfigure}[b]{0.19\linewidth}
    \centering
    \includegraphics[width=\linewidth]{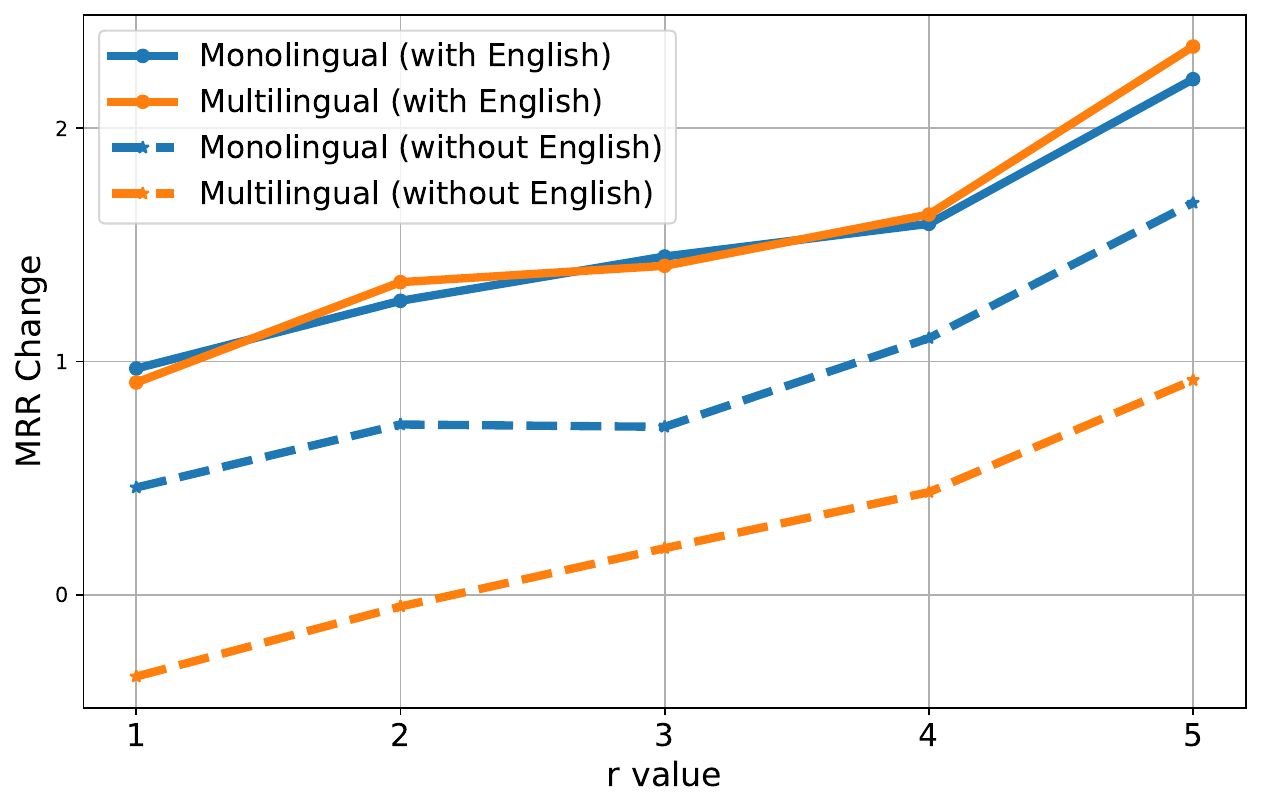}
    \subcaption{GraphCodeBERT}
    \label{fig:t2c_zero_shot_cs_lrd_plot_graphcodebert_mean}
  \end{subfigure}
  \begin{subfigure}[b]{0.19\linewidth}
    \centering
    \includegraphics[width=\linewidth]{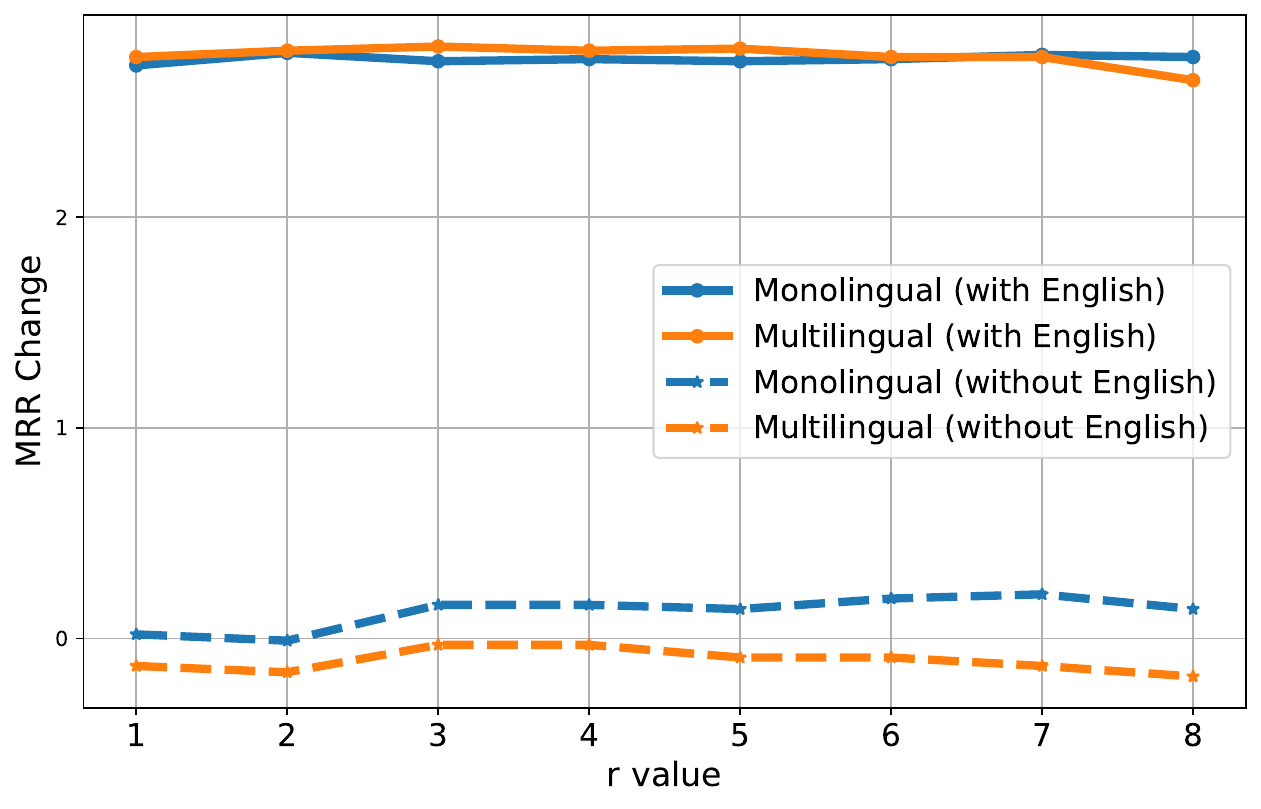}
    \subcaption{UnixCoder}
\label{fig:t2c_zero_shot_cs_lrd_plot_unixcoder_mean}
  \end{subfigure}
  \begin{subfigure}[b]{0.19\linewidth}
    \centering
    \includegraphics[width=\linewidth]{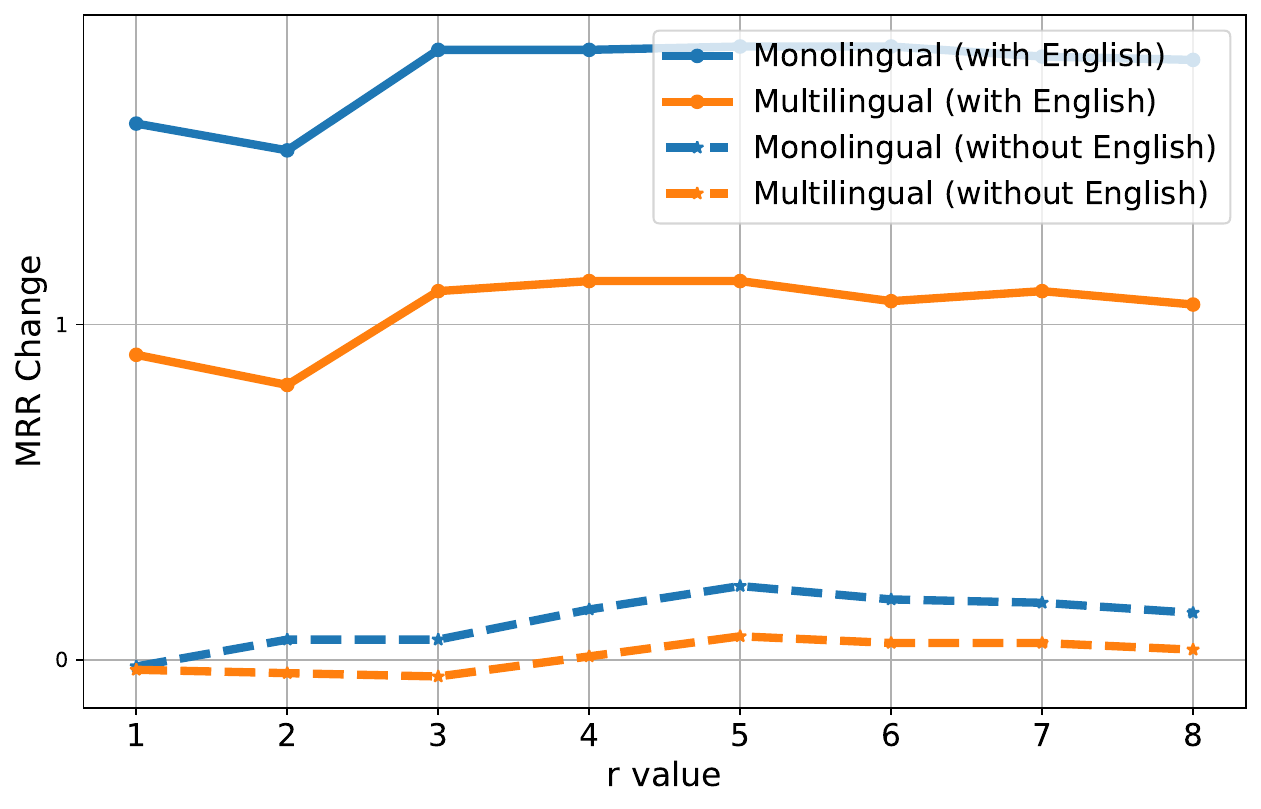}
    \subcaption{StarEncoder}
    \label{fig:t2c_zero_shot_cs_lrd_plot_starencoder_mean}
  \end{subfigure}
  \begin{subfigure}[b]{0.19\linewidth}
    \centering
    \includegraphics[width=\linewidth]{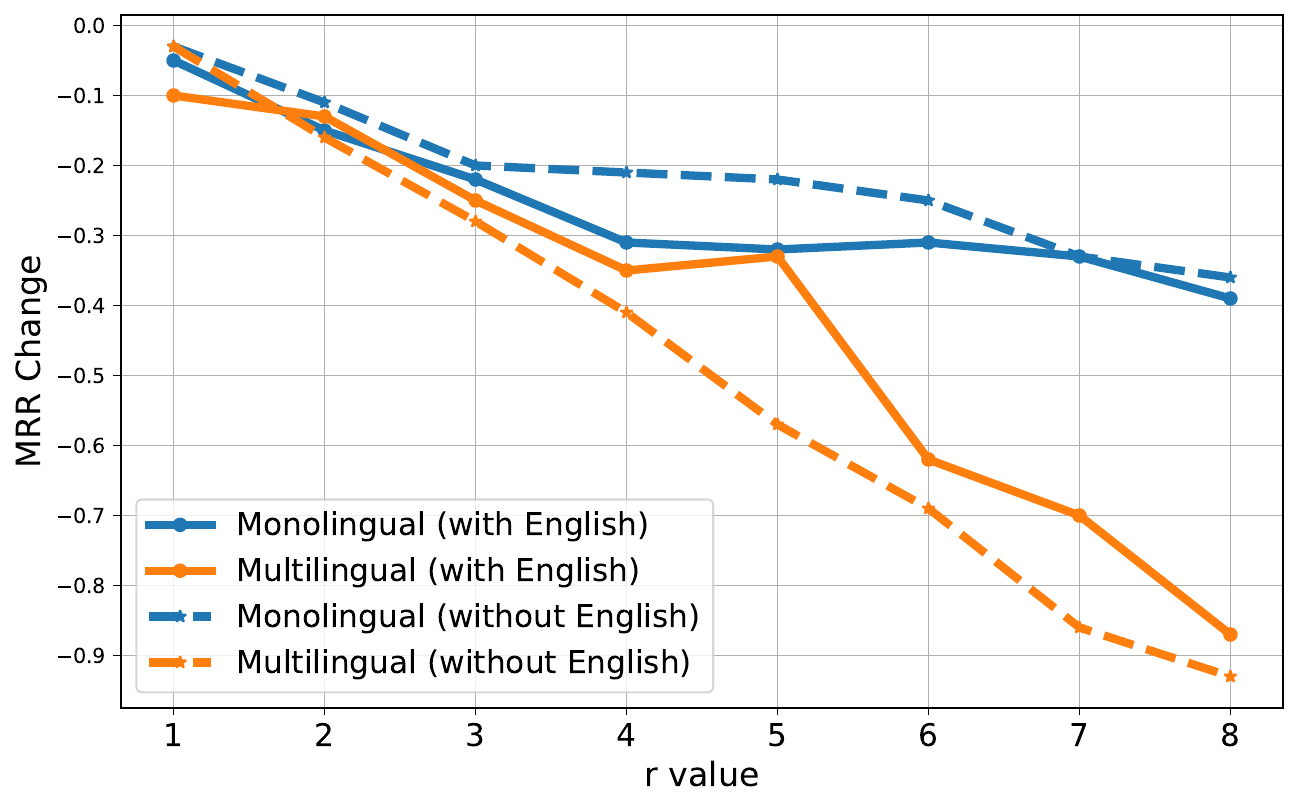}
    \subcaption{CodeT5+}
    \label{fig:t2c_zero_shot_cs_lrd_plot_codet5+_pooler}
  \end{subfigure}

  \caption{Effect of the rank $(r)$ of the language subspace on MRR change in zero-shot Text2Code search. The top row shows it for LRD, and the bottom row for CS-LRD. 'Without English' and 'With English' indicate cases where query embeddings remain untransformed and transformed, respectively.}
  \label{fig:t2c_zero_shot_r_value}
\end{figure*}

%% file: tables/t2c/zeroshot/t2c_unixcoder_mean.tex
\begin{tabular}{c|c|ccccccc}
\hline
Unixcoder (mean)              &             & Go    & Java  & Javascript & PHP   & Python & Ruby  & Avg.                   \\ \hline
\multirow{4}{*}{Monolingual}  & Original    & 61.38 & 44.23 & 40.93      & 35.22 & 42.43  & 55.30 & 46.58                  \\
                              & Centering   & 64.01 & 47.77 & 44.25      & 38.98 & 46.15  & 57.16 & 49.72 (\textbf{+3.14}) \\
                              & LRD(r=10)   & 61.51 & 44.46 & 41.15      & 35.38 & 42.61  & 55.44 & 46.76 (\textbf{+0.18}) \\
                              & CS-LRD(r=9) & 63.10 & 47.62 & 43.94      & 38.61 & 45.98  & 56.79 & 49.34 (\textbf{+2.76}) \\ \hline
\multirow{4}{*}{Multilingual} & Original    & 54.05 & 36.40 & 27.87      & 29.84 & 35.71  & 34.83 & 36.45                  \\
                              & Centering   & 54.65 & 40.14 & 30.42      & 33.21 & 40.20  & 38.11 & 39.46 (\textbf{+3.01}) \\
                              & LRD(r=10)   & 54.18 & 36.62 & 27.96      & 30.03 & 35.85  & 34.91 & 36.59 (\textbf{+0.14}) \\
                              & CS-LRD(r=9) & 55.21 & 39.95 & 30.07      & 33.06 & 39.37  & 36.94 & 39.10 (\textbf{+2.65}) \\ \hline
\end{tabular}%

%% file: plots/c2c_zeroshot_abalation_alignment.tex
\begin{figure*}[htb!]
  \centering
  \begin{subfigure}[b]{0.19\linewidth}
    \centering
    \includegraphics[width=\linewidth]{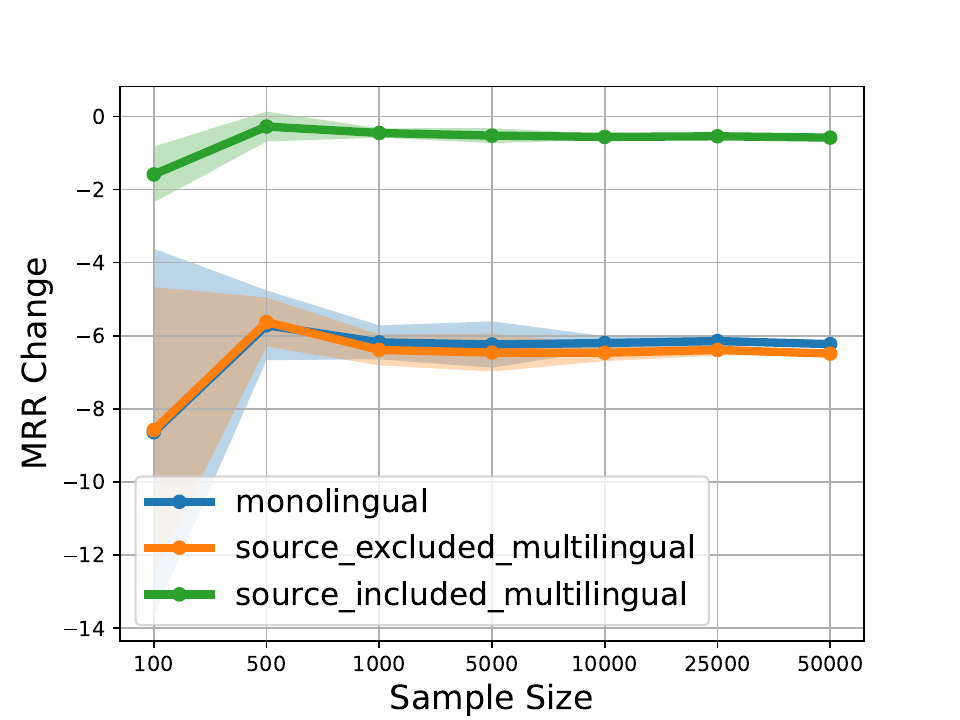}
    \subcaption{CodeBERT}
    \label{fig:centering_abalation_codebert}
  \end{subfigure}
  \begin{subfigure}[b]{0.19\linewidth}
    \centering
    \includegraphics[width=\linewidth]{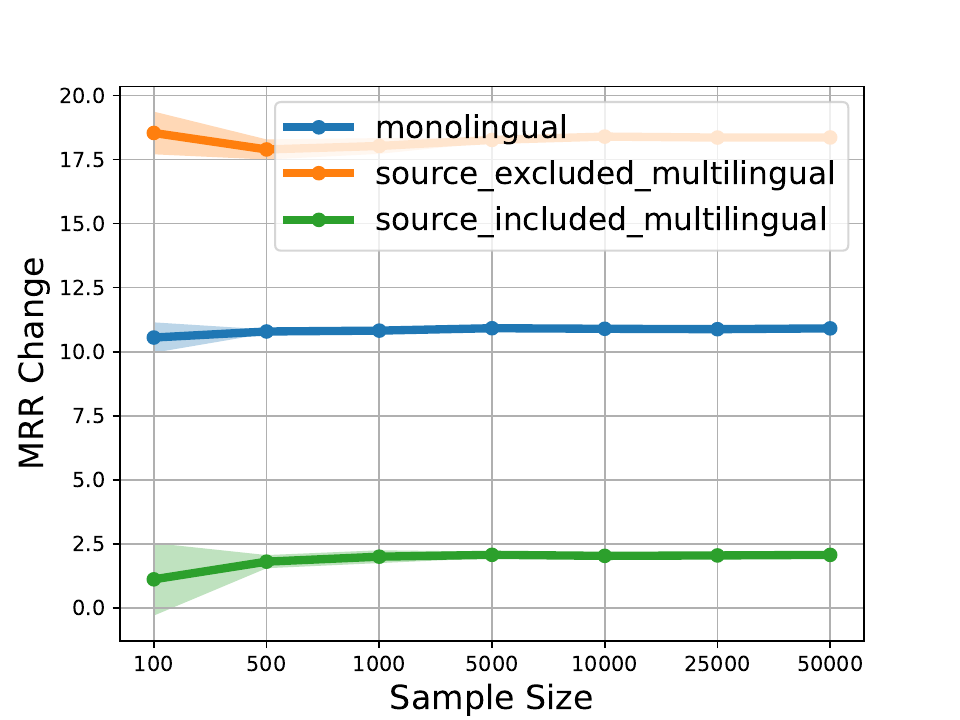}
    \subcaption{GraphCodeBERT}
   \label{fig:centering_abalation_graphcodebert.pdf}
  \end{subfigure}
  \begin{subfigure}[b]{0.19\linewidth}
    \centering
    \includegraphics[width=\linewidth]{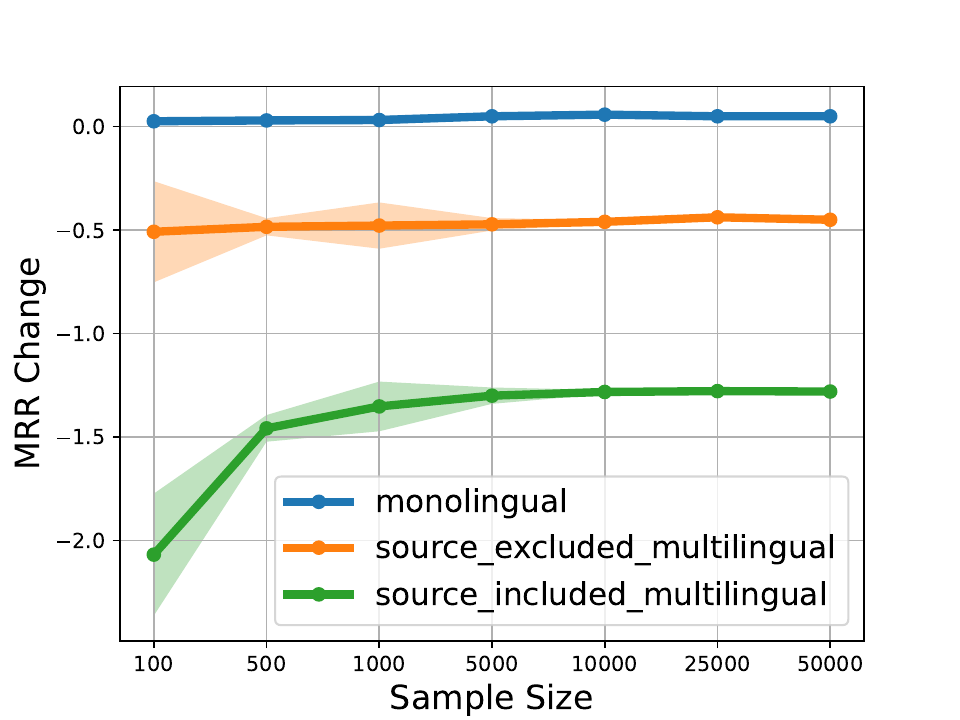}
    \subcaption{UnixCoder}
    \label{fig:centering_abalation_unixcoder}
  \end{subfigure}
  \begin{subfigure}[b]{0.19\linewidth}
    \centering
    \includegraphics[width=\linewidth]{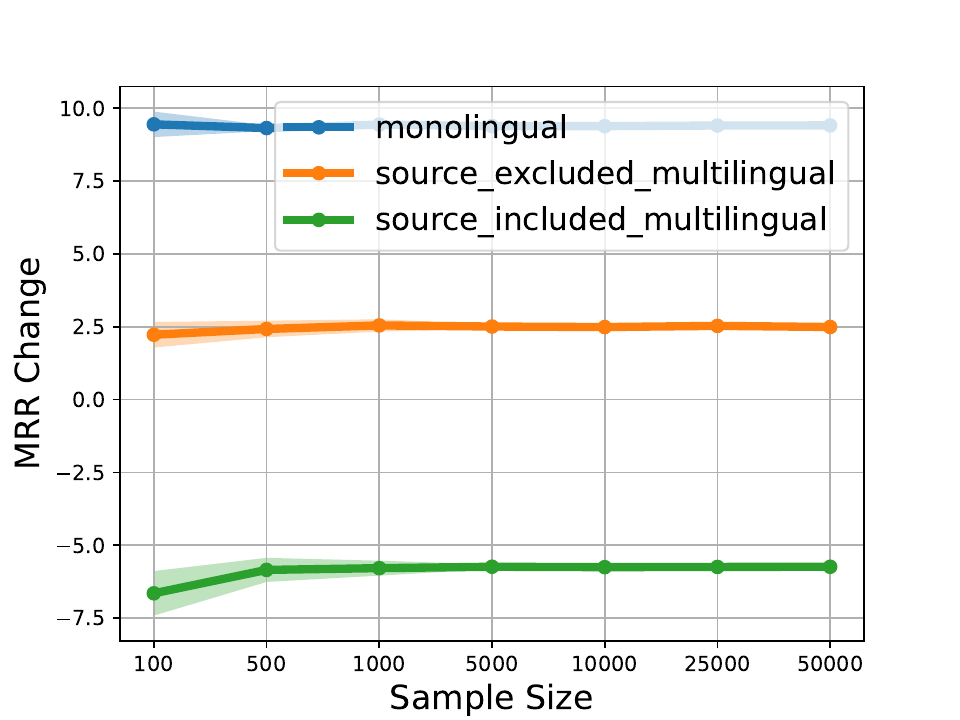}
    \subcaption{StarEncoder}
    \label{fig:centering_abalation_starencoder}
  \end{subfigure}
  \begin{subfigure}[b]{0.19\linewidth}
    \centering
    \includegraphics[width=\linewidth]{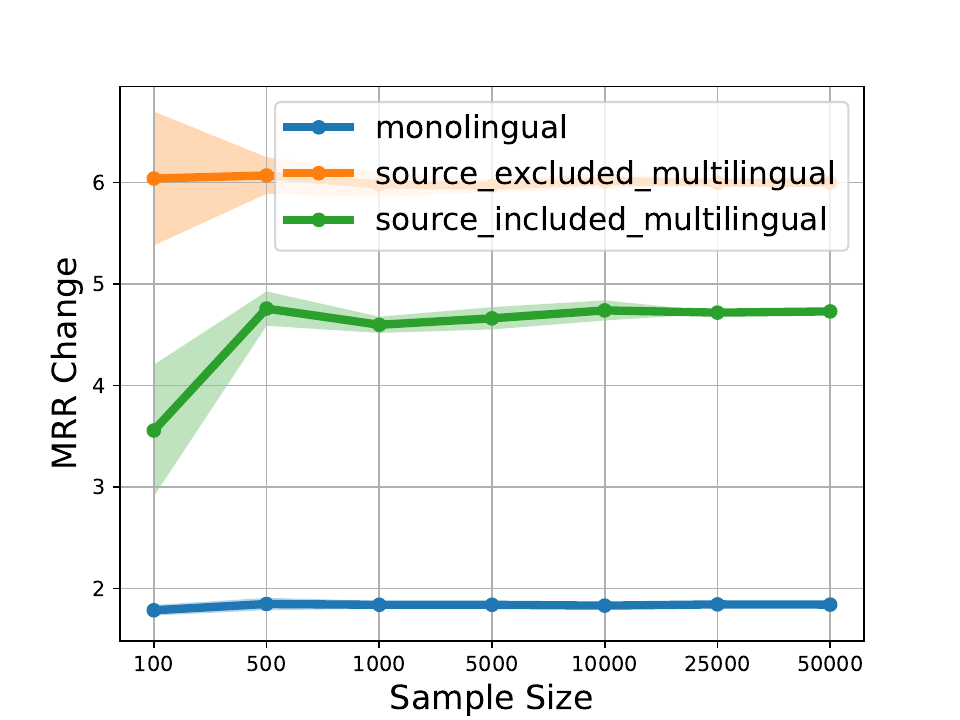}
    \subcaption{CodeT5+}
    \label{fig:centering_abalation_codet5+}
  \end{subfigure}

  \begin{subfigure}[b]{0.19\linewidth}
    \centering
    \includegraphics[width=\linewidth]{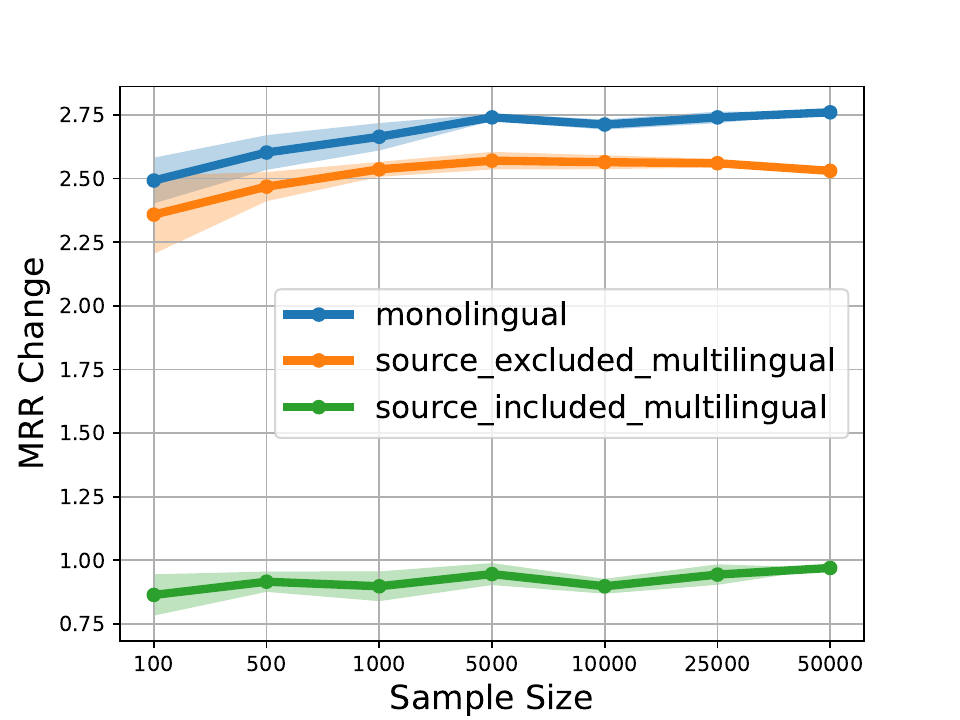}
    \subcaption{CodeBERT}
    \label{fig:LRD_abalation_codebert}
  \end{subfigure}
  \begin{subfigure}[b]{0.19\linewidth}
    \centering
    \includegraphics[width=\linewidth]{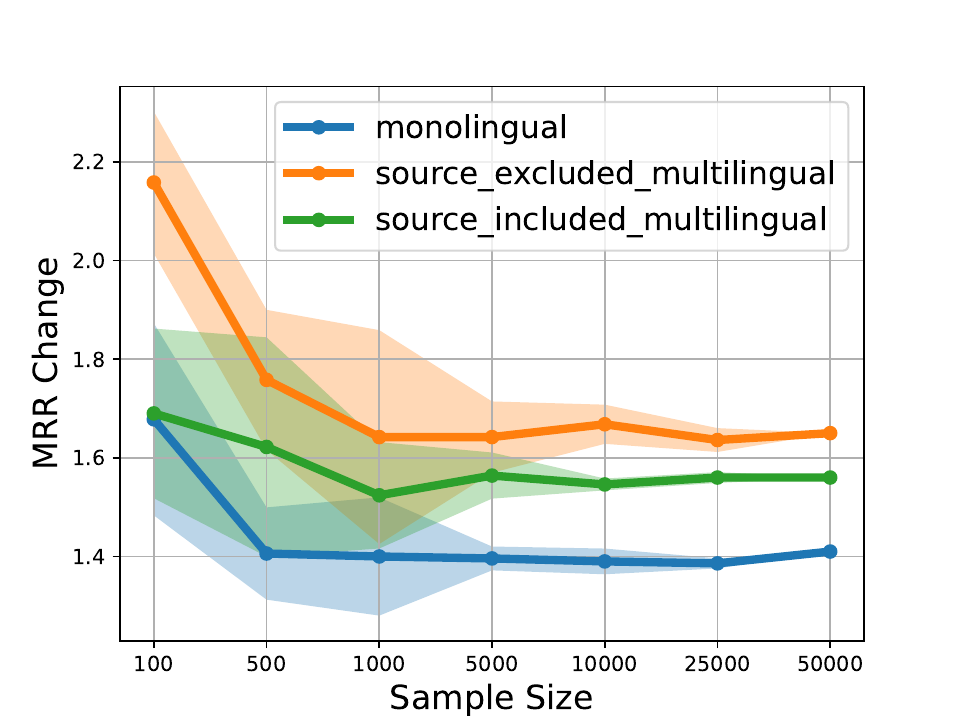}
    \subcaption{GraphCodeBERT}
\label{fig:LRD_abalation_graphcodebert}
  \end{subfigure}
  \begin{subfigure}[b]{0.19\linewidth}
    \centering
    \includegraphics[width=\linewidth]{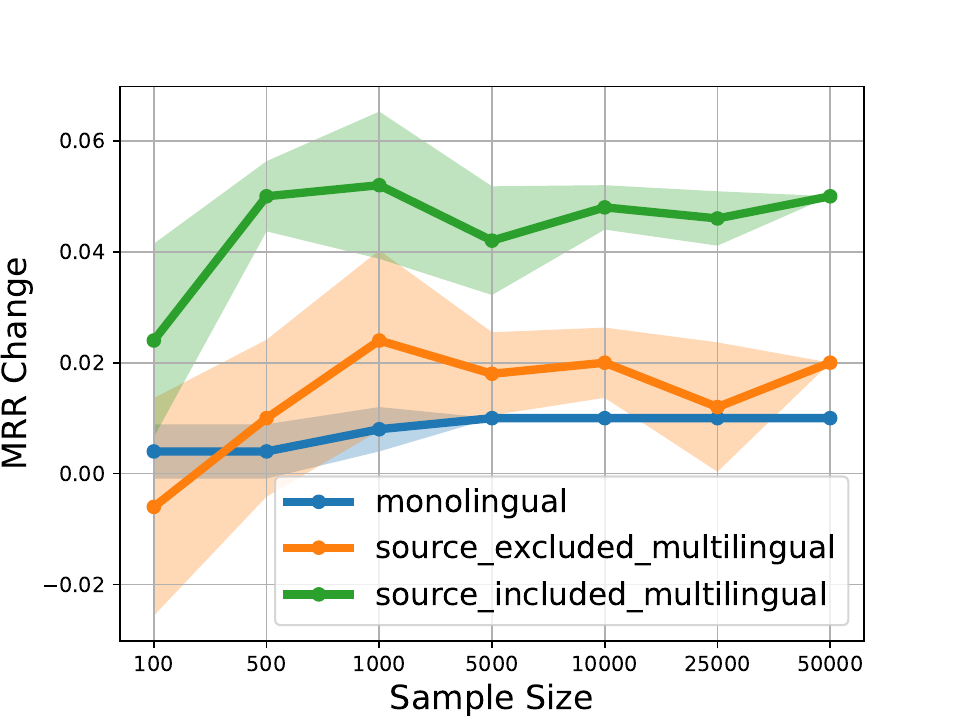}
    \subcaption{UnixCoder}
    \label{fig:LRD_abalation_unixcoder}
  \end{subfigure}
  \begin{subfigure}[b]{0.19\linewidth}
    \centering
    \includegraphics[width=\linewidth]{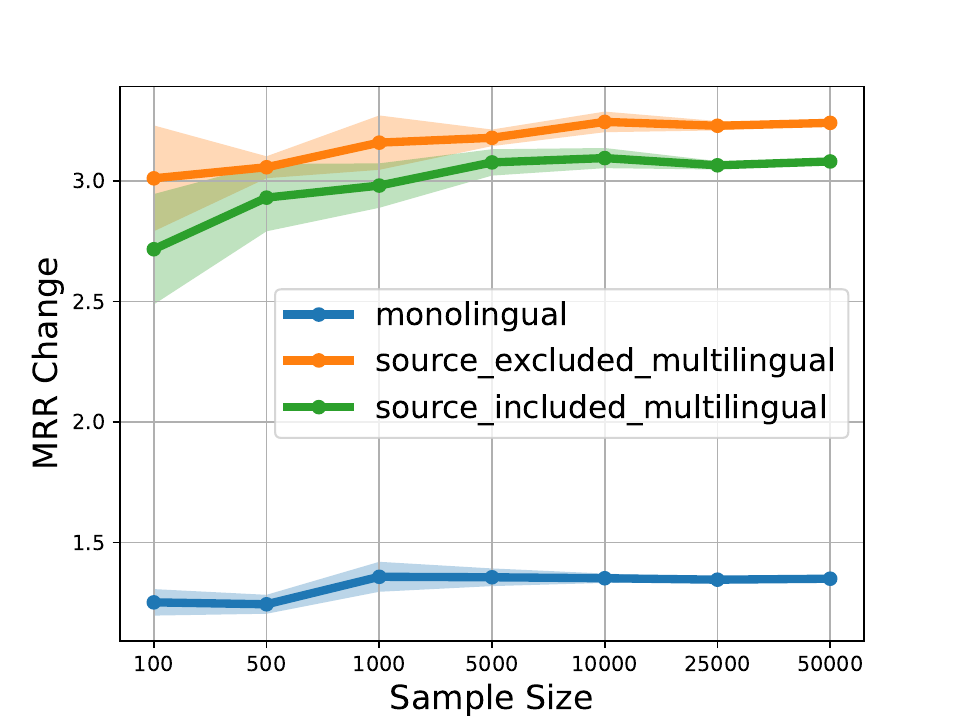}
\subcaption{StarEncoder}    \label{fig:LRD_abalation_starencoder}
  \end{subfigure}
  \begin{subfigure}[b]{0.19\linewidth}
    \centering
    \includegraphics[width=\linewidth]{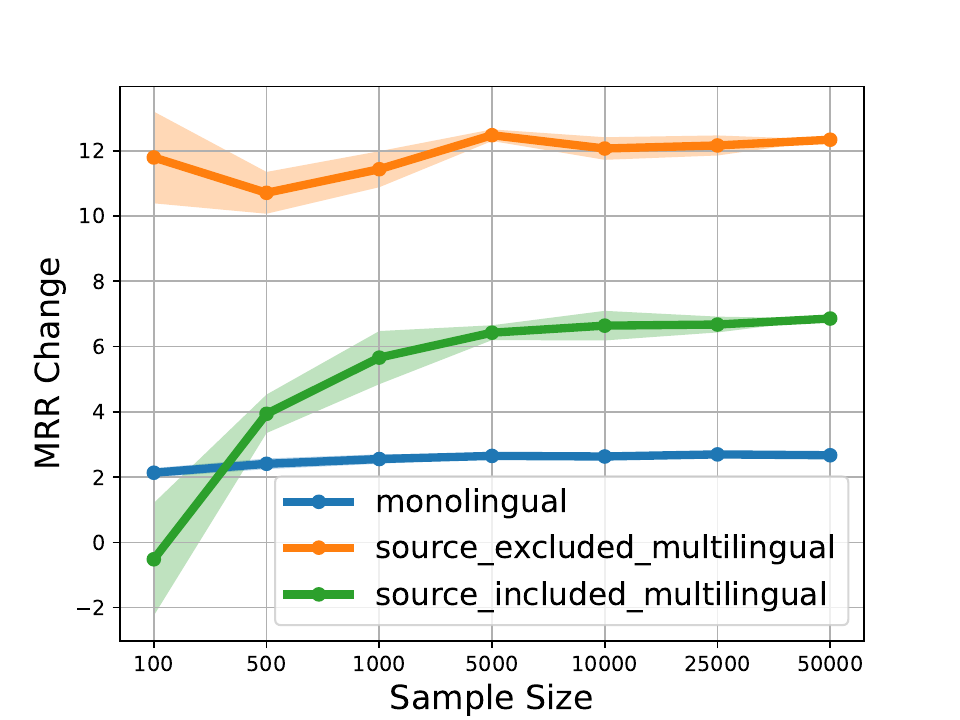}
    \subcaption{CodeT5+}
    \label{fig:LRD_abalation_codet5+}
  \end{subfigure}

  \begin{subfigure}[b]{0.19\linewidth}
    \centering
    \includegraphics[width=\linewidth]{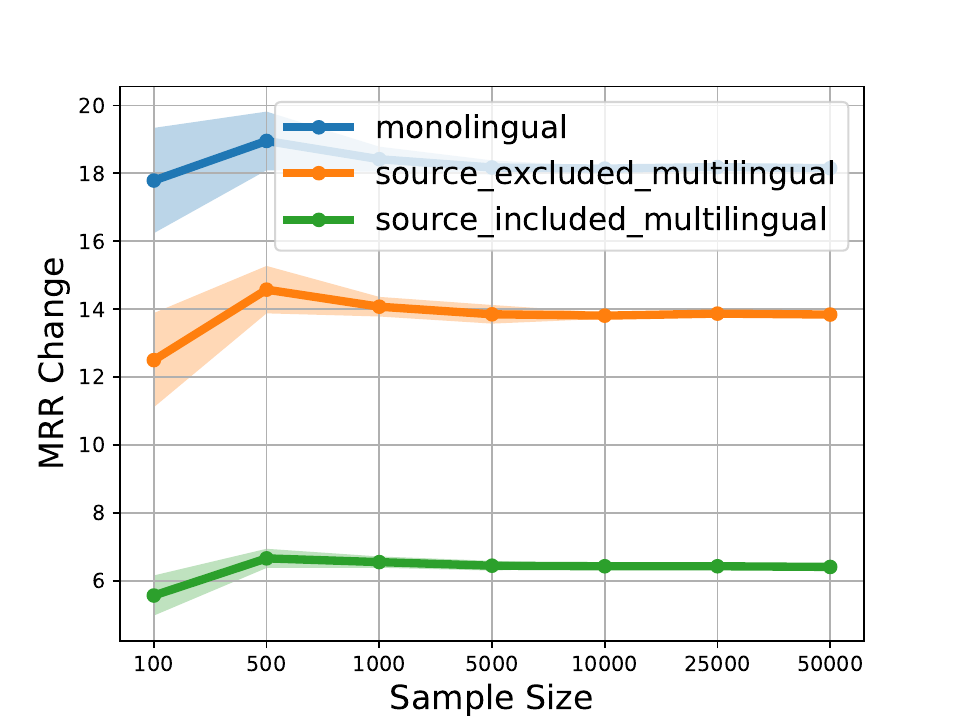}
    \subcaption{CodeBERT}
    \label{fig:CS-LRD_abalation_codebert}
  \end{subfigure}
  \begin{subfigure}[b]{0.19\linewidth}
    \centering
    \includegraphics[width=\linewidth]{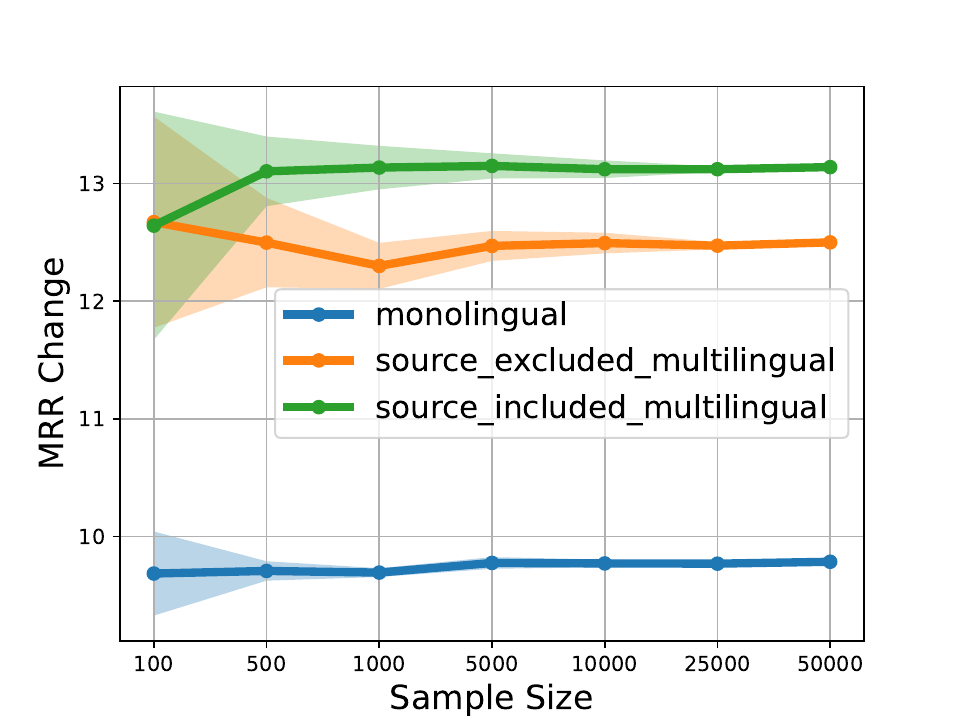}
    \subcaption{GraphCodeBert}
   \label{CS-LRD_abalation_graphcodebert}
  \end{subfigure}
  \begin{subfigure}[b]{0.19\linewidth}
    \centering
    \includegraphics[width=\linewidth]{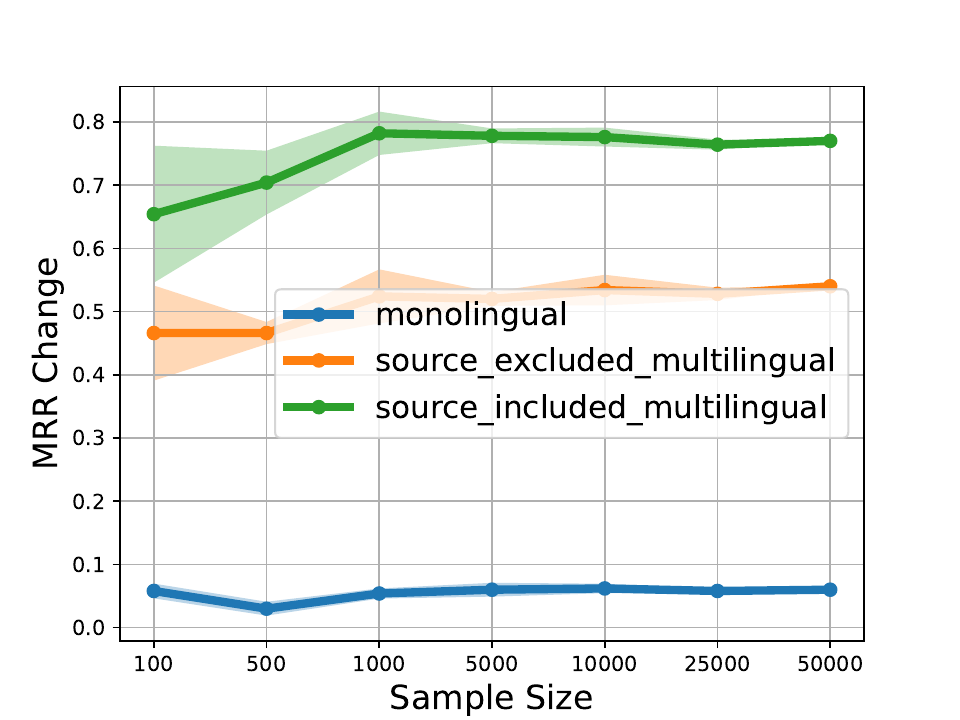}
    \subcaption{UnixCoder}
    \label{fig:CS-LRD_abalation_unixcoder}
  \end{subfigure}
  \begin{subfigure}[b]{0.19\linewidth}
    \centering
    \includegraphics[width=\linewidth]{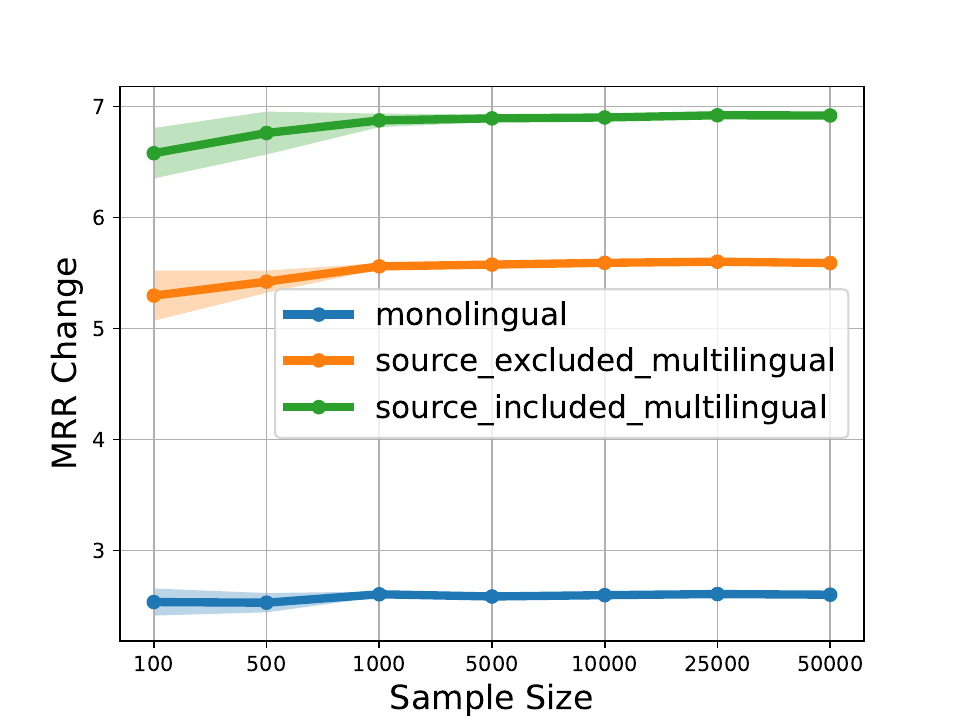}
    \subcaption{StarEncoder}
    \label{fig:CS-LRD_abalation_starencoder}
  \end{subfigure}
  \begin{subfigure}[b]{0.19\linewidth}
    \centering
    \includegraphics[width=\linewidth]{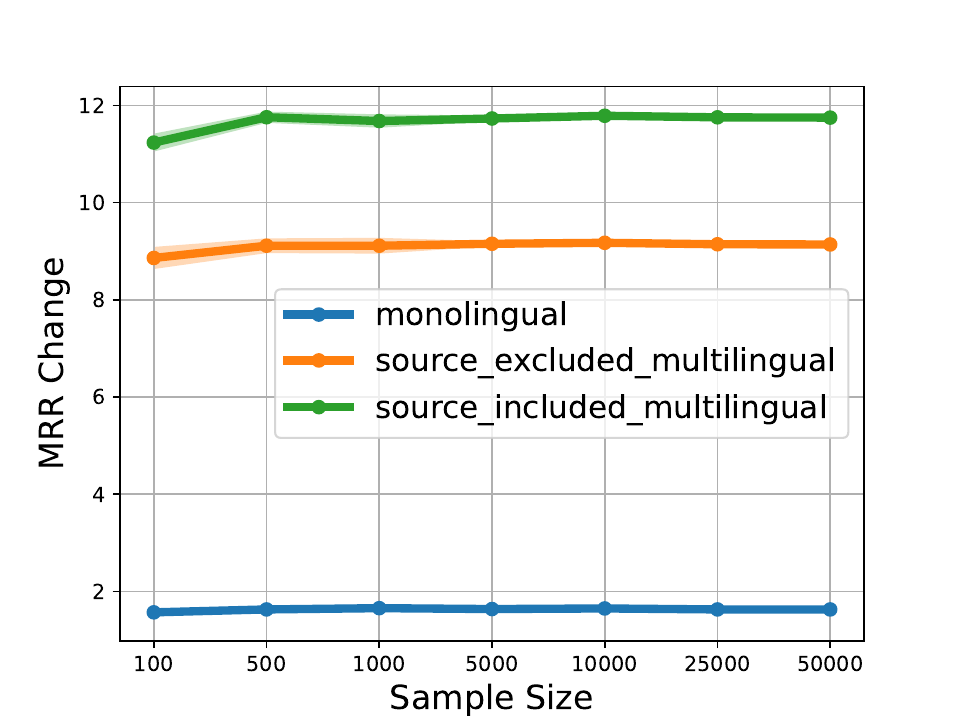}
    \subcaption{CodeT5+}
    \label{fig:CS-LRD_abalation_codet5+}
  \end{subfigure}

\caption{Impact of Estimation Set size on MRR change, with the top, middle, and bottom rows showing effects for Centering, LRD, and CS-LRD, respectively.}
\label{fig:c2c_zeroshot_abalation_alignment}
\end{figure*}

%% file: plots/c2c_zeroshot_abalation_pooling.tex
\begin{figure*}[htb!]
  \centering
\begin{subfigure}[b]{0.245\linewidth}
    \centering
    \includegraphics[width=\linewidth]{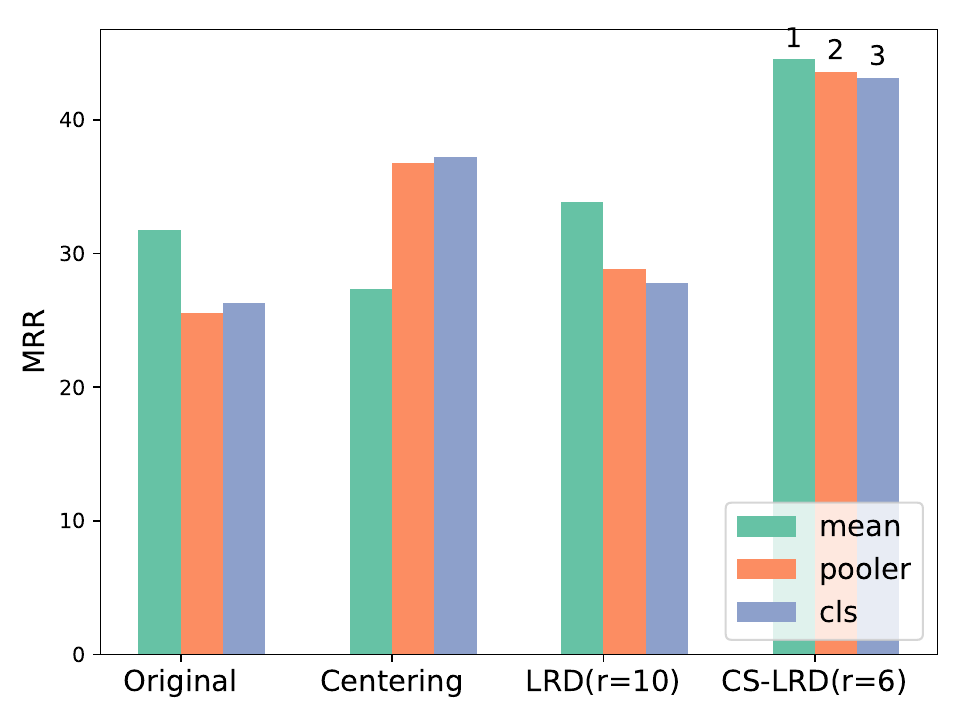}
    \subcaption{CodeBERT}
    \label{fig:pooler_abalation_codebert}
\end{subfigure}
\begin{subfigure}[b]{0.245\linewidth}
    \centering
    \includegraphics[width=\linewidth]{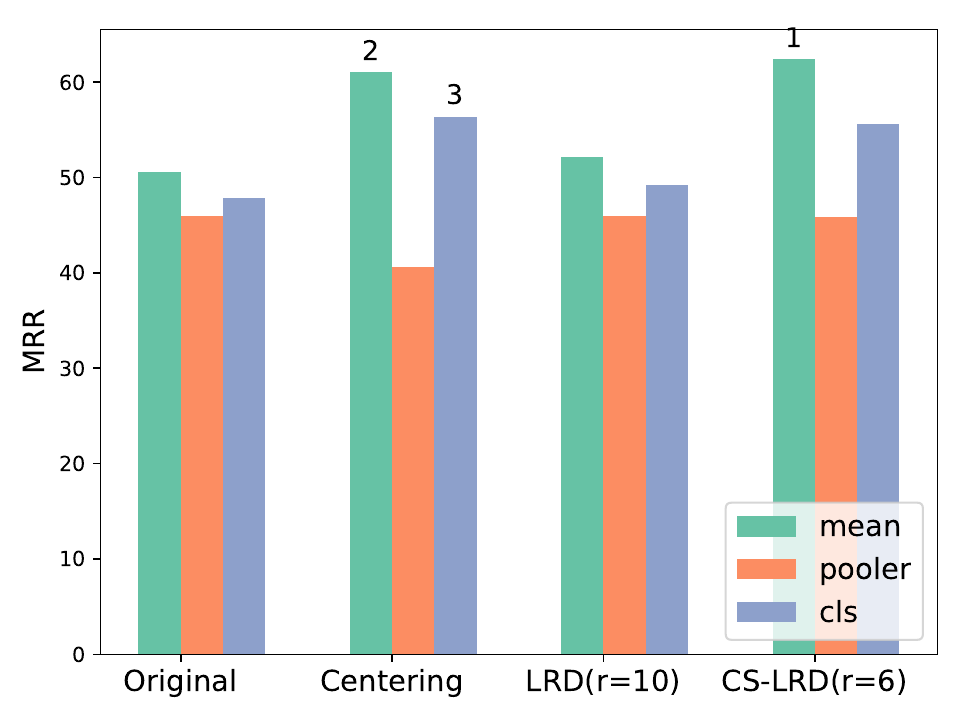}
    \subcaption{GraphCodeBERT}
    \label{fig:pooler_abalation_graphcodebert}
\end{subfigure}
\begin{subfigure}[b]{0.245\linewidth}
    \centering
    \includegraphics[width=\linewidth]{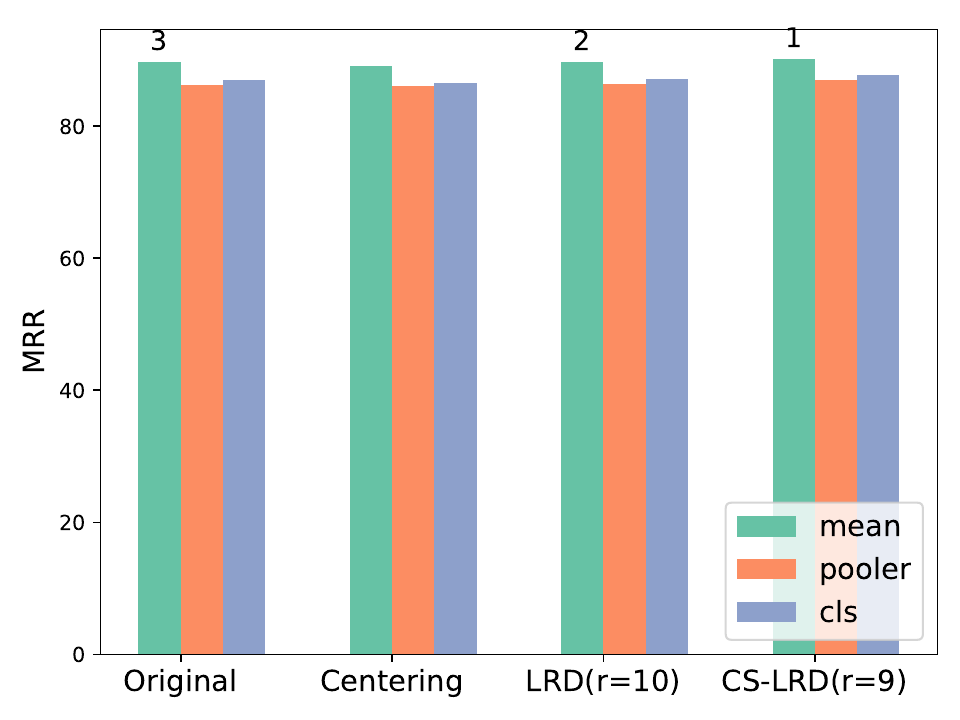}
    \subcaption{UnixCoder}   \label{fig:pooler_abalation_unixcoder}
\end{subfigure}
\begin{subfigure}[b]{0.245\linewidth}
    \centering
    \includegraphics[width=\linewidth]{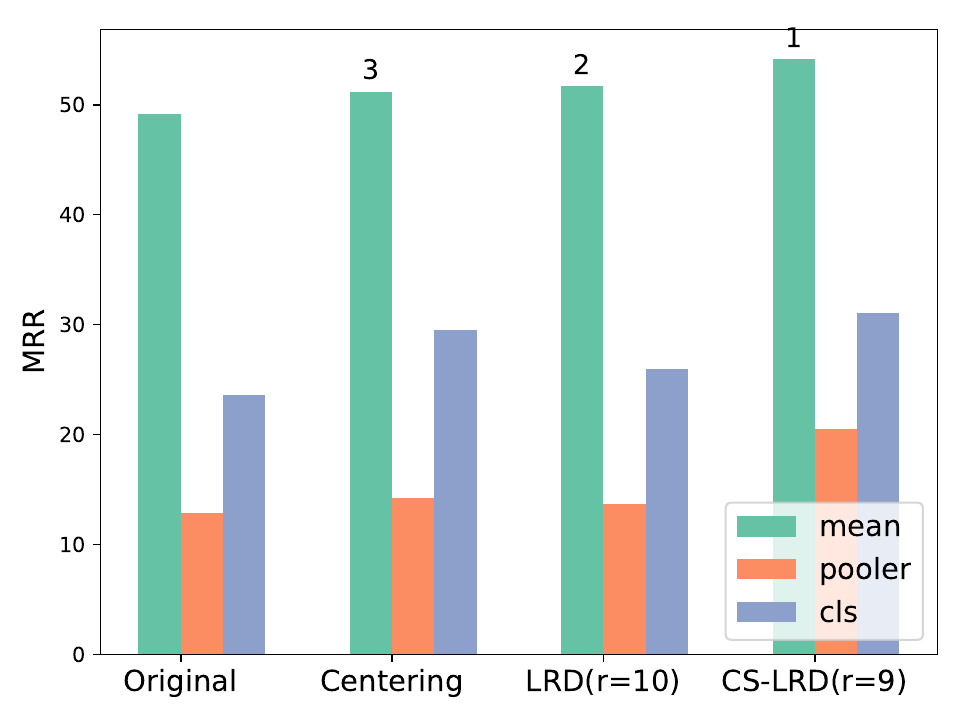}
    \subcaption{StarEncoder}
    \label{fig:pooler_abalation_starencoder}
\end{subfigure}
\caption{The figure presents the averaged Mean Reciprocal Rank (MRR) across three retrieval setups for zero-shot Code2Code search. These results are derived from mean, cls, and pooler embeddings. Annotations highlight the top three values in both the original and after removing language components settings for various pooling strategies.}
\label{fig:c2c_zeroshot_abalation_pooling}
\end{figure*}

%% file: plots/c2c/c2c_finetuned.tex
\begin{figure*}[htbp!]
  \centering

  \begin{subfigure}[b]{0.19\linewidth}
    \centering
    \includegraphics[width=\linewidth]{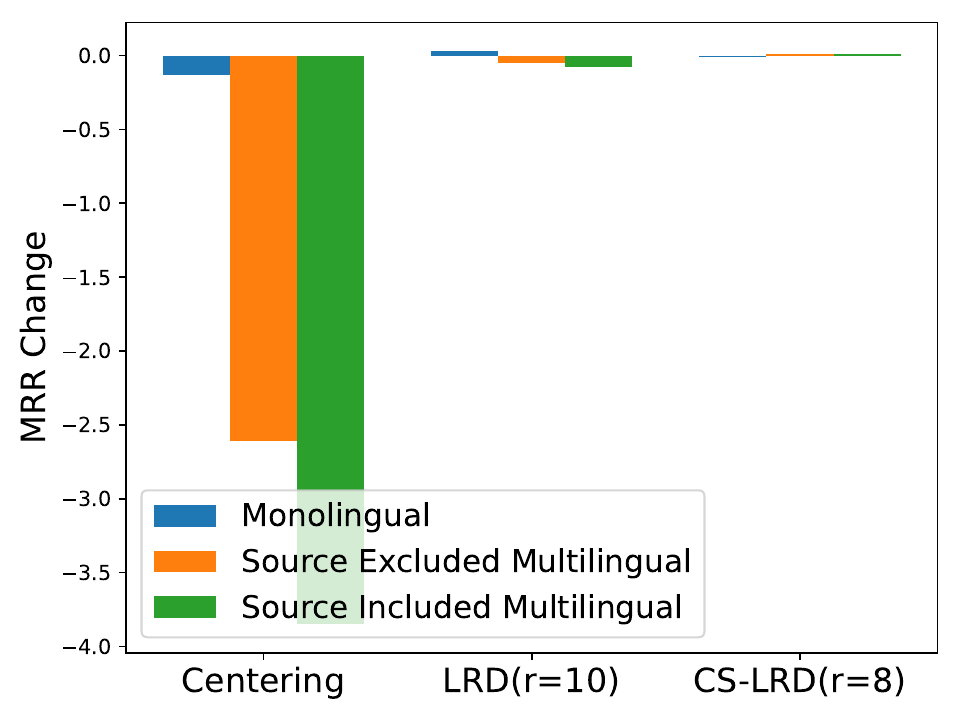}
    \subcaption{CodeBERT}
    \label{fig:c2c_finetuned_codebert_finetunedmean-5}
  \end{subfigure}
  \begin{subfigure}[b]{0.19\linewidth}
    \centering
    \includegraphics[width=\linewidth]{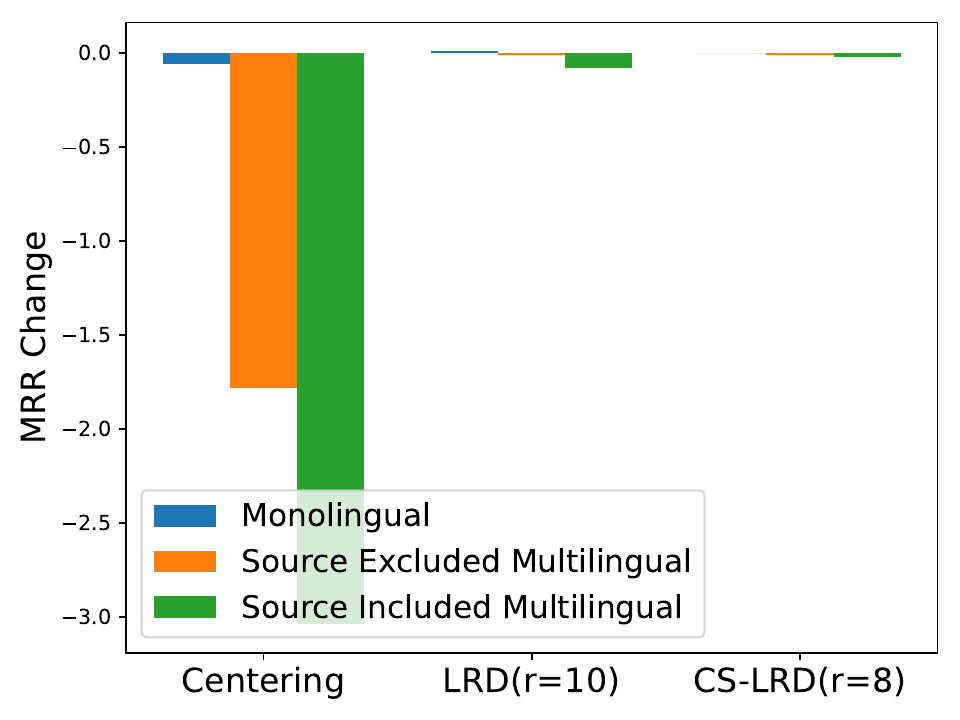}
    \subcaption{GraphCodeBERT}
    \label{fig:c2c_finetuned_graphcodebert_finetunedmean-5}
  \end{subfigure}
  \begin{subfigure}[b]{0.19\linewidth}
    \centering
    \includegraphics[width=\linewidth]{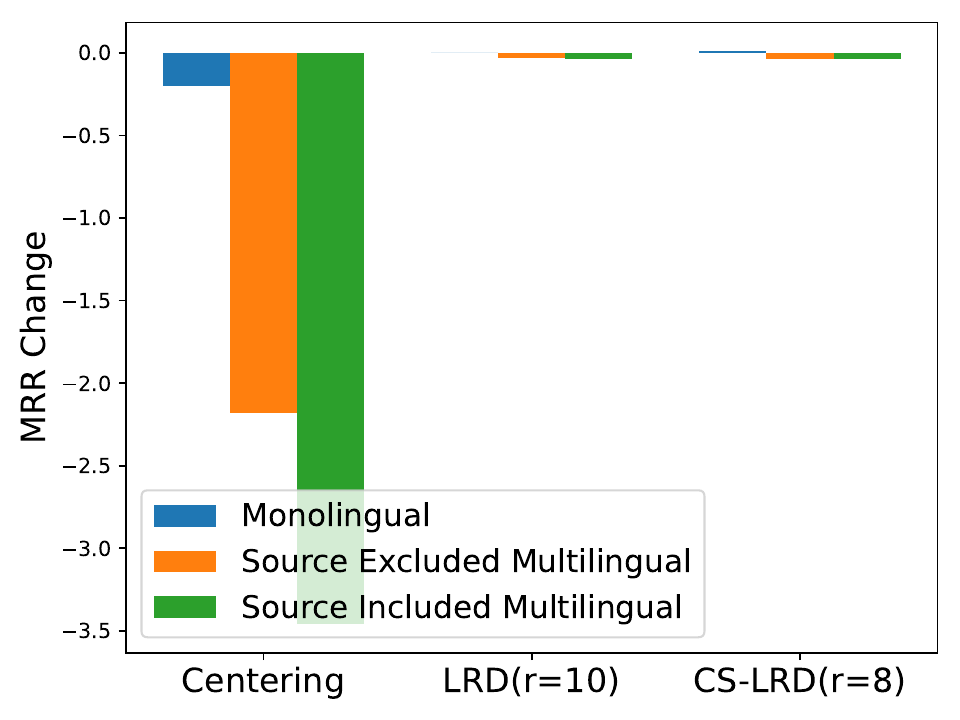}
    \subcaption{UnixCoder}
    \label{fig:c2c_finetuned_unixcoder_finetunedmean-5}
  \end{subfigure}
  \begin{subfigure}[b]{0.19\linewidth}
    \centering
    \includegraphics[width=\linewidth]{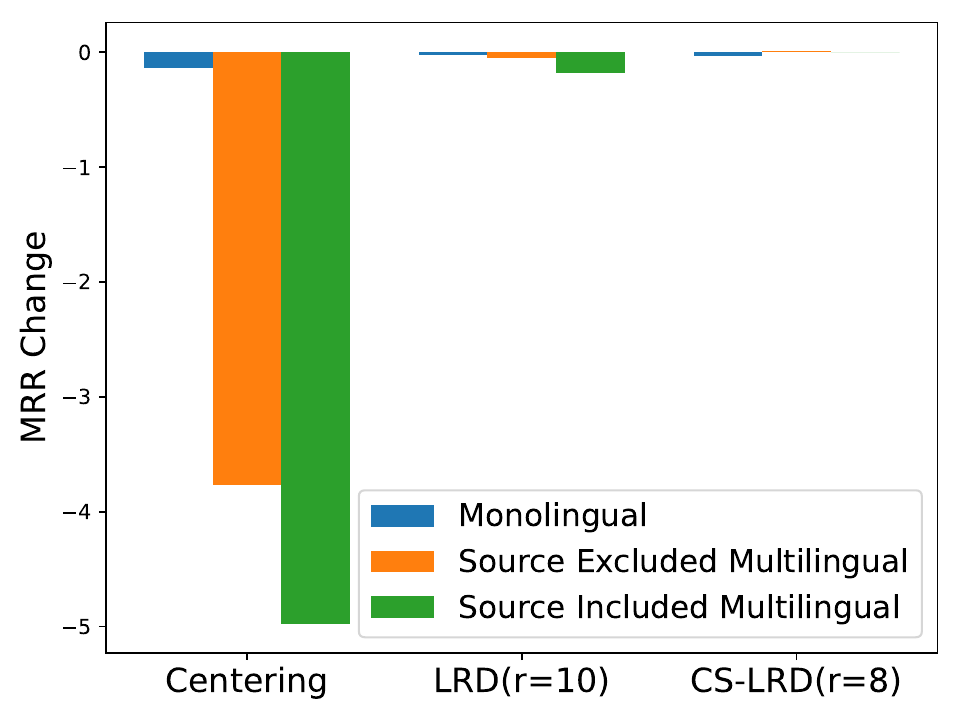}
    \subcaption{StarEncoder}
    \label{fig:c2c_finetuned_starencoder_finetunedmean-}
  \end{subfigure}
  \begin{subfigure}[b]{0.19\linewidth}
    \centering
    \includegraphics[width=\linewidth]{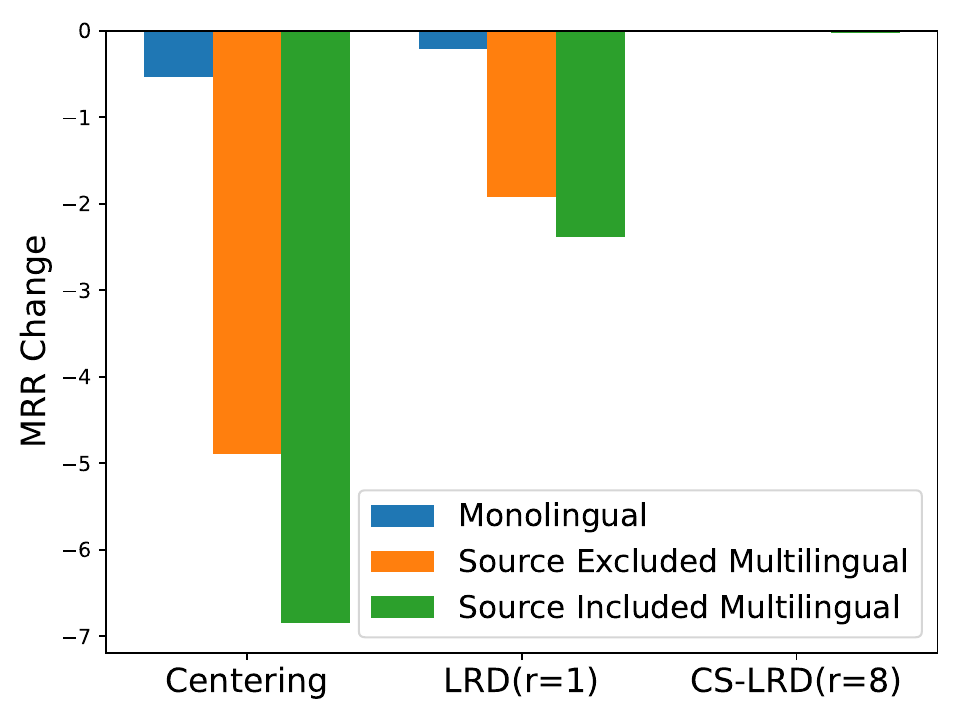}
    \subcaption{CodeT5+}
    \label{fig:c2c_finetuned_codet5+_finetunedmean-5}
  \end{subfigure}
  \caption{Absolute change in Mean Reciprocal Rank (MRR) after removing language components for Code2Code search after contrastive fine-tuning.}
  \label{fig:c2c_finetuned}
\end{figure*}

%% file: plots/t2c/t2c_finetuned.tex
\begin{figure*}[]
  \centering

  \begin{subfigure}[b]{0.19\linewidth}
    \centering
    \includegraphics[width=\linewidth]{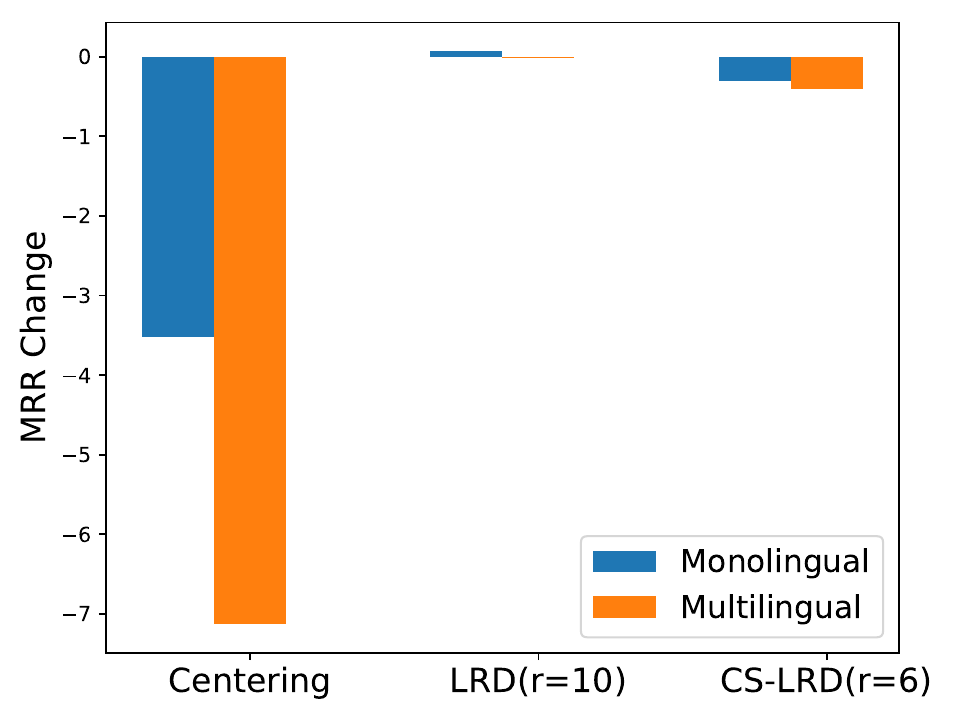}
    \subcaption{CodeBERT}
    \label{fig:t2c_finetuned_codebert_finetunedmean}
  \end{subfigure}
  \begin{subfigure}[b]{0.19\linewidth}
    \centering
    \includegraphics[width=\linewidth]{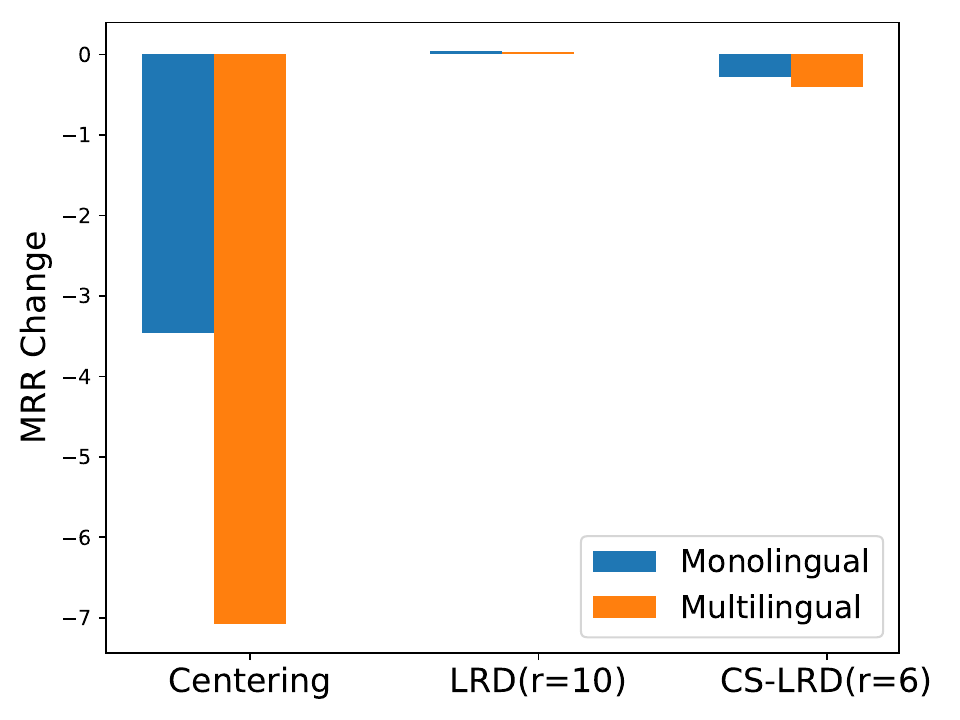}
    \subcaption{GraphCodeBERT}
\label{fig:t2c_finetuned_graphcodebert_finetunedmean}
  \end{subfigure}
  \begin{subfigure}[b]{0.19\linewidth}
    \centering
    \includegraphics[width=\linewidth]{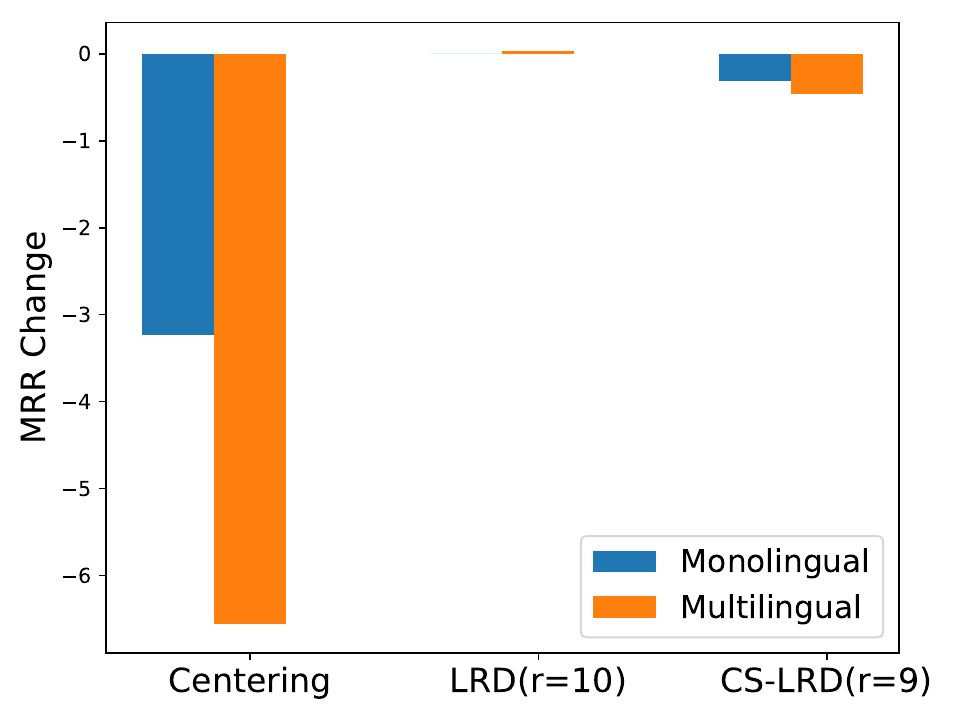}
    \subcaption{UnixCoder}
\label{fig:t2c_finetuned_unixcoder_finetunedmean}
  \end{subfigure}
  \begin{subfigure}[b]{0.19\linewidth}
    \centering
    \includegraphics[width=\linewidth]{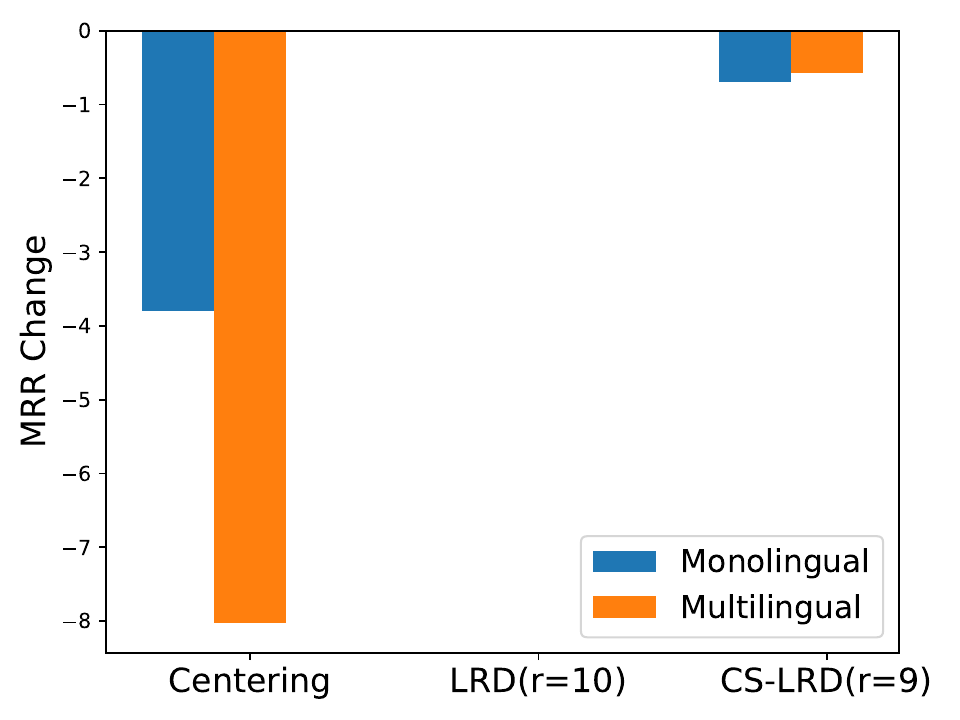}
    \subcaption{StarEncoder}
\label{fig:t2c_finetuned_starencoder_finetunedmean}
  \end{subfigure}
  \begin{subfigure}[b]{0.19\linewidth}
    \centering
    \includegraphics[width=\linewidth]{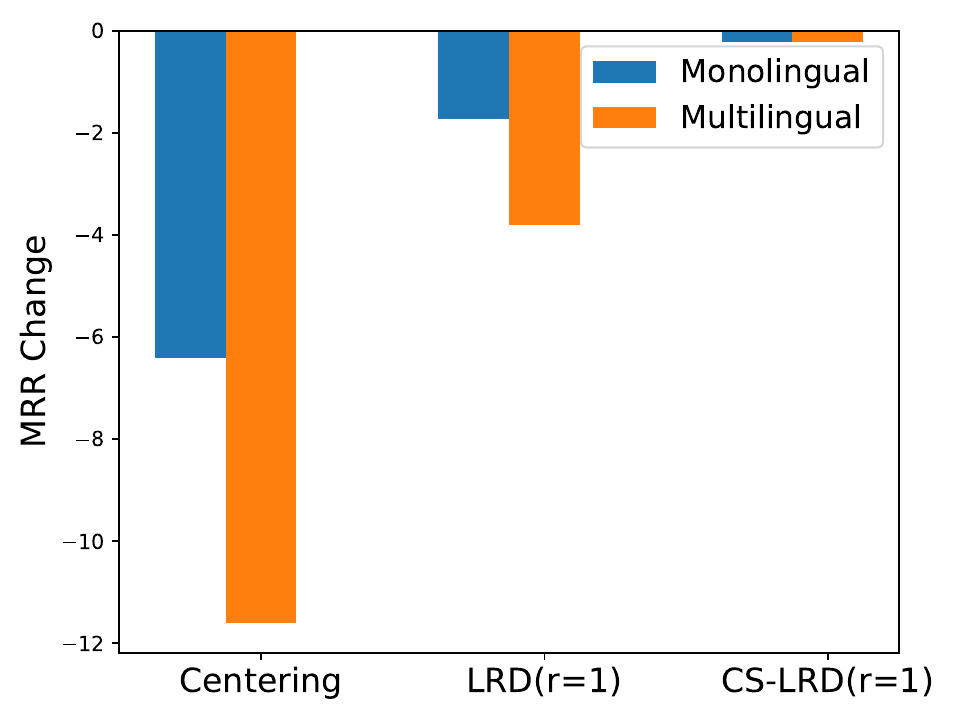}
    \subcaption{CodeT5+}
    \label{fig:t2c_finetuned_codet5+_finetunedpooler}
  \end{subfigure}
  \caption{Absolute change in Mean Reciprocal Rank (MRR) after removing language components for Text2Code search after contrastive fine-tuning.}
  \label{fig:t2c_finetuned}
\end{figure*}

%% file: tables/c2c/zeroshot/c2c_codebert_mean.tex
\begin{tabular}{c|c|ccccc}
\hline
CodeBERT(mean)                                &             & Java  & Javascript & PHP   & Python & Avg.                    \\ \hline
\multirow{4}{*}{Monolingual}                  & Original    & 46.90 & 56.75      & 57.89 & 43.19  & 51.18                   \\
                                              & Centering   & 55.25 & 47.29      & 43.93 & 33.32  & 44.95 (\textbf{-6.23})  \\
                                              & LRD(r=10)   & 49.49 & 59.56      & 60.87 & 45.83  & 53.94 (\textbf{+2.76})  \\
                                              & CS-LRD(r=6) & 68.70 & 75.91      & 76.35 & 56.35  & 69.33 (\textbf{+18.15}) \\ \hline
\multirow{4}{*}{Source Excluded Multilingual} & Original    & 36.78 & 39.92      & 44.76 & 31.37  & 38.21                   \\
                                              & Centering   & 41.37 & 31.93      & 29.15 & 24.44  & 31.72 (\textbf{-6.49})  \\
                                              & LRD(r=10)   & 38.97 & 43.62      & 46.82 & 33.56  & 40.74 (\textbf{+2.53})  \\
                                              & CS-LRD(r=6) & 49.06 & 60.58      & 56.83 & 41.74  & 52.05 (\textbf{+13.84}) \\ \hline
\multirow{4}{*}{Source Included Multilingual} & Original    & 4.09  & 6.11       & 8.01  & 5.34   & 5.89                    \\
                                              & Centering   & 2.06  & 5.24       & 8.18  & 5.76   & 5.31 (\textbf{-0.58})   \\
                                              & LRD(r=10)   & 4.80  & 7.25       & 9.25  & 6.12   & 6.86 (\textbf{+0.97})   \\
                                              & CS-LRD(r=6) & 7.41  & 15.25      & 15.35 & 11.21  & 12.30 (\textbf{+6.41})  \\ \hline
\end{tabular}%

%% file: tables/c2c/zeroshot/c2c_codebert_cls.tex
\begin{tabular}{c|c|ccccc}
\hline
CodeBERT (cls)                                &             & Java  & Javascript & PHP   & Python & Avg.                    \\ \hline
\multirow{4}{*}{Monolingual}                  & Original    & 51.34 & 54.02      & 57.42 & 26.15  & 47.23                   \\
                                              & Centering   & 64.16 & 62.21      & 71.11 & 53.86  & 62.84 (\textbf{+15.61}) \\
                                              & LRD(r=10)   & 53.02 & 55.63      & 59.52 & 28.43  & 49.15 (\textbf{+1.92})  \\
                                              & CS-LRD(r=6) & 77.72 & 78.42      & 76.87 & 60.80  & 73.45 (\textbf{+26.22}) \\ \hline
\multirow{4}{*}{Source Excluded Multilingual} & Original    & 34.28 & 28.55      & 44.16 & 14.43  & 30.35                   \\
                                              & Centering   & 48.48 & 46.28      & 47.32 & 37.69  & 44.94 (\textbf{+14.59}) \\
                                              & LRD(r=10)   & 36.25 & 31.63      & 46.86 & 16.05  & 32.70 (\textbf{+2.35})  \\
                                              & CS-LRD(r=6) & 48.14 & 60.89      & 52.38 & 41.43  & 50.71 (\textbf{+20.36}) \\ \hline
\multirow{4}{*}{Source Included Multilingual} & Original    & 1.11  & 1.89       & 1.23  & 1.38   & 1.4                     \\
                                              & Centering   & 2.69  & 3.00       & 5.30  & 4.38   & 3.84 (\textbf{+2.44})   \\
                                              & LRD(r=10)   & 1.29  & 2.14       & 1.40  & 1.60   & 1.61 (\textbf{+0.21})   \\
                                              & CS-LRD(r=6) & 4.08  & 6.61       & 6.11  & 4.67   & 5.37 (\textbf{+3.97})   \\ \hline
\end{tabular}%

%% file: tables/c2c/zeroshot/c2c_codebert_pooler.tex
\begin{tabular}{c|c|ccccc}
\hline
CodeBERT (pooler)                                &             & Java  & Javascript & PHP   & Python & Avg.                    \\ \hline
\multirow{4}{*}{Monolingual}                  & Original    & 49.11 & 51.24      & 56.12 & 23.23  & 44.92                   \\
                                              & Centering   & 64.12 & 61.88      & 70.77 & 48.81  & 61.40 (\textbf{+16.48}) \\
                                              & LRD(r=10)   & 52.46 & 55.89      & 59.83 & 28.64  & 49.20 (\textbf{+4.28})  \\
                                              & CS-LRD(r=6) & 77.53 & 77.39      & 75.30 & 61.49  & 72.93 (\textbf{+28.01}) \\ \hline
\multirow{4}{*}{Source Excluded Multilingual} & Original    & 36.60 & 29.57      & 41.70 & 12.96  & 30.21                   \\
                                              & Centering   & 50.17 & 45.56      & 47.41 & 35.03  & 44.54 (\textbf{+14.33}) \\
                                              & LRD(r=10)   & 40.31 & 36.32      & 47.63 & 17.00  & 35.32 (\textbf{+5.11})  \\
                                              & CS-LRD(r=6) & 53.23 & 59.81      & 52.07 & 42.85  & 51.99 (\textbf{+21.78}) \\ \hline
\multirow{4}{*}{Source Included Multilingual} & Original    & 1.27  & 1.92       & 1.62  & 1.43   & 1.56                    \\
                                              & Centering   & 3.39  & 3.29       & 6.45  & 4.39   & 4.38 (\textbf{+2.82})   \\
                                              & LRD(r=10)   & 1.62  & 2.67       & 2.09  & 1.79   & 2.04 (\textbf{+0.48})   \\
                                              & CS-LRD(r=6) & 4.19  & 7.05       & 7.49  & 4.95   & 5.92 (\textbf{+4.36})   \\ \hline
\end{tabular}%

%% file: tables/c2c/zeroshot/c2c_graphcodebert_mean.tex
\begin{tabular}{c|c|ccccc}
\hline
GraphCodeBERT (mean)                          &             & Java  & Javascript & PHP   & Python & Avg.                    \\ \hline
\multirow{4}{*}{Monolingual}                  & Original    & 72.56 & 82.39      & 85.85 & 91.55  & 83.09                   \\
                                              & Centering   & 92.75 & 92.50      & 94.20 & 96.53  & 94.00 (\textbf{+10.91}) \\
                                              & LRD(r=10)   & 74.47 & 84.03      & 87.35 & 92.17  & 84.50 (\textbf{+1.41})  \\
                                              & CS-LRD(r=6) & 90.47 & 92.36      & 93.47 & 95.21  & 92.88 (\textbf{+9.79})  \\ \hline
\multirow{4}{*}{Source Excluded Multilingual} & Original    & 45.57 & 66.15      & 61.95 & 68.24  & 60.48                   \\
                                              & Centering   & 81.51 & 83.49      & 66.08 & 84.28  & 78.84 (\textbf{+18.36}) \\
                                              & LRD(r=10)   & 47.80 & 68.35      & 62.83 & 69.54  & 62.13 (\textbf{+1.65})  \\
                                              & CS-LRD(r=6) & 66.78 & 78.46      & 70.22 & 76.46  & 72.98 (\textbf{+12.50}) \\ \hline
\multirow{4}{*}{Source Included Multilingual} & Original    & 5.09  & 10.63      & 5.59  & 11.59  & 8.23                    \\
                                              & Centering   & 2.67  & 11.66      & 9.61  & 17.28  & 10.30 (\textbf{+2.07})  \\
                                              & LRD(r=10)   & 6.20  & 12.89      & 6.66  & 13.42  & 9.79 (\textbf{+1.56})   \\
                                              & CS-LRD(r=6) & 12.88 & 27.92      & 16.25 & 28.42  & 21.37 (\textbf{+13.14}) \\ \hline
\end{tabular}%

%% file: tables/c2c/zeroshot/c2c_graphcodebert_cls.tex
\begin{tabular}{c|cccccc}
\hline
GraphCodeBERT (cls)                           &             & Java  & Javascript & PHP   & Python & Avg.                    \\ \hline
\multirow{4}{*}{Monolingual}                  & Original    & 62.00 & 82.02      & 78.29 & 71.09  & 73.35                   \\
                                              & Centering   & 80.86 & 90.83      & 83.13 & 81.92  & 84.18 (\textbf{+10.83}) \\
                                              & LRD(r=10)   & 63.56 & 83.54      & 79.47 & 71.78  & 74.59 (\textbf{+1.24})  \\
                                              & CS-LRD(r=6) & 72.81 & 87.49      & 84.80 & 73.72  & 79.71 (\textbf{+6.36})  \\ \hline
\multirow{4}{*}{Source Excluded Multilingual} & Original    & 52.12 & 63.03      & 42.00 & 62.08  & 54.81                   \\
                                              & Centering   & 70.88 & 72.83      & 41.10 & 70.51  & 63.83 (\textbf{+9.02})  \\
                                              & LRD(r=10)   & 53.93 & 64.47      & 42.79 & 63.02  & 56.05 (\textbf{+1.24})  \\
                                              & CS-LRD(r=6) & 64.44 & 71.68      & 46.04 & 65.87  & 62.01 (\textbf{+7.20})  \\ \hline
\multirow{4}{*}{Source Included Multilingual} & Original    & 7.94  & 25.25      & 15.02 & 13.36  & 15.39                   \\
                                              & Centering   & 15.57 & 28.44      & 16.53 & 23.70  & 21.06 (\textbf{+5.67})  \\
                                              & LRD(r=10)   & 9.29  & 27.03      & 16.33 & 15.08  & 16.93 (\textbf{+1.54})  \\
                                              & CS-LRD(r=6) & 17.34 & 35.19      & 21.38 & 26.41  & 25.08 (\textbf{+9.69})  \\ \hline
\end{tabular}%

%% file: tables/c2c/zeroshot/c2c_graphcodebert_pooler.tex
\begin{tabular}{c|c|ccccc}
\hline
GraphCodeBERT (pooler)                        &             & Java  & Javascript & PHP   & Python & Avg.                   \\ \hline
\multirow{4}{*}{Monolingual}                  & Original    & 60.64 & 80.56      & 73.22 & 67.14  & 70.39                  \\
                                              & Centering   & 58.20 & 78.84      & 70.94 & 68.08  & 69.02 (\textbf{-1.37}) \\
                                              & LRD(r=10)   & 60.52 & 80.62      & 73.47 & 66.94  & 70.39 (\textbf{+0.00}) \\
                                              & CS-LRD(r=6) & 60.50 & 80.24      & 72.73 & 67.21  & 70.17 (\textbf{-0.22}) \\ \hline
\multirow{4}{*}{Source Excluded Multilingual} & Original    & 51.41 & 60.75      & 40.34 & 59.03  & 52.88                  \\
                                              & Centering   & 47.27 & 58.01      & 37.17 & 58.06  & 50.13 (\textbf{-2.75}) \\
                                              & LRD(r=10)   & 51.23 & 60.75      & 40.51 & 58.91  & 52.85 (\textbf{-0.03}) \\
                                              & CS-LRD(r=6) & 51.20 & 60.57      & 40.14 & 59.06  & 52.74 (\textbf{-0.14}) \\ \hline
\multirow{4}{*}{Source Included Multilingual} & Original    & 7.32  & 24.18      & 14.81 & 12.74  & 14.76                  \\
                                              & Centering   & 0.81  & 4.70       & 2.55  & 2.27   & 2.58 (\textbf{-12.18}) \\
                                              & LRD(r=10)   & 7.27  & 23.83      & 14.60 & 12.56  & 14.56 (\textbf{-0.20}) \\
                                              & CS-LRD(r=6) & 7.30  & 24.01      & 14.64 & 12.71  & 14.66 (\textbf{-0.10}) \\ \hline
\end{tabular}%

%% file: tables/c2c/zeroshot/c2c_starencoder_mean.tex
\begin{tabular}{c|c|cccccccc}
\hline
StarEncoder (mean) &             & C     & C\#   & C++   & Java  & Javascript & PHP   & Python & Avg.                   \\ \hline
\multirow{4}{*}{Monolingual}                  & Original & 20.27 & 91.66 & 86.45 & 90.11 & 90.28 & 90.37 & 90.46 & 79.94 \\
                   & Centering   & 76.68 & 90.04 & 93.28 & 89.51 & 89.16      & 93.12 & 93.63  & 89.35 (\textbf{+9.41}) \\
                   & LRD(r=10)   & 21.95 & 92.28 & 88.64 & 91.18 & 91.39      & 91.31 & 92.26  & 81.29 (\textbf{+1.35}) \\
                   & CS-LRD(r=9) & 21.30 & 93.91 & 90.58 & 92.74 & 91.52      & 93.46 & 94.26  & 82.54 (\textbf{+2.60}) \\ \hline
\multirow{4}{*}{Source Excluded Multilingual} & Original & 8.26  & 38.03 & 54.99 & 39.17 & 61.52 & 69.98 & 61.46 & 47.63 \\
                   & Centering   & 15.30 & 39.10 & 58.78 & 41.94 & 61.10      & 67.43 & 67.20  & 50.12 (\textbf{+2.49}) \\
                   & LRD(r=10)   & 8.77  & 40.97 & 59.98 & 41.74 & 65.03      & 72.09 & 67.48  & 50.87 (\textbf{+3.24}) \\
                   & CS-LRD(r=9) & 8.79  & 44.02 & 62.24 & 43.93 & 65.51      & 75.72 & 72.30  & 53.22 (\textbf{+5.59}) \\ \hline
\multirow{4}{*}{Source Included Multilingual} & Original & 7.01  & 27.61 & 20.61 & 29.83 & 28.61 & 5.76  & 18.88 & 19.76 \\
                   & Centering   & 6.67  & 16.52 & 21.35 & 19.71 & 13.44      & 2.66  & 17.76  & 14.02 (\textbf{-5.74}) \\
                   & LRD(r=10)   & 7.52  & 30.09 & 25.46 & 32.07 & 32.80      & 7.82  & 24.13  & 22.84 (\textbf{+3.08}) \\
                   & CS-LRD(r=9) & 7.77  & 33.15 & 31.06 & 34.78 & 35.87      & 10.38 & 33.70  & 26.67 (\textbf{+6.91}) \\ \hline
\end{tabular}%

%% file: tables/c2c/zeroshot/c2c_starencoder_cls.tex
\begin{tabular}{c|c|cccccccc}
\hline
StarEncoder (cls) &             & C     & C\#   & C++   & Java  & Javascript & PHP   & Python & Avg.                    \\ \hline
\multirow{4}{*}{Monolingual}                  & Original & 8.35 & 50.35 & 37.75 & 46.48 & 58.53 & 59.99 & 59.21 & 45.81 \\
                  & Centering   & 48.76 & 57.74 & 65.89 & 52.88 & 67.19      & 61.13 & 65.81  & 59.91 (\textbf{+14.10}) \\
                  & LRD(r=10)   & 8.90  & 54.66 & 44.38 & 50.47 & 61.94      & 64.84 & 64.84  & 50.00 (\textbf{+4.19})  \\
                  & CS-LRD(r=9) & 10.34 & 64.56 & 57.82 & 61.20 & 69.79      & 68.80 & 77.70  & 58.60 (\textbf{+12.79}) \\ \hline
\multirow{4}{*}{Source Excluded Multilingual} & Original & 3.92 & 16.92 & 12.56 & 17.26 & 29.98 & 32.14 & 25.54 & 19.76 \\
                  & Centering   & 8.58  & 17.17 & 26.63 & 17.75 & 35.76      & 32.17 & 27.17  & 23.60 (\textbf{+3.84})  \\
                  & LRD(r=10)   & 4.23  & 17.18 & 15.93 & 17.58 & 33.98      & 36.22 & 29.92  & 22.15 (\textbf{+2.39})  \\
                  & CS-LRD(r=9) & 5.07  & 17.83 & 23.96 & 18.04 & 43.05      & 42.13 & 43.27  & 27.62 (\textbf{+7.86})  \\ \hline
\multirow{4}{*}{Source Included Multilingual} & Original & 2.93 & 10.92 & 2.47  & 11.43 & 2.71  & 0.93  & 5.62  & 5.29  \\
                  & Centering   & 3.88  & 9.07  & 3.05  & 9.92  & 2.76       & 0.25  & 5.39   & 4.90 (\textbf{-0.39})   \\
                  & LRD(r=10)   & 3.33  & 11.23 & 2.75  & 11.87 & 3.41       & 1.21  & 7.16   & 5.85 (\textbf{+0.56})   \\
                  & CS-LRD(r=9) & 3.89  & 12.33 & 3.24  & 13.11 & 4.76       & 1.53  & 10.09  & 6.99 (\textbf{+1.70})   \\ \hline
\end{tabular}

%% file: tables/c2c/zeroshot/c2c_starencoder_pooler.tex
\begin{tabular}{c|c|cccccccc}
\hline
StarEncoder (pooler) &             & C    & C\#    & C++   & Java  & Javascript & PHP   & Python & Avg.                    \\ \hline
\multirow{4}{*}{Monolingual}                  & Original & 3.99 & 27.80 & 11.77 & 28.98 & 30.90 & 21.52 & 30.14 & 22.16 \\
                     & Centering   & 6.80 & 33.20 & 25.15 & 30.42 & 37.13      & 24.72 & 25.24  & 26.09 (\textbf{+3.93})  \\
                     & LRD(r=10)   & 4.08 & 29.15 & 12.64 & 30.13 & 33.07      & 23.68 & 32.21  & 23.57 (\textbf{+1.41})  \\
                     & CS-LRD(r=9) & 5.98 & 43.66 & 25.21 & 43.38 & 49.81      & 40.37 & 50.78  & 37.03 (\textbf{+14.87}) \\ \hline
\multirow{4}{*}{Source Excluded Multilingual} & Original & 2.94 & 14.49 & 7.93  & 15.52 & 16.19 & 10.41 & 15.50 & 11.85 \\
                     & Centering   & 3.53 & 15.68 & 8.55  & 15.30 & 19.94      & 11.46 & 13.88  & 12.62 (\textbf{+0.77})  \\
                     & LRD(r=10)   & 3.03 & 14.82 & 8.46  & 15.82 & 17.84      & 11.57 & 16.95  & 12.64 (\textbf{+0.79})  \\
                     & CS-LRD(r=9) & 3.75 & 16.53 & 13.88 & 17.39 & 26.75      & 22.39 & 26.55  & 18.18 (\textbf{+6.33})  \\ \hline
\multirow{4}{*}{Source Included Multilingual} & Original & 2.40 & 9.75  & 2.28  & 10.59 & 2.69  & 0.85  & 3.66  & 4.6   \\
                     & Centering   & 1.73 & 8.80  & 2.25  & 8.44  & 3.11       & 0.51  & 3.43   & 4.04 (\textbf{-0.56})   \\
                     & LRD(r=10)   & 2.45 & 10.06 & 2.37  & 11.01 & 2.98       & 0.99  & 4.29   & 4.88 (\textbf{+0.28})   \\
                     & CS-LRD(r=9) & 2.98 & 11.94 & 2.73  & 12.95 & 4.59       & 1.46  & 6.61   & 6.18 (\textbf{+1.58})   \\ \hline
\end{tabular}%

%% file: tables/c2c/zeroshot/c2c_unixcoder_mean.tex
\begin{tabular}{c|c|cccccccc}
\hline
UnixCoder (mean)                              &             & C     & C\#    & C++   & Java  & Javascript & PHP   & Python & Avg.                   \\ \hline
\multirow{4}{*}{Monolingual}                  & Original    & 95.27 & 98.31 & 98.12 & 98.19 & 97.48      & 97.76 & 98.18  & 97.62                  \\
                                              & Centering   & 95.59 & 98.28 & 98.23 & 98.43 & 97.30      & 97.76 & 98.13  & 97.67 (\textbf{+0.05}) \\
                                              & LRD(r=10)   & 95.35 & 98.31 & 98.12 & 98.20 & 97.47      & 97.76 & 98.19  & 97.63 (\textbf{+0.01}) \\
                                              & CS-LRD(r=9) & 95.16 & 98.34 & 98.27 & 98.43 & 97.57      & 97.76 & 98.22  & 97.68 (\textbf{+0.06}) \\ \hline
\multirow{4}{*}{Source Excluded Multilingual} & Original    & 78.34 & 87.63 & 94.19 & 87.44 & 91.51      & 90.18 & 92.77  & 88.87                  \\
                                              & Centering   & 77.94 & 87.02 & 93.94 & 87.55 & 90.77      & 89.48 & 92.24  & 88.42 (\textbf{-0.45}) \\
                                              & LRD(r=10)   & 78.53 & 87.68 & 94.21 & 87.44 & 91.56      & 90.16 & 92.64  & 88.89 (\textbf{+0.02}) \\
                                              & CS-LRD(r=9) & 79.32 & 88.45 & 94.39 & 88.10 & 91.76      & 90.57 & 93.28  & 89.41 (\textbf{+0.54}) \\ \hline
\multirow{4}{*}{Source Included Multilingual} & Original    & 71.19 & 82.03 & 85.87 & 81.45 & 84.75      & 85.21 & 87.57  & 82.58                  \\
                                              & Centering   & 69.67 & 80.80 & 85.86 & 80.92 & 83.05      & 82.98 & 85.85  & 81.30 (\textbf{-1.28}) \\
                                              & LRD(r=10)   & 71.26 & 82.07 & 86.02 & 81.47 & 84.75      & 85.25 & 87.57  & 82.63 (\textbf{+0.05}) \\
                                              & CS-LRD(r=9) & 72.28 & 82.92 & 86.71 & 82.07 & 85.29      & 85.81 & 88.36  & 83.35 (\textbf{+0.77}) \\ \hline
\end{tabular}%

%% file: tables/c2c/zeroshot/c2c_unixcoder_cls.tex
\begin{tabular}{c|c|cccccccc}
\hline
UnixCoder (cls)                               &             & C     & C\#    & C++   & Java  & Javascript & PHP   & Python & Avg.                   \\ \hline
\multirow{4}{*}{Monolingual}                  & Original    & 93.63 & 97.29 & 97.54 & 97.79 & 96.88      & 96.83 & 97.55  & 96.79                  \\
                                              & Centering   & 93.76 & 97.57 & 97.66 & 97.99 & 97.12      & 97.14 & 97.68  & 96.99 (\textbf{+0.20}) \\
                                              & LRD(r=10)   & 93.55 & 97.29 & 97.56 & 97.82 & 96.92      & 96.83 & 97.55  & 96.79 (\textbf{+0.00}) \\
                                              & CS-LRD(r=9) & 94.29 & 97.64 & 97.68 & 98.04 & 97.02      & 96.95 & 97.61  & 97.03 (\textbf{+0.24}) \\ \hline
\multirow{4}{*}{Source Excluded Multilingual} & Original    & 75.57 & 81.52 & 92.98 & 82.07 & 89.90      & 88.45 & 91.66  & 86.02                  \\
                                              & Centering   & 74.99 & 81.35 & 92.77 & 82.34 & 89.74      & 87.65 & 91.83  & 85.81 (\textbf{-0.21}) \\
                                              & LRD(r=10)   & 75.75 & 81.63 & 93.02 & 82.11 & 89.89      & 88.50 & 91.66  & 86.08 (\textbf{+0.06}) \\
                                              & CS-LRD(r=9) & 77.26 & 82.87 & 93.12 & 83.37 & 90.45      & 88.52 & 92.46  & 86.86 (\textbf{+0.84}) \\ \hline
\multirow{4}{*}{Source Included Multilingual} & Original    & 67.67 & 75.89 & 81.38 & 75.50 & 80.70      & 81.39 & 84.61  & 78.16                  \\
                                              & Centering   & 65.04 & 73.93 & 81.83 & 74.42 & 79.12      & 78.96 & 83.49  & 76.68 (\textbf{-1.48}) \\
                                              & LRD(r=10)   & 67.84 & 76.03 & 81.64 & 75.60 & 80.81      & 81.56 & 84.68  & 78.31 (\textbf{+0.15}) \\
                                              & CS-LRD(r=9) & 69.63 & 77.71 & 83.09 & 77.18 & 81.50      & 81.61 & 85.32  & 79.43 (\textbf{+1.27}) \\ \hline
\end{tabular}%

%% file: tables/c2c/zeroshot/c2c_unixcoder_pooler.tex
\begin{tabular}{c|c|cccccccc}
\hline
Unixcoder (pooler)                            &             & C     & C\#    & C++   & Java  & Javascript & PHP   & Python & Avg.                   \\ \hline
\multirow{4}{*}{Monolingual}                  & Original    & 93.17 & 97.16 & 97.30 & 97.68 & 96.45      & 96.59 & 97.26  & 96.52                  \\
                                              & Centering   & 93.31 & 97.36 & 97.47 & 98.01 & 96.78      & 96.91 & 97.54  & 96.77 (\textbf{+0.25}) \\
                                              & LRD(r=10)   & 93.09 & 97.22 & 97.40 & 97.75 & 96.50      & 96.66 & 97.29  & 96.56 (\textbf{+0.04}) \\
                                              & CS-LRD(r=9) & 93.59 & 97.38 & 97.53 & 97.92 & 96.64      & 96.72 & 97.43  & 96.74 (\textbf{+0.22}) \\ \hline
\multirow{4}{*}{Source Excluded Multilingual} & Original    & 74.44 & 80.66 & 92.35 & 80.81 & 88.70      & 87.47 & 90.75  & 85.03                  \\
                                              & Centering   & 74.42 & 80.47 & 92.28 & 81.85 & 88.59      & 87.43 & 91.29  & 85.19 (\textbf{+0.16}) \\
                                              & LRD(r=10)   & 74.50 & 80.92 & 92.57 & 81.09 & 88.87      & 87.72 & 90.82  & 85.21 (\textbf{+0.18}) \\
                                              & CS-LRD(r=9) & 75.75 & 81.76 & 92.61 & 82.48 & 89.35      & 87.89 & 91.71  & 85.94 (\textbf{+0.91}) \\ \hline
\multirow{4}{*}{Source Included Multilingual} & Original    & 67.00 & 75.07 & 80.49 & 74.39 & 79.36      & 80.87 & 83.59  & 77.25                  \\
                                              & Centering   & 65.36 & 73.11 & 81.22 & 74.22 & 77.48      & 79.12 & 82.82  & 76.19 (\textbf{-1.06}) \\
                                              & LRD(r=10)   & 67.05 & 75.41 & 80.83 & 74.67 & 79.61      & 80.99 & 83.93  & 77.50 (\textbf{+0.25}) \\
                                              & CS-LRD(r=9) & 68.59 & 76.69 & 81.80 & 76.62 & 80.20      & 81.09 & 84.44  & 78.49 (\textbf{+1.24}) \\ \hline
\end{tabular}%

%% file: tables/t2c/zeroshot/t2c_codebert_mean.tex
\begin{tabular}{c|c|ccccccc}
\hline
CodeBERT (mean)               &             & Go   & Java & Javascript & PHP  & Python & Ruby & Avg.                  \\ \hline
\multirow{4}{*}{Monolingual}  & Original    & 0.15 & 0.04 & 0.06       & 0.03 & 0.06   & 0.37 & 0.12                  \\
                              & Centering   & 0.13 & 0.26 & 0.29       & 0.19 & 0.31   & 1.04 & 0.37 (\textbf{+0.25}) \\
                              & LRD(r=10)   & 0.18 & 0.04 & 0.06       & 0.03 & 0.07   & 0.40 & 0.13 (\textbf{+0.01}) \\
                              & CS-LRD(r=6) & 0.33 & 0.15 & 0.18       & 0.06 & 0.27   & 0.87 & 0.31 (\textbf{+0.19}) \\ \hline
\multirow{4}{*}{Multilingual} & Original    & 0.07 & 0.01 & 0.00       & 0.00 & 0.02   & 0.27 & 0.06                  \\
                              & Centering   & 0.02 & 0.03 & 0.04       & 0.05 & 0.24   & 0.27 & 0.11 (\textbf{+0.05}) \\
                              & LRD(r=10)   & 0.09 & 0.01 & 0.00       & 0.00 & 0.02   & 0.30 & 0.07 (\textbf{+0.01}) \\
                              & CS-LRD(r=6) & 0.13 & 0.05 & 0.02       & 0.00 & 0.19   & 0.41 & 0.13 (\textbf{+0.07}) \\ \hline
\end{tabular}%

%% file: tables/t2c/zeroshot/t2c_graphcodebert_mean.tex
\begin{tabular}{c|c|ccccccc}
\hline
GraphCodeBert (mean)          &             & Go    & Java  & Javascript & PHP   & Python & Ruby  & Avg.                   \\ \hline
\multirow{4}{*}{Monolingual}  & Original    & 12.48 & 8.60  & 7.30       & 8.08  & 10.38  & 20.80 & 11.27                  \\
                              & Centering   & 19.30 & 17.32 & 18.14      & 14.62 & 18.53  & 31.59 & 19.92 (\textbf{+8.65}) \\
                              & LRD(r=10)   & 14.85 & 10.10 & 8.58       & 9.20  & 12.07  & 22.94 & 12.96 (\textbf{+1.69}) \\
                              & CS-LRD(r=6) & 15.94 & 11.86 & 8.07       & 10.22 & 13.09  & 24.05 & 13.87 (\textbf{+2.60}) \\ \hline
\multirow{4}{*}{Multilingual} & Original    & 5.49  & 7.41  & 3.01       & 4.05  & 7.39   & 6.63  & 5.66                   \\
                              & Centering   & 8.60  & 12.07 & 7.58       & 9.28  & 12.26  & 20.01 & 11.63 (\textbf{+5.97}) \\
                              & LRD(r=10)   & 6.75  & 8.70  & 3.47       & 4.77  & 8.70   & 7.85  & 6.71 (\textbf{+1.05})  \\
                              & CS-LRD(r=6) & 7.60  & 9.29  & 3.28       & 4.89  & 10.08  & 14.50 & 8.27 (\textbf{+2.61})  \\ \hline
\end{tabular}%

%% file: tables/t2c/zeroshot/t2c_starencoder_mean.tex
\begin{tabular}{c|c|ccccccc}
\hline
StarEncoder (mean)            &             & Go    & Ruby  & Java  & Javascript & PHP  & Python & Avg.                    \\ \hline
\multirow{4}{*}{Monolingual}  & Original    & 1.85  & 4.41  & 1.89  & 1.55       & 0.57 & 2.14   & 2.07                    \\
                              & Centering   & 18.00 & 18.98 & 10.65 & 10.52      & 6.95 & 10.71  & 12.64 (\textbf{+10.57}) \\
                              & LRD(r=10)   & 2.08  & 4.88  & 2.21  & 1.76       & 0.72 & 2.52   & 2.36 (\textbf{+0.29})   \\
                              & CS-LRD(r=9) & 3.07  & 7.60  & 3.93  & 2.71       & 1.68 & 4.09   & 3.85 (\textbf{+1.78})   \\ \hline
\multirow{4}{*}{Multilingual} & Original    & 0.96  & 1.88  & 1.33  & 0.75       & 0.16 & 1.80   & 1.15                    \\
                              & Centering   & 5.80  & 9.92  & 5.92  & 4.48       & 1.51 & 8.41   & 6.01 (\textbf{+4.86})   \\
                              & LRD(r=10)   & 1.06  & 2.18  & 1.60  & 0.88       & 0.21 & 2.08   & 1.34 (\textbf{+0.19})   \\
                              & CS-LRD(r=9) & 1.09  & 4.28  & 2.69  & 1.31       & 0.49 & 3.43   & 2.22 (\textbf{+1.07})   \\ \hline
\end{tabular}%

%% file: tables/t2c/zeroshot/t2c_codet5+_pooler.tex
\begin{tabular}{c|c|ccccccc}
\hline
CodeT5+ (pooler)              &             & Go    & Ruby  & Java  & Javascript & PHP   & Python & Avg.                   \\ \hline
\multirow{4}{*}{Monolingual}  & Original    & 90.74 & 74.45 & 71.82 & 69.18      & 67.82 & 71.72  & 74.29                  \\
                              & Centering   & 89.98 & 73.38 & 70.36 & 67.71      & 65.57 & 70.07  & 72.84 (\textbf{-1.45}) \\
                              & LRD(r=1)    & 90.42 & 73.86 & 71.18 & 68.45      & 67.03 & 71.10  & 73.67 (\textbf{-0.62}) \\
                              & CS-LRD(r=1) & 90.69 & 74.32 & 71.90 & 69.13      & 67.81 & 71.60  & 74.24 (\textbf{-0.05}) \\ \hline
\multirow{4}{*}{Multilingual} & Original    & 89.40 & 55.82 & 65.60 & 58.65      & 63.36 & 67.32  & 66.69                  \\
                              & Centering   & 86.89 & 58.09 & 59.46 & 52.24      & 55.43 & 65.75  & 62.98 (\textbf{-3.71}) \\
                              & LRD(r=1)    & 88.83 & 56.69 & 63.76 & 55.46      & 60.74 & 67.03  & 65.42 (\textbf{-1.27}) \\
                              & CS-LRD(r=1) & 89.37 & 55.65 & 65.93 & 58.35      & 63.17 & 67.08  & 66.59 (\textbf{-0.10}) \\ \hline
\end{tabular}%

%% file: acl_latex.bbl
\begin{thebibliography}{37}
\expandafter\ifx\csname natexlab\endcsname\relax\def\natexlab#1{#1}\fi

\bibitem[{Ahmad et~al.(2021)Ahmad, Chakraborty, Ray, and Chang}]{ahmad2021unified}
Wasi Ahmad, Saikat Chakraborty, Baishakhi Ray, and Kai-Wei Chang. 2021.
\newblock Unified pre-training for program understanding and generation.
\newblock In \emph{Proceedings of the 2021 Conference of the North American Chapter of the Association for Computational Linguistics: Human Language Technologies}, pages 2655--2668.

\bibitem[{Allal et~al.(2023)Allal, Li, Kocetkov, Mou, Akiki, Ferrandis, Muennighoff, Mishra, Gu, Dey et~al.}]{allal2023santacoder}
Loubna~Ben Allal, Raymond Li, Denis Kocetkov, Chenghao Mou, Christopher Akiki, Carlos~Munoz Ferrandis, Niklas Muennighoff, Mayank Mishra, Alex Gu, Manan Dey, et~al. 2023.
\newblock Santacoder: don't reach for the stars!
\newblock \emph{arXiv preprint arXiv:2301.03988}.

\bibitem[{Athiwaratkun et~al.(2022)Athiwaratkun, Gouda, Wang, Li, Tian, Tan, Ahmad, Wang, Sun, Shang et~al.}]{athiwaratkun2022multi}
Ben Athiwaratkun, Sanjay~Krishna Gouda, Zijian Wang, Xiaopeng Li, Yuchen Tian, Ming Tan, Wasi~Uddin Ahmad, Shiqi Wang, Qing Sun, Mingyue Shang, et~al. 2022.
\newblock Multi-lingual evaluation of code generation models.
\newblock In \emph{The Eleventh International Conference on Learning Representations}.

\bibitem[{Brown et~al.(2020)Brown, Mann, Ryder, Subbiah, Kaplan, Dhariwal, Neelakantan, Shyam, Sastry, Askell et~al.}]{brown2020language}
Tom Brown, Benjamin Mann, Nick Ryder, Melanie Subbiah, Jared~D Kaplan, Prafulla Dhariwal, Arvind Neelakantan, Pranav Shyam, Girish Sastry, Amanda Askell, et~al. 2020.
\newblock Language models are few-shot learners.
\newblock \emph{Advances in neural information processing systems}, 33:1877--1901.

\bibitem[{Chang et~al.(2022)Chang, Tu, and Bergen}]{chang2022geometry}
Tyler Chang, Zhuowen Tu, and Benjamin Bergen. 2022.
\newblock The geometry of multilingual language model representations.
\newblock In \emph{Proceedings of the 2022 Conference on Empirical Methods in Natural Language Processing}, pages 119--136.

\bibitem[{Chen et~al.(2021)Chen, Tworek, Jun, Yuan, Pinto, Kaplan, Edwards, Burda, Joseph, Brockman et~al.}]{chen2021evaluating}
Mark Chen, Jerry Tworek, Heewoo Jun, Qiming Yuan, Henrique Ponde de~Oliveira Pinto, Jared Kaplan, Harri Edwards, Yuri Burda, Nicholas Joseph, Greg Brockman, et~al. 2021.
\newblock Evaluating large language models trained on code.
\newblock \emph{arXiv preprint arXiv:2107.03374}.

\bibitem[{Devlin et~al.(2019)Devlin, Chang, Lee, and Toutanova}]{devlin-etal-2019-bert}
Jacob Devlin, Ming-Wei Chang, Kenton Lee, and Kristina Toutanova. 2019.
\newblock {BERT}: Pre-training of deep bidirectional transformers for language understanding.
\newblock In \emph{Proceedings of the 2019 Conference of the North {A}merican Chapter of the Association for Computational Linguistics: Human Language Technologies, Volume 1 (Long and Short Papers)}, pages 4171--4186.

\bibitem[{Feng et~al.(2020)Feng, Guo, Tang, Duan, Feng, Gong, Shou, Qin, Liu, Jiang et~al.}]{feng2020codebert}
Zhangyin Feng, Daya Guo, Duyu Tang, Nan Duan, Xiaocheng Feng, Ming Gong, Linjun Shou, Bing Qin, Ting Liu, Daxin Jiang, et~al. 2020.
\newblock Codebert: A pre-trained model for programming and natural languages.
\newblock In \emph{Findings of the Association for Computational Linguistics: EMNLP 2020}, pages 1536--1547.

\bibitem[{Guo et~al.(2022)Guo, Lu, Duan, Wang, Zhou, and Yin}]{guo2022unixcoder}
Daya Guo, Shuai Lu, Nan Duan, Yanlin Wang, Ming Zhou, and Jian Yin. 2022.
\newblock Unixcoder: Unified cross-modal pre-training for code representation.
\newblock In \emph{Proceedings of the 60th Annual Meeting of the Association for Computational Linguistics (Volume 1: Long Papers)}, pages 7212--7225.

\bibitem[{Guo et~al.(2020)Guo, Ren, Lu, Feng, Tang, Shujie, Zhou, Duan, Svyatkovskiy, Fu et~al.}]{guo2020graphcodebert}
Daya Guo, Shuo Ren, Shuai Lu, Zhangyin Feng, Duyu Tang, LIU Shujie, Long Zhou, Nan Duan, Alexey Svyatkovskiy, Shengyu Fu, et~al. 2020.
\newblock Graphcodebert: Pre-training code representations with data flow.
\newblock In \emph{International Conference on Learning Representations}.

\bibitem[{Husain et~al.(2019)Husain, Wu, Gazit, Allamanis, and Brockschmidt}]{husain2019codesearchnet}
Hamel Husain, Ho-Hsiang Wu, Tiferet Gazit, Miltiadis Allamanis, and Marc Brockschmidt. 2019.
\newblock Codesearchnet challenge: Evaluating the state of semantic code search.
\newblock \emph{arXiv preprint arXiv:1909.09436}.

\bibitem[{Jain et~al.(2021)Jain, Jain, Zhang, Abbeel, Gonzalez, and Stoica}]{jain-etal-2021-contrastive}
Paras Jain, Ajay Jain, Tianjun Zhang, Pieter Abbeel, Joseph Gonzalez, and Ion Stoica. 2021.
\newblock Contrastive code representation learning.
\newblock In \emph{Proceedings of the 2021 Conference on Empirical Methods in Natural Language Processing}, pages 5954--5971.

\bibitem[{Kocetkov et~al.(2022)Kocetkov, Li, Jia, Mou, Jernite, Mitchell, Ferrandis, Hughes, Wolf, Bahdanau et~al.}]{kocetkov2022stack}
Denis Kocetkov, Raymond Li, LI~Jia, Chenghao Mou, Yacine Jernite, Margaret Mitchell, Carlos~Mu{\~n}oz Ferrandis, Sean Hughes, Thomas Wolf, Dzmitry Bahdanau, et~al. 2022.
\newblock The stack: 3 tb of permissively licensed source code.
\newblock \emph{Transactions on Machine Learning Research}.

\bibitem[{Kulshreshtha et~al.(2020)Kulshreshtha, Garcia, and Chang}]{kulshreshtha2020cross}
Saurabh Kulshreshtha, Jose Luis~Redondo Garcia, and Ching~Yun Chang. 2020.
\newblock Cross-lingual alignment methods for multilingual bert: A comparative study.
\newblock In \emph{Findings of the Association for Computational Linguistics: EMNLP 2020}, pages 933--942.

\bibitem[{Li et~al.(2023)Li, Allal, Zi, Muennighoff, Kocetkov, Mou, Marone, Akiki, Li, Chim et~al.}]{li2023starcoder}
Raymond Li, Loubna~Ben Allal, Yangtian Zi, Niklas Muennighoff, Denis Kocetkov, Chenghao Mou, Marc Marone, Christopher Akiki, Jia Li, Jenny Chim, et~al. 2023.
\newblock Starcoder: may the source be with you!
\newblock \emph{arXiv preprint arXiv:2305.06161}.

\bibitem[{Libovick{\`y} et~al.(2020)Libovick{\`y}, Rosa, and Fraser}]{libovicky2020language}
Jind{\v{r}}ich Libovick{\`y}, Rudolf Rosa, and Alexander Fraser. 2020.
\newblock On the language neutrality of pre-trained multilingual representations.
\newblock In \emph{Findings of the Association for Computational Linguistics: EMNLP 2020}, pages 1663--1674.

\bibitem[{Luo et~al.(2023)Luo, Xu, Zhao, Sun, Geng, Hu, Tao, Ma, Lin, and Jiang}]{luo2023wizardcoder}
Ziyang Luo, Can Xu, Pu~Zhao, Qingfeng Sun, Xiubo Geng, Wenxiang Hu, Chongyang Tao, Jing Ma, Qingwei Lin, and Daxin Jiang. 2023.
\newblock Wizardcoder: Empowering code large language models with evol-instruct.
\newblock \emph{arXiv preprint arXiv:2306.08568}.

\bibitem[{Nijkamp et~al.(2023)Nijkamp, Hayashi, Xiong, Savarese, and Zhou}]{nijkamp2023codegen2}
Erik Nijkamp, Hiroaki Hayashi, Caiming Xiong, Silvio Savarese, and Yingbo Zhou. 2023.
\newblock Codegen2: Lessons for training llms on programming and natural languages.
\newblock \emph{arXiv preprint arXiv:2305.02309}.

\bibitem[{Nijkamp et~al.(2022)Nijkamp, Pang, Hayashi, Tu, Wang, Zhou, Savarese, and Xiong}]{nijkamp2022codegen}
Erik Nijkamp, Bo~Pang, Hiroaki Hayashi, Lifu Tu, Huan Wang, Yingbo Zhou, Silvio Savarese, and Caiming Xiong. 2022.
\newblock Codegen: An open large language model for code with multi-turn program synthesis.
\newblock In \emph{The Eleventh International Conference on Learning Representations}.

\bibitem[{Oord et~al.(2018)Oord, Li, and Vinyals}]{oord2018representation}
Aaron van~den Oord, Yazhe Li, and Oriol Vinyals. 2018.
\newblock Representation learning with contrastive predictive coding.
\newblock \emph{arXiv preprint arXiv:1807.03748}.

\bibitem[{Piratla et~al.(2020)Piratla, Netrapalli, and Sarawagi}]{piratla2020efficient}
Vihari Piratla, Praneeth Netrapalli, and Sunita Sarawagi. 2020.
\newblock Efficient domain generalization via common-specific low-rank decomposition.
\newblock In \emph{International Conference on Machine Learning}, pages 7728--7738. PMLR.

\bibitem[{Pires et~al.(2019)Pires, Schlinger, and Garrette}]{pires2019multilingual}
Telmo Pires, Eva Schlinger, and Dan Garrette. 2019.
\newblock How multilingual is multilingual bert?
\newblock In \emph{Proceedings of the 57th Annual Meeting of the Association for Computational Linguistics}, pages 4996--5001.

\bibitem[{Radford et~al.(2019)Radford, Wu, Child, Luan, Amodei, Sutskever et~al.}]{radford2019language}
Alec Radford, Jeffrey Wu, Rewon Child, David Luan, Dario Amodei, Ilya Sutskever, et~al. 2019.
\newblock Language models are unsupervised multitask learners.
\newblock \emph{OpenAI blog}.

\bibitem[{Raffel et~al.(2020)Raffel, Shazeer, Roberts, Lee, Narang, Matena, Zhou, Li, and Liu}]{raffel2020exploring}
Colin Raffel, Noam Shazeer, Adam Roberts, Katherine Lee, Sharan Narang, Michael Matena, Yanqi Zhou, Wei Li, and Peter~J Liu. 2020.
\newblock Exploring the limits of transfer learning with a unified text-to-text transformer.
\newblock \emph{The Journal of Machine Learning Research}, 21(1):5485--5551.

\bibitem[{Roy et~al.(2020)Roy, Constant, Al-Rfou, Barua, Phillips, and Yang}]{roy2020lareqa}
Uma Roy, Noah Constant, Rami Al-Rfou, Aditya Barua, Aaron Phillips, and Yinfei Yang. 2020.
\newblock Lareqa: Language-agnostic answer retrieval from a multilingual pool.
\newblock In \emph{Proceedings of the 2020 Conference on Empirical Methods in Natural Language Processing (EMNLP)}, pages 5919--5930.

\bibitem[{Roziere et~al.(2023)Roziere, Gehring, Gloeckle, Sootla, Gat, Tan, Adi, Liu, Remez, Rapin et~al.}]{roziere2023code}
Baptiste Roziere, Jonas Gehring, Fabian Gloeckle, Sten Sootla, Itai Gat, Xiaoqing~Ellen Tan, Yossi Adi, Jingyu Liu, Tal Remez, J{\'e}r{\'e}my Rapin, et~al. 2023.
\newblock Code llama: Open foundation models for code.
\newblock \emph{arXiv preprint arXiv:2308.12950}.

\bibitem[{Schmidt(1907)}]{schmidt1907theorie}
Erhard Schmidt. 1907.
\newblock Zur theorie der linearen und nichtlinearen integralgleichungen.
\newblock \emph{Mathematische Annalen}, 63(4):433--476.

\bibitem[{Schuster et~al.(2019)Schuster, Ram, Barzilay, and Globerson}]{schuster2019cross}
Tal Schuster, Ori Ram, Regina Barzilay, and Amir Globerson. 2019.
\newblock Cross-lingual alignment of contextual word embeddings, with applications to zero-shot dependency parsing.
\newblock In \emph{Proceedings of the 2019 Conference of the North American Chapter of the Association for Computational Linguistics: Human Language Technologies, Volume 1 (Long and Short Papers)}, pages 1599--1613.

\bibitem[{Wang et~al.(2021{\natexlab{a}})Wang, Wang, Mi, Zhou, Wan, Liu, Li, Wu, Liu, and Jiang}]{wang2021syncobert}
Xin Wang, Yasheng Wang, Fei Mi, Pingyi Zhou, Yao Wan, Xiao Liu, Li~Li, Hao Wu, Jin Liu, and Xin Jiang. 2021{\natexlab{a}}.
\newblock Syncobert: Syntax-guided multi-modal contrastive pre-training for code representation.
\newblock \emph{arXiv preprint arXiv:2108.04556}.

\bibitem[{Wang et~al.(2023)Wang, Le, Gotmare, Bui, Li, and Hoi}]{wang2023codet5+}
Yue Wang, Hung Le, Akhilesh~Deepak Gotmare, Nghi~DQ Bui, Junnan Li, and Steven~CH Hoi. 2023.
\newblock Codet5+: Open code large language models for code understanding and generation.
\newblock \emph{arXiv preprint arXiv:2305.07922}.

\bibitem[{Wang et~al.(2021{\natexlab{b}})Wang, Wang, Joty, and Hoi}]{wang2021codet5}
Yue Wang, Weishi Wang, Shafiq Joty, and Steven~CH Hoi. 2021{\natexlab{b}}.
\newblock Codet5: Identifier-aware unified pre-trained encoder-decoder models for code understanding and generation.
\newblock In \emph{Proceedings of the 2021 Conference on Empirical Methods in Natural Language Processing}, pages 8696--8708.

\bibitem[{Xie et~al.(2022)Xie, Zhao, Yu, and Li}]{xie2022discovering}
Zhihui Xie, Handong Zhao, Tong Yu, and Shuai Li. 2022.
\newblock Discovering low-rank subspaces for language-agnostic multilingual representations.
\newblock In \emph{Proceedings of the 2022 Conference on Empirical Methods in Natural Language Processing}, pages 5617--5633.

\bibitem[{Xu et~al.(2022)Xu, Alon, Neubig, and Hellendoorn}]{xu2022systematic}
Frank~F Xu, Uri Alon, Graham Neubig, and Vincent~Josua Hellendoorn. 2022.
\newblock A systematic evaluation of large language models of code.
\newblock In \emph{Proceedings of the 6th ACM SIGPLAN International Symposium on Machine Programming}, pages 1--10.

\bibitem[{Yang et~al.(2021)Yang, Yang, Cer, and Darve}]{yang2021simple}
Ziyi Yang, Yinfei Yang, Daniel Cer, and Eric Darve. 2021.
\newblock A simple and effective method to eliminate the self language bias in multilingual representations.
\newblock In \emph{Proceedings of the 2021 Conference on Empirical Methods in Natural Language Processing}, pages 5825--5832.

\bibitem[{Zheng et~al.(2023)Zheng, Xia, Zou, Dong, Wang, Xue, Wang, Shen, Wang, Li et~al.}]{zheng2023codegeex}
Qinkai Zheng, Xiao Xia, Xu~Zou, Yuxiao Dong, Shan Wang, Yufei Xue, Zihan Wang, Lei Shen, Andi Wang, Yang Li, et~al. 2023.
\newblock Codegeex: A pre-trained model for code generation with multilingual evaluations on humaneval-x.
\newblock \emph{arXiv preprint arXiv:2303.17568}.

\bibitem[{Zhu et~al.(2022)Zhu, Jain, Suresh, Ravindran, Tipirneni, and Reddy}]{zhu2022xlcost}
Ming Zhu, Aneesh Jain, Karthik Suresh, Roshan Ravindran, Sindhu Tipirneni, and Chandan~K Reddy. 2022.
\newblock Xlcost: A benchmark dataset for cross-lingual code intelligence.
\newblock \emph{arXiv preprint arXiv:2206.08474}.

\bibitem[{Z{\"u}gner et~al.(2020)Z{\"u}gner, Kirschstein, Catasta, Leskovec, and G{\"u}nnemann}]{zugner2020language}
Daniel Z{\"u}gner, Tobias Kirschstein, Michele Catasta, Jure Leskovec, and Stephan G{\"u}nnemann. 2020.
\newblock Language-agnostic representation learning of source code from structure and context.
\newblock In \emph{International Conference on Learning Representations}.

\end{thebibliography}
